\documentclass{article} %
\usepackage{configuration/iclr2025_conference,times}

\usepackage{hyperref}
\usepackage{wrapfig}
\usepackage[utf8]{inputenc}
\usepackage{longtable}

\usepackage{amsmath,amssymb,amsthm}
\newtheorem{theorem}{Theorem}[section]
\newtheorem{lemma}[theorem]{Lemma}

\newtheorem{definition}[theorem]{Definition}

\newtheorem{remark}[theorem]{Remark}

\providecommand{\abs}[1]{\left\lvert#1\right\rvert}

\providecommand{\R}{\mathbb{R}} %

\renewcommand{\aa}{\mathbf{a}}

\providecommand{\hh}{\mathbf{h}}

\providecommand{\mm}{\mathbf{m}}

\providecommand{\pp}{\mathbf{p}}

\providecommand{\xx}{\mathbf{x}}
\providecommand{\yy}{\mathbf{y}}

\providecommand{\mD}{\mathbf{D}}

\providecommand{\mP}{\mathbf{P}}

\providecommand{\mTheta}{\boldsymbol{\Theta}}

\providecommand{\mtheta}{\boldsymbol{\theta}}

\providecommand{\cT}{\mathcal{T}}

\newenvironment{talign*}
{\csname align*\endcsname}
{\endalign}

\newcommand*\circled[1]{\tikz[baseline=(char.base)]{
        \node[shape=circle,draw,inner sep=0.5pt] (char) {#1};}}

\usepackage[utf8]{inputenc}         %
\usepackage[T1]{fontenc}            %
\usepackage{url}                    %
\usepackage{booktabs}               %
\usepackage{amsfonts}               %
\usepackage{nicefrac}               %
\usepackage{microtype}              %
\usepackage{xcolor}                 %
\usepackage{algorithm}
\usepackage{algorithmic}
\usepackage{graphicx}
\usepackage{subcaption}
\usepackage[flushleft]{threeparttable}
\usepackage{float}
\usepackage{multirow}
\usepackage{xspace}
\usepackage{natbib}
\usepackage{enumitem}
\usepackage[font=small]{caption}
\usepackage{autobreak}
\usepackage{sidecap}
\usepackage{wrapfig}
\usepackage{bbding}
\usepackage[toc, page, header]{appendix}
\usepackage{tikz}
\usepackage{xcolor}
\usepackage{pifont}
\usepackage{mdframed}

\hypersetup{
    colorlinks=true,
    linkcolor=blue,
    citecolor=blue,
    urlcolor=blue
}

\definecolor{coral}{RGB}{255,127,80}
\definecolor{darkgreen}{RGB}{0,100,0}
\definecolor{darkyellow}{RGB}{204,153,0}
\definecolor{salmon}{RGB}{250,128,114}

\usepackage{xspace}
\newcommand{\alg}{\textsc{ELICIT}\xspace}

\usepackage[textwidth=3.5cm]{todonotes}

\title{\alg: LLM Augmentation via External In-Context Capability}

\author{Futing Wang \textsuperscript{1, 2} \thanks{~~These authors contributed equally to this work.}  \quad
    Jianhao Yan \textsuperscript{1, 2} \footnotemark[1] \quad
    Yue Zhang \textsuperscript{2, 3} \thanks{~~Corresponding author.}  \quad
    Tao Lin \textsuperscript{2, 4} \footnotemark[2] \quad \\
    Zhejiang University  \textsuperscript{1}\quad 
    Westlake University  \textsuperscript{2} \\
    Institute of Advanced Technology, Westlake Institute for Advanced Study  \textsuperscript{3}\\
    Research Center for Industries of the Future, Westlake University \textsuperscript{4}\\
    \texttt{\{wangfuting, yanjianhao, zhangyue, lintao\}@westlake.edu.cn}
}

\iclrfinalcopy %
\begin{document}

\maketitle

\begin{abstract}
    Enhancing the adaptive capabilities of large language models is a critical pursuit in both research and application.
    Traditional fine-tuning methods require substantial data and computational resources, especially for enhancing specific capabilities, while in-context learning is limited by the need for appropriate demonstrations and efficient token usage.
    Inspired by the expression of in-context learned capabilities through task vectors and the concept of modularization, we propose \alg, a framework consisting of two modules designed to effectively store and reuse task vectors to elicit the diverse  capabilities of models without additional training or inference tokens.
    Our comprehensive experiments and analysis demonstrate that our pipeline is highly transferable across different input formats, tasks, and model architectures.
    \alg serves as a plug-and-play performance booster to enable adaptive elicitation of model capabilities.
    By externally storing and reusing vectors that represent in-context learned capabilities, \alg not only demonstrates the potential to operate modular capabilities but also significantly enhances the performance, versatility, adaptability, and scalability of large language models.
    Our code is publicly available \footnote{\href{https://github.com/LINs-lab/ELICIT}{https://github.com/LINs-lab/ELICIT}}.
    \looseness=-1
\end{abstract}

\section{Introduction}

\begin{wrapfigure}{r}{0.5\textwidth}
    \centering
    \vspace{-10pt}
    \includegraphics[width=0.9\linewidth]{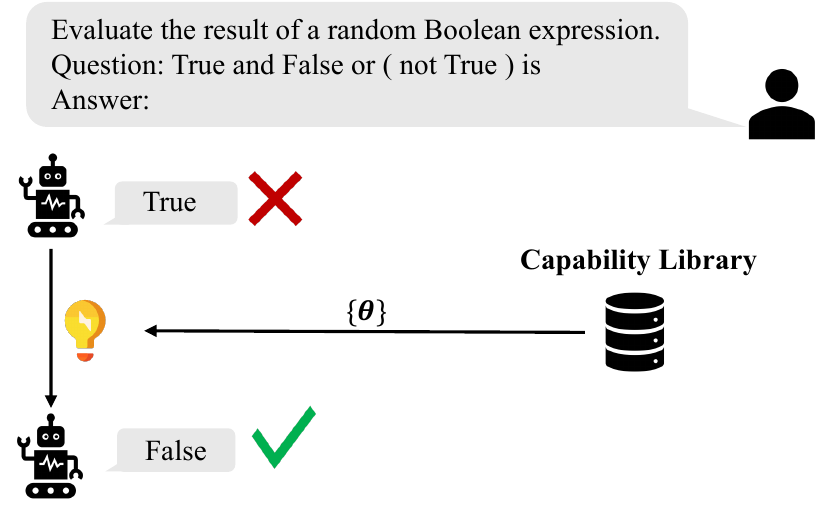}
    \vspace{-12pt}
    \caption{\small
        \textbf{Illustration of \alg}, which dynamically retrieves and integrates task vectors from a capability library to augment a language model's performance on arbitrary queries, without increasing token usage during inference.
        \looseness=-1
    }
    \label{fig:illustration}
    \vspace{-15pt}
\end{wrapfigure}

Large Language Models (LLMs) have revolutionized the field of Natural Language Processing (NLP), demonstrating remarkable versatility in tackling a wide array of tasks and real-world challenges~\citep{devlin2018bert,brown2020language,han2021pre,achiam2023gpt,touvron2023llama}.
The power of these models lies in their ability to seamlessly integrate various \emph{capabilities}, from logical reasoning~\citep{bommasani2021opportunities} to common sense understanding~\citep{talmor2018commonsenseqa}.
In our rapidly evolving world, a crucial aspect of LLM is the ability to efficiently adapt to new tasks or scenarios.

Traditional fine-tuning methods, while effective in enhancing specific model capabilities~\citep{devlin2018bert, thirunavukarasu2023large, gururangan2020don}, often fall short in providing the necessary adaptability.
These approaches are computationally intensive, leaving LLMs ill-equipped to handle the dynamic nature of real-world applications.
In-Context Learning (ICL)~\citep{brown2020language} has emerged as a promising alternative, allowing LLMs to adapt to new tasks without additional training by leveraging their inherent capabilities~\citep{team2023gemini,vacareanu2024words, agarwal2024many}.
ICL, while effective, relies on carefully crafted demonstrations, requires extra overhead for each inference, and interrupts the context, potentially limiting its efficiency and flexibility~\citep{lu2021fantastically,zhao2021calibrate,dong2022survey, liu2023context}.

We envision a next step in LLM adaptation: \emph{Can we elicit and harness the potential of LLMs' inherent capabilities when adapting to new tasks, as demonstrated by ICL, while simultaneously maintaining efficiency and flexibility?}

Our research explores this question by introducing a novel approach \alg inspired by the concept of modularization~\citep{pfeiffer2023modular, fedus2022review,ding2023parameter,zhang2023emergent, xiao2024configurable}.
\alg involves the establishment of a comprehensive library of task vectors, each eliciting one in-context capability within the LLM.
For any arbitrary query text, \alg dynamically leverage this capability library to selectively elicit the capability, effectively and flexibly tapping into the model's inherent capabilities on an as-needed basis.
We empirically verify the advantanges of \alg under $20$ tasks and $4$ models:

\begin{itemize}[nosep, leftmargin=12pt]
    \item \textbf{Efficient Capability Elicitation}: \alg aims to improve the model's task-specific capabilities with minimal additional computational cost during inference.
          Across $20$ tasks and $4$ models, \alg achieves an average improvement of $11.4$\% over zero-shot performance while maintaining the same token usage (Section~\ref{sec:effectiveness}).
    \item \textbf{Flexible Task Handling}: \alg can adapt to various tasks without requiring task-specific demonstrations or rigid templates, enhancing performance on both in-domain and unseen tasks (Sections~\ref{sec:effectiveness} and \ref{sec:generalizable}).
    \item \textbf{Selective Capability Activation}: \alg allows for targeted activation of specific model capabilities based on the input query.
          In our experiments with a math-only capability library, \alg boosted Math performance dramatically (e.g., from $2.6$\% to $21.3$\% for Mistral) while maintaining or slightly improving performance in other domains (Section~\ref{sec:adaptive_activation}).
    \item \textbf{Complementary Integration}: \alg shows potential for complementary use with existing methods, offering further performance gains.
          When combined with BM25 retrieval, \alg enhanced Pythia's average performance from $22.1$\% to $28.3$\% (Section~\ref{sec:orthogonal}).
\end{itemize}

\textbf{Our key contributions are summarized as follows:}
\begin{itemize}[nosep, leftmargin=12pt]
    \item We introduce a novel, modular framework for enhancing the adaptive capabilities of LLMs on demand with minimal computational overhead.
    \item We conduct extensive experiments to evaluate our method, showcasing its effectiveness across different query formats, language models, and tasks.
    \item We provide a thorough analysis of our method, offering insights into the design choices and their contributions to overall performance.
\end{itemize}

\vspace{-10pt}
\section{Related Work}
\vspace{-5pt}
\paragraph*{In-Context Learning.}
While \cite{brown2020language} introduced In-Context Learning (ICL) as a simple yet effective way to enhance LLM performance by incorporating demonstrations into prompts, its applications have rapidly expanded across diverse domains.
ICL enables model to adapt to a wide array of tasks ranging from traditional NLP benchmarks to more specialized tasks such as egression \citep{vacareanu2024words}, kNN classification \citep{agarwal2024many,dinh2022lift}, and even jailbreaking \citep{anil2024many}.
Researchers have actively explored various avenues to further enhance ICL's adaptability and effectiveness. These efforts include increasing demonstration quantity \citep{bertsch2024context,agarwal2024many,zhang2023sentiment,team2023gemini}, fine-tuning models for ICL \citep{min2021metaicl}, leveraging prompt engineering \citep{nie2022improving}, and implenmenting demonstration retrieval \citep{liu2021makes, rubin2021learning, li2023unified,shi2022xricl}.

Concurrently, deeper insights into ICL's underlying mechanisms have been sought through diverse perspectives. Some researchers view ICL as a process of compressing training data into task-specific vectors \citep{hendel2023context}, while others relate it to gradient descent \citep{von2023transformers} or analyze it through the lens of repetition \citep{yan2023understanding} and memorization \citep{golchin2024memorization}.
Building upon ICL advancements, we explored eliciting and harnessing LLMs' inherent capabilities for new task adaptation, akin to ICL, while maintaining efficiency and flexibility.

\paragraph*{Task representation for ICL.}
Inspired by findings that intermediate representations in LLMs encode semantic meaning~\citep{zou2023representation}, researchers have explored injecting in-context learning demonstrations, encoded as function vectors, into intermediate representations to trigger desired predictions~\citep{liu2023context,hendel2023context,todd2023function,li2024context}.
The scope of this research line has broadened to include different modalities,with recent work demonstrating its effectiveness in both visual~\citep{hojel2025finding} and multimodal domains~\citep{huang2024multimodal}.
However, this line of work focuses on manipulating internal representations.
We are the first to comprehensively explore the modular approach of externally storing and retrieving such task representations to augment large language model capabilities.
\looseness=-1

\paragraph*{Modular LLM.}
Examining and understanding the modular nature of large language models (LLMs) has become a crucial area of study for researchers~\citep{pfeiffer2023modular,fedus2022review,ding2023parameter,zhang2023emergent,xiao2024configurable}. Initial investigations suggest that LLMs possess the capability to be broken down into distinct specialized components or modules.
Some approaches introduce additional modules or parameters for optimization, including parameter-efficient tuning techniques like adapter layers~\citep{houlsby2019parameter,pfeiffer2020mad}, prompt tuning~\citep{liu2023pre,ding2021openprompt}, and parameter subset optimization methods such as BitFit~\citep{zaken2021bitfit} and binary masks~\citep{guo2020parameter,zhao2022consistent}.
Other approaches involve training dedicated models for task composition~\citep{shao2023compositional,mu2024learning} or merging fine-tuned parameter adjustments~\citep{ilharco2022editing,panigrahi2023task,merullo2023language,yu2024language}.
Inspired by such modular perspectives, we explore the question of using task vectors in a modular way to dynamically elicit capabilities within the model.

\section{Method}
To elicit the hidden capability inside LLMs, we build our \alg by introducing a capability library which condenses each in-context learned capability into a task vector, and utilizing a retrieval module to strengthen the model when a task vector is helpful.

This section describes our implementation of \alg.
We first formally define in-context learning Task Vectors~(Section \ref{sec:background}), and motivate our work.
Then, we discuss the design choices of building capabilities libraries~(Section \ref{sec_construct_library}), including the layer selection and intervention strategies.
Finally, we introduce our retrieval module~(Section \ref{sec:retrieval}) to dynamic elicit and leverage model's capability.

\subsection{From ICL to Task Vectors: Formal Definitions}
\label{sec:background}
\paragraph*{In-Context Learning (ICL).}
Firstly, we define the framework for ICL.
Let $\cT$ represent a collection of tasks.
For each task $t \in \cT$, there exists a dataset $\mP_t$ of in-context prompts.
Each prompt $\pp_i^t \in \mP_t$ is a sequence of tokens that represents the $i$-th prompt for task $t$.
Specifically, each prompt $\pp_i^t$ consists of two components: (1) a set of $N$ input-output demonstrations $\mD = {(\xx_{ij}, \yy_{ij})}_{j=1}^N$ from task $t$, where $j$ indexes the sequence of pairs ranging from $1$ to $N$, and (2) a query input $\xx_{iq}$, which is distinct from the inputs in $\mD$.
We formally represent an ICL prompt $\pp_i^t$ as:
\begin{align}
    \label{icl_prompt}
    \pp_i^t = [(\xx_{i1}, \yy_{i1}), \ldots, (\xx_{iN}, \yy_{iN}), \xx_{iq}] \,.
\end{align}

The Language Model (LM) aims to predict the corresponding target response $\yy_{iq}$ for the query input $\xx_{iq}$.
Through learning from the demonstrated input-output mappings in $\mD$, ICL can enhances the model's capability to perform this task.
We firstly introduce the hidden state in Transformers below.

\paragraph*{Task Vector.}
Previous research \citep{hendel2023context} introduced the concept of a task vector in the context of ICL.
We build upon this foundation in our work.
We first introduce the definition of hidden state representations in transformer models and task vector is derived from it.
\begin{definition}[\textbf{Hidden State Representation in Transformers}]
    \label{def_hidden_state}
    Let $T$ be an auto-regressive transformer language model with $L$ layers.
    For each layer $l \in {1, \ldots, L}$, we define $\hh_l \in \R^d$ as the vector representation of the last token at layer $l$.
    The computation of $\hh_l$ follows the recurrence relation \citep{vaswani2017attention}: $\hh_l = \hh_{l-1} + \mm_l + \aa_l$, where $\mm_l$ is the output of a multilayer perceptron at layer $l$, and $\aa_l$ is the projection of the attention output into the hidden state at layer $l$.
\end{definition}
Having established the notion of hidden states in transformer models, we can now formally define the task vector within the ICL framework.
\begin{definition}[\textbf{Task Vector} $\mtheta$]
    \label{def_task_vector}
    ICL functions by learning a task-specific mapping from demonstrations.
    This mapping is represented as a task vector $\mtheta$.
    The task vector is derived from the activation state $\hh_l$ (as defined in Definition \ref{def_hidden_state}) at a specific layer $l$, corresponding to the last token of the prompt.
    This vector subsequently steers the transformer to yield pertinent outputs for given queries.
\end{definition}
The task vector, as defined, encapsulates the essence of the task.
This leads to the following lemma, highlighting its role in simulating ICL behavior.
\begin{lemma}[\textbf{Task Vector for ICL Simulation}]
    \label{def_task_vector_function}
    Given a task vector $\mtheta$ that effectively captures the information from demonstrations in an ICL setting, we can simulate the behavior of regular ICL with only query as follows:
    \begin{align*}
        T[\pp_i^t] \approx f(\mtheta; \xx_{iq}) \,,
    \end{align*}
    where:
    \begin{itemize}[leftmargin=12pt, nosep]
        \item $T[\pp_i^t]$ represents the output of the transformer model given a ICL prompt $\pp_i^t$ defined as \eqref{icl_prompt}.
        \item $f(\mtheta; \xx_q)$ denotes a function that processes the query input $\xx_q$ in a zero-shot manner, guided by the information encoded in the task vector $\mtheta$.
    \end{itemize}
\end{lemma}
\begin{remark} [\textbf{Intervention of Task Vector} $\mtheta$]
    \label{def_intervention}
    The function $f(\mtheta; \xx_q)$ mentioned in Lemma \ref{def_task_vector_function} is an abstract concept expressing that the task vector can be used to influence the model's inference process.
    In practice, $f(\mtheta; \xx_q)$ is implemented through operations on the hidden states $\hh_l$ and the task vector $\mtheta$. Specifically, these operations can take the following forms:
    \begin{enumerate}[leftmargin=12pt, nosep]
        \item \textbf{Replacement}~\citep{hendel2023context}: The task vector $\mtheta$ directly replaces the hidden state $\hh_l$, i.e., $\tilde{\hh}_l = \mtheta$.
        \item \textbf{Linear combination}~\citep{todd2023function}: The task vector $\mtheta$ is combined linearly with the hidden state $\hh_l$, i.e., $\tilde{\hh}_l = \hh_l + \alpha \mtheta$, where $\alpha$ is an adjustable scalar parameter.
    \end{enumerate}
\end{remark}

While previous research has demonstrated the existence and extractability of task vectors, it also has shown the potential for serving a technique to elicit the inherent capabilities when adapting to difference tasks as ICL, while simultaneously maintaing computational efficiency and flexibility.

\begin{figure}[!t]
    \centering
    \includegraphics[width=0.65\textwidth,]{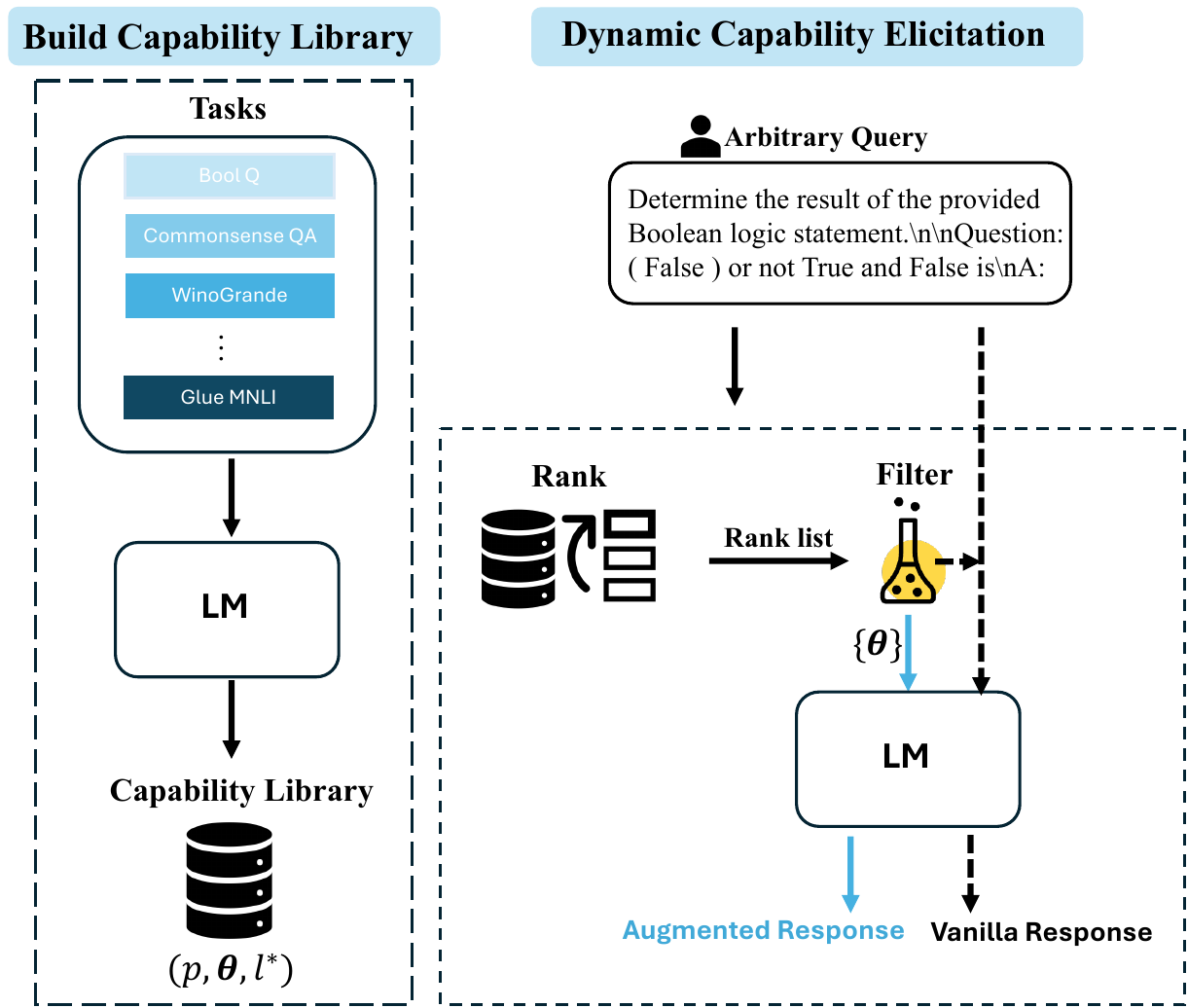}
    \caption{\small
        \textbf{Overview of the proposed \alg framework for Large Language Model Augmentation.}
        \alg consists of two modular components: (1) Build Capability Library - constructing a library of task-specific task vectors by learning from diverse task; (2) Dynamic Capability Elicitation - dynamically retrieving and integrating relevant task vectors from the library to augment the model's capability for an arbitrary input query.
        \looseness=-1
    }
    \label{fig:moativation}
\end{figure}

We investigate the research problem through task vectors by proposing \alg, a framework designed to leverage these vectors for enhancing model capabilities. As shown in Figure~\ref{fig:moativation}, \alg consists of two main components:
\begin{itemize}[leftmargin=12pt ,nosep]
    \item \textbf{Build Capability Library}: A capability library that stores task vectors representing various in-context learned capabilities.
    \item \textbf{Dynamic Capability Elicitation}: A dynamic retrieval module that selectively activates relevant task vectors based on the input query.
\end{itemize}

\subsection{Building Capabilities Library} \label{sec_construct_library}
To investigate the idea of \alg, we first create a library of in-context learned capabilities $\mTheta = \{\{\mtheta _i ^t\}_{i=1} ^ k\}_{t \in \cT}$.
Each element in this library is represented by a task vector (as defined in Definition \ref{def_task_vector}).
Here, $k$ denotes the number of ICL prompts for each task $t$, and we use $k=10$ for illustration.
Definition \ref{def_task_vector} describes $\mtheta \in \R^d$ as a single layer's hidden state.
In our implementation, we collect $\mtheta \in \R^{L \times d}$, which includes representations from all $L$ layers, to enable the exploration of various designs for the sequential components of \alg.

The implementation of creating capability library involves two critical considerations: \circled{1} \emph{Dynamic Layer Selection for $l^*$}, and \circled{2} \emph{Intervention Strategies}.
\circled{1} determines the appropriate layer $l$ to intervene into during further reuse, utilizing the corresponding task vector, while
\circled{2} decides how to appropriately intervene the task vector to influence the model's behavior (possible methods are described in Remark \ref{def_intervention}).
Our framework addresses these considerations as follows.

\textbf{\circled{1} Dynamic Layer Selection for $l^*$.}
The selection of the optimal layer for task vector intervention is crucial for maximizing the effectiveness of our approach~\citep{todd2023function,hendel2023context}. Appendix~\ref{appendix:best_layer_distribution} further illustrates the variation in the optimal layer across different tasks.
We implement a dynamic layer selection strategy to determine the optimal layer $l^*$ for task vector intervention.
While using a validation set to identify the optimal layer is not a novel concept, \emph{our contribution lies in addressing the challenge of determining the intervention layer when applying the library in our proposed pipeline:
    we equip each task vector with its corresponding optimal layer, pre-identified during the library construction phase, thereby enabling efficient and effective reuse of task vectors during inference}.
Our process is as follows:

\begin{itemize}[leftmargin=12pt, nosep]
    \item We obtain the task vector $\mtheta \in \R^{L \times d}$ by processing the ICL prompt $\pp_i^t$ (defined in \eqref{icl_prompt}, using randomly selected $N=16$ demonstrations).
          This is done by extracting the hidden states $\{\hh_l\}_{l=1}^L$ from all layers of the language model.
          We also store these ICL prompts for future use.
    \item Using the validation set, we perform a layer-by-layer analysis:
          \begin{itemize}[leftmargin=12pt, nosep]
              \item For each layer $l$, we intervene $\mtheta_l$ during zero-shot query processing.
              \item We measure zero-shot accuracy on the validation set for each intervention.
              \item We identify $l^*$ as the layer yielding the best accuracy.
          \end{itemize}
    \item We repeat this process for each task-model combination, creating a library where each task vector $\mtheta$ has its corresponding best layer $l^*$.
\end{itemize}
When reusing task vectors in library, for any $\mtheta \in \mTheta$, we intervene at the pre-identified optimal layer $l^*$ for each task-model combination.
This dynamic selection method ensures the performance of the task vector and provides a generalizable framework adaptable to different tasks.

\textbf{\circled{2} Intervention Strategies.}
The concept of intervention, formally introduced in Remark \ref{def_intervention}, also outlines two methods for incorporating the task vector $\mtheta_l$ into the model's inference process.
We evaluate these two intervention strategies:
1) linear combination of the original hidden state and task vector with varying intervene strength $\alpha$, and 2) direct replacement of the original hidden state with the task vector.
We examine the impact of these intervention strategies on both task performance and language modeling capability, with the latter measured using cross-entropy loss on the pre-training dataset (i.e, WikiText).
\looseness=-1

\begin{figure}[t]
    \centering
    \includegraphics[width=1.0\textwidth,]{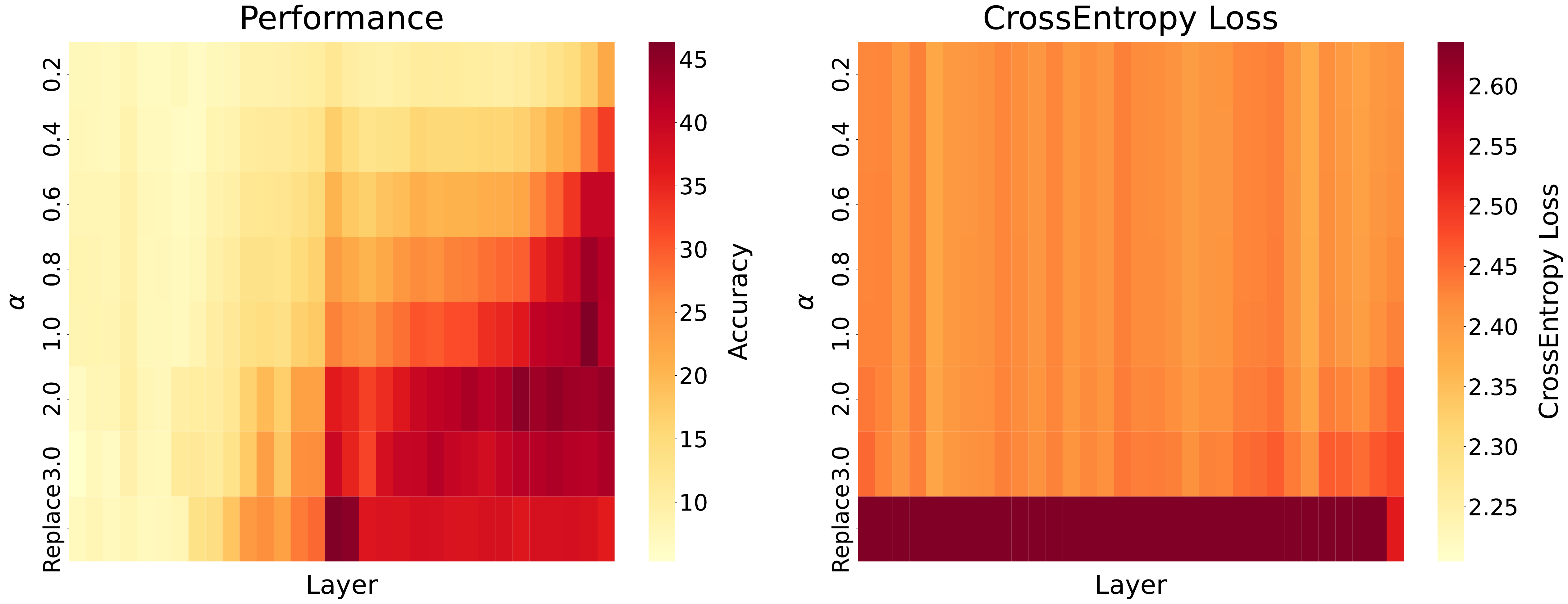}
    \vspace{-1em}
    \caption{\small
        \textbf{Varying intervention strengths affect accuracy and cross-entropy loss in Llama3-8B on valid set of $20$ tasks across different layer}.
        Higher in intervention strengths improve average task performance across layers but negatively impact language modeling capabilities.
        This reveals a trade-off between task-specific enhancement and general language modeling proficiency using task vectors.
    }
    \vspace{-1.2em}
    \label{fig:sweep_alpha_layer}
\end{figure}

Figure~\ref{fig:sweep_alpha_layer} provides a detailed visualization of how varying $\alpha$ affect both accuracy and cross-entropy loss in the Llama3-8B model across a diverse set of $20$ tasks.
\emph{Results reveal a clear trade-off between task performance and language modeling capability as intervention strength increases.}
Among the strategies tested, the additive approach $\tilde{\hh}_l = \hh_l + 2 \times \mtheta_l$ consistently demonstrates superior performance across a wide range of tasks while minimizing degradation in language modeling ability.
Results for other models are presented in the Appendix~\ref{appendix:intervention_strategy}, showing similar trends.

\textbf{In conclusion.}
Our library contains $k \times \abs{\cT}$ items for each model, each consisting of three key components: (1) the ICL prompt $\pp_i^t$, (2) the corresponding task vector $\mtheta \in \R^{L \times d}$, and (3) the pre-identified optimal layer $l^*$.

\subsection{Dynamic Capability Elicitation}
\label{sec:retrieval}
After the creation of the capability library, as described in Section~\ref{sec_construct_library}, we consider two considerations: \circled{i} \emph{Relevant Task Vector Selection}, and \circled{ii} \emph{Threshold-Based Filtering}.
\circled{i} Relevant Task Vector Selection focuses on identifying the most relevant task vectors from the library for a given test query $q$.
We aim to find the most relevant task vectors $\mtheta^q \in \mTheta$ stored in the library.
Unlike traditional in-context learning (ICL), we lack meta-information about the query.
\circled{ii} Threshold-Based Filtering determines whether to utilize a retrieved task vector or not, to avoid compromising performance when no suitable task vectors are available in the library.
Our framework addresses these challenges as follows:
\looseness=-1

\circled{i} \textbf{Relevant Task Vector Selection.}
We address the challenge of selecting the most relevant task vectors by employing a binary classifier to calculate similarity scores.
This classifier is built upon the SimCSE RoBERTa model\footnote{princeton-nlp/sup-simcse-roberta-base}, augmented with a $2$-layer Multi-Layer Perceptron (MLP) head.
The architecture incorporates ReLU activation functions and a dropout rate of $0.2$ for regularization.

We fine-tuned this model over 15 epochs using a learning rate of $2e{-}5$ on our curated dataset of $10{,}000$ examples.
The trained classifier is then used to compute similarity scores between a given query and each ICL prompt in our library.
These scores are used to rank all library items, producing a similarity list of size $k \times \abs{ \cT }$.
The top-ranked task vector from this list is selected as our target for further processing.

\begin{figure}[t]
    \centering
    \begin{subfigure}[b]{0.48\textwidth}
        \centering
        \includegraphics[width=\textwidth]{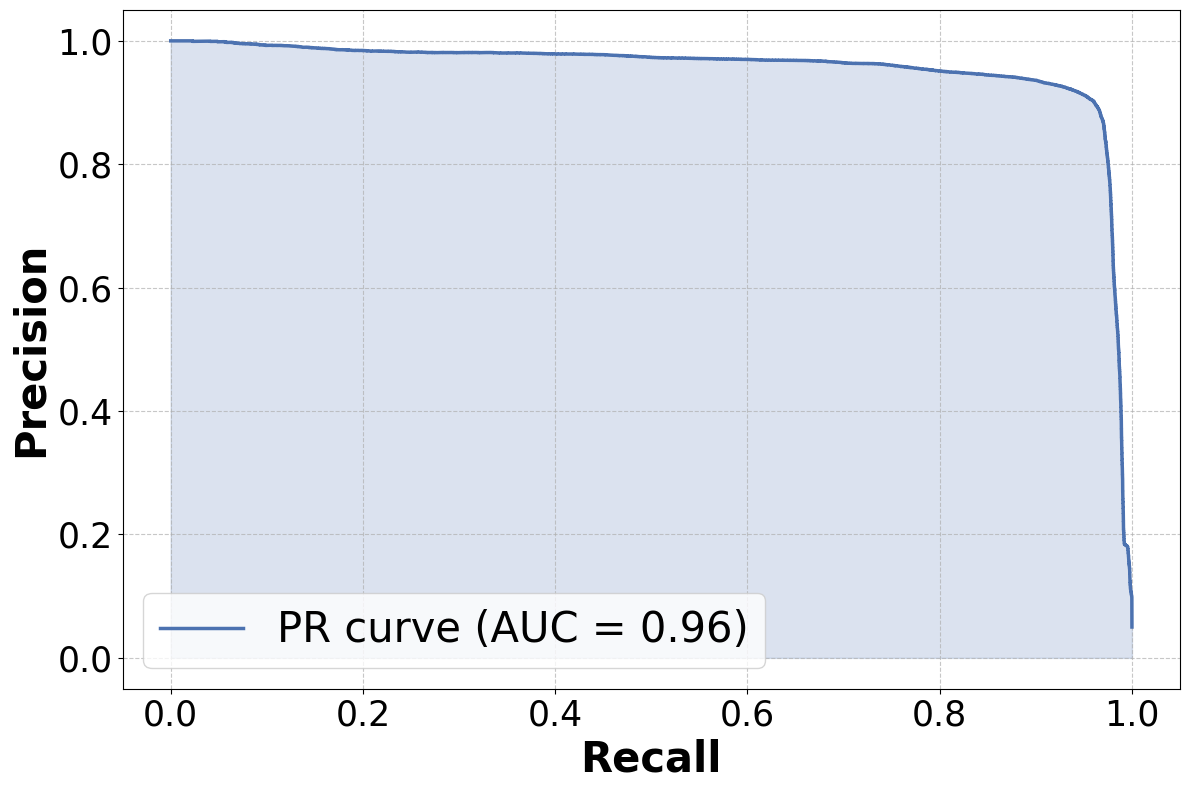}
        \caption{Precision-Recall Curves}
        \label{fig:text_recall_precision}
    \end{subfigure}
    \hfill
    \begin{subfigure}[b]{0.48\textwidth}
        \centering
        \includegraphics[width=\textwidth]{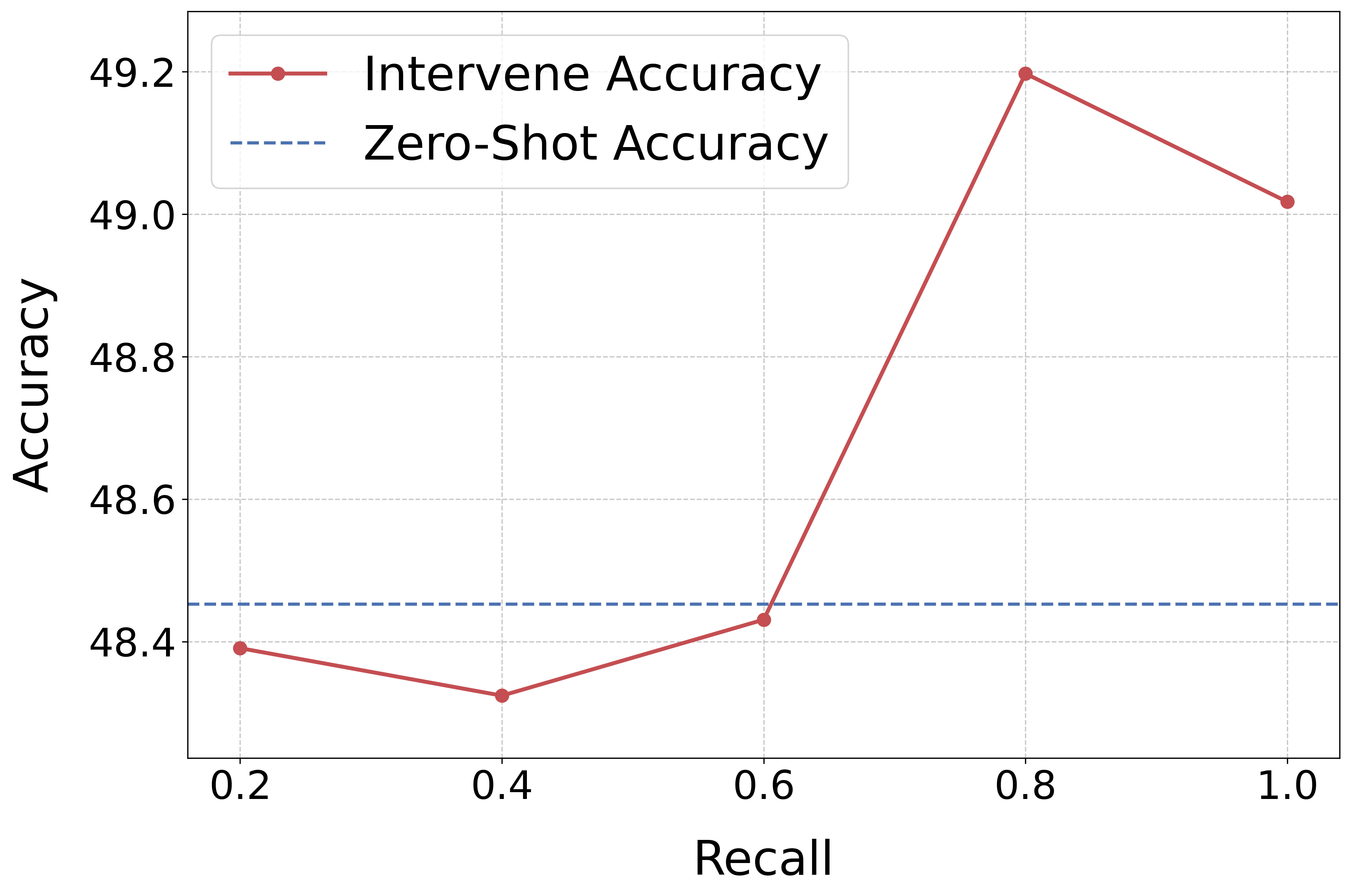}
        \caption{Valid Performance varying different recall.}
        \label{fig:text_recall_sweep}
    \end{subfigure}
    \vspace{-0.8em}
    \caption{\small \textbf{Precision-Recall Curves and recall sweeping on Llama3-8B in valid set across 20 tasks.}
        (a)  Precision-Recall curves for the retriever across 20 tasks (AUC=0.96), guiding threshold selection for high recall and precision.\
        (b) Validation set accuracy after intervention using different recall thresholds.}
    \label{fig:recall_sweep}
    \vspace{-1.5em}
\end{figure}

\circled{ii} \textbf{Threshold-Based Filtering.}
To determine whether to utilize stored task vectors from our library, we implement a threshold-based approach using similarity scores.
This threshold is established through a comprehensive analysis of the recall-precision trade-off across our validation set, as illustrated in Figures~\ref{fig:text_recall_precision}, utilizing the aggregated similarity lists for all samples.
The AUC scores of our precision-recall curves (i.e., $0.96$) demonstrate the high effectiveness of our threshold-based approach in accurately determining whether stored task vectors require intervention.
Our evaluation of various recall levels, as shown in Figures~\ref{fig:text_recall_sweep}, reveals that a recall of $0.8$ provides the optimal balance for our pipeline, other models' results shown in Appendix~\ref{appendix:recall_swep}.

Our decision process of how to utilize the similarity list and threshold to choose whether to use the stored task vectors and what task vectors to apply is as follows:
\begin{itemize}[leftmargin=12pt,nosep]
    \item We implement Dynamic Top-K Thresholding (DTT). If the highest similarity score exceeds the threshold, we select the top $10$ task vectors from the ranked list for further processing.
    \item We then employ a majority voting mechanism among the optimal layers suggested by these top vectors to determine the final layer for intervention.
    \item In cases where the highest similarity score falls below the threshold, we refrain from using any stored task vector, relying instead on the model's base capabilities.
\end{itemize}

\section{Experiments}
To comprehensively evaluate the effectiveness of \alg, we conduct a series of experiments designed to explore the following key questions:
\begin{itemize}[leftmargin=12pt, nosep]
    \item \emph{Capability Elicitation Efficiency:} Can \alg effectively elicit the model's capabilities without incurring significant additional computational costs?
    \item \emph{Selective Activation:} Is \alg capable of selectively activating relevant capabilities as needed for specific tasks?
    \item \emph{Complementarity:} How well does \alg integrate with and complement existing methods in the field? \looseness=-1
    \item \emph{Generalization:} Can \alg handle novel queries, particularly those that diverge significantly from the task vectors currently stored in the library?
\end{itemize}

\looseness=-1

\subsection{Experiment Setup}

\paragraph*{Model.}
We utilize decoder-only auto-regressive language models (Pythia-2.8B \citep{biderman2023pythia}, LLaMA3-8B \citep{dubey2024llama}, and Mistral-7B \citep{jiang2023mistral}) and recurrent neural network (Mamba-2.8B \citep{gu2023mamba}).
Table \ref{tab:model_info} provides a comprehensive overview of these models, detailing their key characteristics including the number of parameters, layer numbers, and training context window size.
For all models, we use the corresponding \texttt{huggingface} implementations~\citep{wolf2020transformers}.

\begin{table}[t]
    \caption{\small
        \textbf{Models used in this work.}
        We consider decoder-only auto-regressive language models and recurrent neural networks that are capable of ICL.
        For each model, we present the number of parameters, context window during training, and the number of layers $\abs{L}$.
    }
    \vspace{-1em}
    \label{tab:model_info}
    \resizebox{\textwidth}{!}{
        \begin{tabular}{llllll}
            \toprule
            \textbf{Model} & \textbf{HuggingFace ID}        & \textbf{Citation}         & \textbf{Parameters} & \textbf{Train Length} & \textbf{$|L|$} \\
            \midrule
            Llama 3        & meta-llama/Meta-Llama-3-8B     & \cite{dubey2024llama}     & 8B                  & 8k                    & 32             \\
            Mistral        & TIGER-Lab/Mistral-7B-Base-V0.2 & \cite{jiang2023mistral}   & 7B                  & 32k                   & 32             \\
            Pythia         & EleutherAI/pythia-2.8b         & \cite{biderman2023pythia} & 2.8B                & 2k                    & 32             \\
            \midrule
            Mamba          & state-spaces/mamba-2.8b-hf     & \cite{gu2023mamba}        & 2.8B                & 2k                    & 64             \\
            \bottomrule
        \end{tabular}
    }
    \vspace{-1.5em}
\end{table}
\vspace{-8pt}
\paragraph*{Tasks.}
To assess the efficacy of our proposed pipeline across a diverse array of scenarios, we have meticulously constructed a benchmark comprising $20$ distinct tasks.
This benchmark is designed to evaluate the model's performance on both classification and multiple-choice problems, spanning a wide spectrum of applications and complexities.
The tasks are categorized into five domains:
\begin{itemize}[leftmargin=12pt, nosep]
    \item \textbf{Knowledge}: CommonsenseQA~\citep{talmor2018commonsenseqa}, OpenBookQA~\citep{mihaylov2018can}, HellaSwag~\citep{zellers2019hellaswag}, and BoolQ~\citep{clark2019boolq};
    \item \textbf{Reasoning}: Four subsets from Big-Bench Hard (BBH)~\citep{suzgun2022challenging} and ARC-Challenge~\citep{clark2018think};
    \item \textbf{Mathematics}: MathQA~\citep{amini2019mathqa} and MMLU Pro-MATH~\citep{wang2024mmlu};
    \item \textbf{Safety}: Crows-Pairs~\citep{nangia2020crows}, BBQ-Age~\citep{parrish2021bbq}, Ethics-Commonsense, and Ethics-Justice~\citep{merity2016pointer};
    \item \textbf{Natural Language Understanding (NLU)}: GLUE (SST-2, QNLI, MNLI)~\citep{wang2018glue} and SuperGLUE (WIC, RTE)~\citep{wang2019superglue}.
\end{itemize}
\vspace{-8pt}
\paragraph*{Evaluation.} \label{para:evaluation}
To evaluate \alg under a real usage scenario, where the demonstrations can hardly be at the same format with the test query, we augment the test query with two additional formats different from the demonstration in library.
Furthermore, in our preliminary experiments, we find that zero-shot LLMs cannot answer properly with contextual guidance. Thus, to ensure a fair comparison with the zero-shot scenario, we add task templates before the test query. More details and examples can be found in Appendix~\ref{appendix:evaluation_setting}.
\vspace{-8pt}
\paragraph{Baselines.}
Our primary baseline is the zero-shot performance of LLMs, as our method maintains the same token usage.
For reference, we also include in-context learning (ICL) and BM25 \citep{robertson2009probabilistic} retrieval of 16 examples from the same pool of examples used in constructing the capability library.
However, these are not directly comparable to our method, due to the raised nearly 20 times more tokens consuming.
The ICL baseline is task-specific, requiring knowledge of each query's task type to use corresponding demonstrations.
In contrast, our method is task-agnostic, applicable across various tasks without needing task-specific information or prompts.

\subsection{Efficient Capability Elicitation} \label{sec:effectiveness}
\begin{table}[t]
    \caption{
        \small
        \textbf{Performance of \alg across model and tasks.}
        \alg significantly enhances performance while maintaining the same token usage as Zero-shot, often achieving results comparable to or better than 16-shot and 16-shot BM25 ICL retriever methods.
        This improvement is consistent across various models and tasks, demonstrating \alg's efficiency and effectiveness in boosting model capabilities without increasing computational demands.
        We sample 100 examples per task across three random seeds.
    }
    \label{tab:main_results}
    \resizebox{\columnwidth}{!}{%
        \begin{tabular}{cc|c|ccccc|c}
            \toprule
            {\color[HTML]{0D0D0D} \textbf{Model}} & \multicolumn{1}{l|}{}          & \textbf{\# Tokens}                   & \textbf{NLU}                      & \textbf{Reasoning}                & \textbf{Knowledge}                & \textbf{Math}                     & \textbf{Safety}                   & \textbf{Avg.}                              \\ \midrule
                                                  & {\color[HTML]{999999} 16-shot} & {\color[HTML]{999999} 1883.8 ± 0.9}  & {\color[HTML]{999999} 60.6 ± 1.0} & {\color[HTML]{999999} 56.0 ± 0.4} & {\color[HTML]{999999} 70.6 ± 1.0} & {\color[HTML]{999999} 26.7 ± 2.0} & {\color[HTML]{999999} 62.1 ± 0.4} & {\color[HTML]{999999} 55.2 ± 0.4}          \\
                                                  & {\color[HTML]{999999} bm25}    & {\color[HTML]{999999} 2350.7 ± 24.9} & {\color[HTML]{999999} 56.1 ± 1.5} & {\color[HTML]{999999} 68.8 ± 0.2} & {\color[HTML]{999999} 69.5 ± 0.9} & {\color[HTML]{999999} 28.0 ± 2.3} & {\color[HTML]{999999} 56.7 ± 2.0} & {\color[HTML]{999999} 55.8 ± 0.7}          \\
                                                  & Zero-shot                      & 108.3 ± 1.4                          & 32.2 ± 1.2                        & 31.6 ± 0.2                        & 42.5 ± 1.2                        & 14.0 ± 1.0                        & 35.5 ± 1.2                        & 31.2 ± 0.7                                 \\
            \multirow{-4}{*}{\textbf{Llama3}}     & \alg                           & 108.3 ± 1.4                          & \textbf{41.6 ± 0.4}               & \textbf{46.7 ± 0.1}               & \textbf{60.6 ± 1.4}               & \textbf{19.1 ± 1.4}               & \textbf{49.9 ± 2.1}               & \textbf{43.5 ± 0.8}                        \\ \midrule
                                                  & {\color[HTML]{999999} 16-shot} & {\color[HTML]{999999} 2161.3 ± 0.9}  & {\color[HTML]{999999} 55.3 ± 0.5} & {\color[HTML]{999999} 52.1 ± 0.5} & {\color[HTML]{999999} 70.8 ± 0.4} & {\color[HTML]{999999} 23.7 ± 1.7} & {\color[HTML]{999999} 63.1 ± 0.6} & {\color[HTML]{999999} 53.0 ± 0.1}          \\
                                                  & {\color[HTML]{999999} bm25}    & {\color[HTML]{999999} 2655.2 ± 27.3} & {\color[HTML]{999999} 55.2 ± 0.3} & {\color[HTML]{999999} 66.0 ± 0.5} & {\color[HTML]{999999} 70.2 ± 1.9} & {\color[HTML]{999999} 24.1 ± 0.4} & {\color[HTML]{999999} 62.1 ± 0.5} & {\color[HTML]{999999} 55.5 ± 0.4}          \\
                                                  & Zero-shot                      & 123.5 ± 1.7                          & 29.6 ± 1.2                        & 26.9 ± 0.4                        & 45.5 ± 1.3                        & 2.8 ± 0.1                         & 36.1 ± 0.3                        & 28.2 ± 0.5                                 \\
            \multirow{-4}{*}{\textbf{Mistral}}    & \alg                           & 123.5 ± 1.7                          & \textbf{41.9 ± 1.0}               & \textbf{48.3 ± 0.3}               & \textbf{59.4 ± 0.9}               & \textbf{20.3 ± 0.9}               & \textbf{48.7 ± 1.8}               & \textbf{43.7 ± 0.6}                        \\ \midrule
                                                  & {\color[HTML]{B7B7B7} 16-shot} & {\color[HTML]{B7B7B7} 1942.4 ± 0.9}  & {\color[HTML]{B7B7B7} 50.2 ± 0.5} & {\color[HTML]{B7B7B7} 19.6 ± 0.1} & {\color[HTML]{B7B7B7} 12.8 ± 0.9} & {\color[HTML]{B7B7B7} 9.2 ± 1.6}  & {\color[HTML]{B7B7B7} 31.8 ± 0.9} & {\color[HTML]{B7B7B7} 24.7 ± 0.2}          \\
                                                  & bm25                           & {\color[HTML]{B7B7B7} 2422.8 ± 26.0} & {\color[HTML]{B7B7B7} 33.3 ± 2.2} & {\color[HTML]{B7B7B7} 25.8 ± 0.4} & {\color[HTML]{B7B7B7} 12.9 ± 0.5} & {\color[HTML]{B7B7B7} 11.0 ± 1.8} & {\color[HTML]{B7B7B7} 27.3 ± 2.1} & {\color[HTML]{B7B7B7} \textbf{22.1 ± 0.5}} \\
                                                  & Zero-shot                      & 110.0 ± 1.5                          & 43.0 ± 0.4                        & 18.3 ± 0.3                        & 22.0 ± 1.5                        & 7.3 ± 0.1                         & 32.5 ± 1.2                        & 24.6 ± 0.4                                 \\
            \multirow{-4}{*}{\textbf{Pythia}}     & \alg                           & 110.0 ± 1.5                          & \textbf{64.0 ± 1.6}               & \textbf{23.6 ± 1.1}               & \textbf{20.4 ± 1.4}               & \textbf{14.5 ± 1.0}               & \textbf{41.2 ± 2.5}               & \textbf{32.7 ± 0.5}                        \\ \midrule
                                                  & {\color[HTML]{B7B7B7} 16-shot} & {\color[HTML]{B7B7B7} 1942.4 ± 0.9}  & {\color[HTML]{B7B7B7} 37.5 ± 1.0} & {\color[HTML]{B7B7B7} 31.5 ± 0.5} & {\color[HTML]{B7B7B7} 31.6 ± 0.8} & {\color[HTML]{B7B7B7} 14.2 ± 0.5} & {\color[HTML]{B7B7B7} 41.7 ± 1.2} & {\color[HTML]{B7B7B7} 31.3 ± 0.3}          \\
                                                  & {\color[HTML]{B7B7B7} bm25}    & {\color[HTML]{B7B7B7} 2422.8 ± 26.0} & {\color[HTML]{B7B7B7} 29.3 ± 2.2} & {\color[HTML]{B7B7B7} 34.9 ± 0.9} & {\color[HTML]{B7B7B7} 24.7 ± 0.5} & {\color[HTML]{B7B7B7} 15.1 ± 2.2} & {\color[HTML]{B7B7B7} 35.4 ± 1.2} & {\color[HTML]{B7B7B7} 27.9 ± 0.3}          \\
                                                  & Zero-shot                      & 110.0 ± 1.5                          & 36.1 ± 1.5                        & 19.3 ± 0.5                        & 17.3 ± 1.2                        & 5.8 ± 1.2                         & 30.1 ± 0.1                        & 21.7 ± 0.2                                 \\
            \multirow{-4}{*}{\textbf{Mamba}}      & \alg                           & 110.0 ± 1.5                          & \textbf{51.1 ± 0.7}               & \textbf{28.7 ± 0.8}               & \textbf{29.2 ± 1.3}               & \textbf{15.3 ± 1.1}               & \textbf{48.2 ± 1.8}               & \textbf{34.5 ± 0.6}                        \\ \bottomrule
        \end{tabular}%
    }
    \vspace{-1.5em}
\end{table}

\paragraph{\emph{\alg achieves efficiently eliciting models' capabilities.}}
\begin{wrapfigure}{r}{0.35\textwidth}
    \centering
    \vspace{-25pt}
    \includegraphics[width=0.9\linewidth]{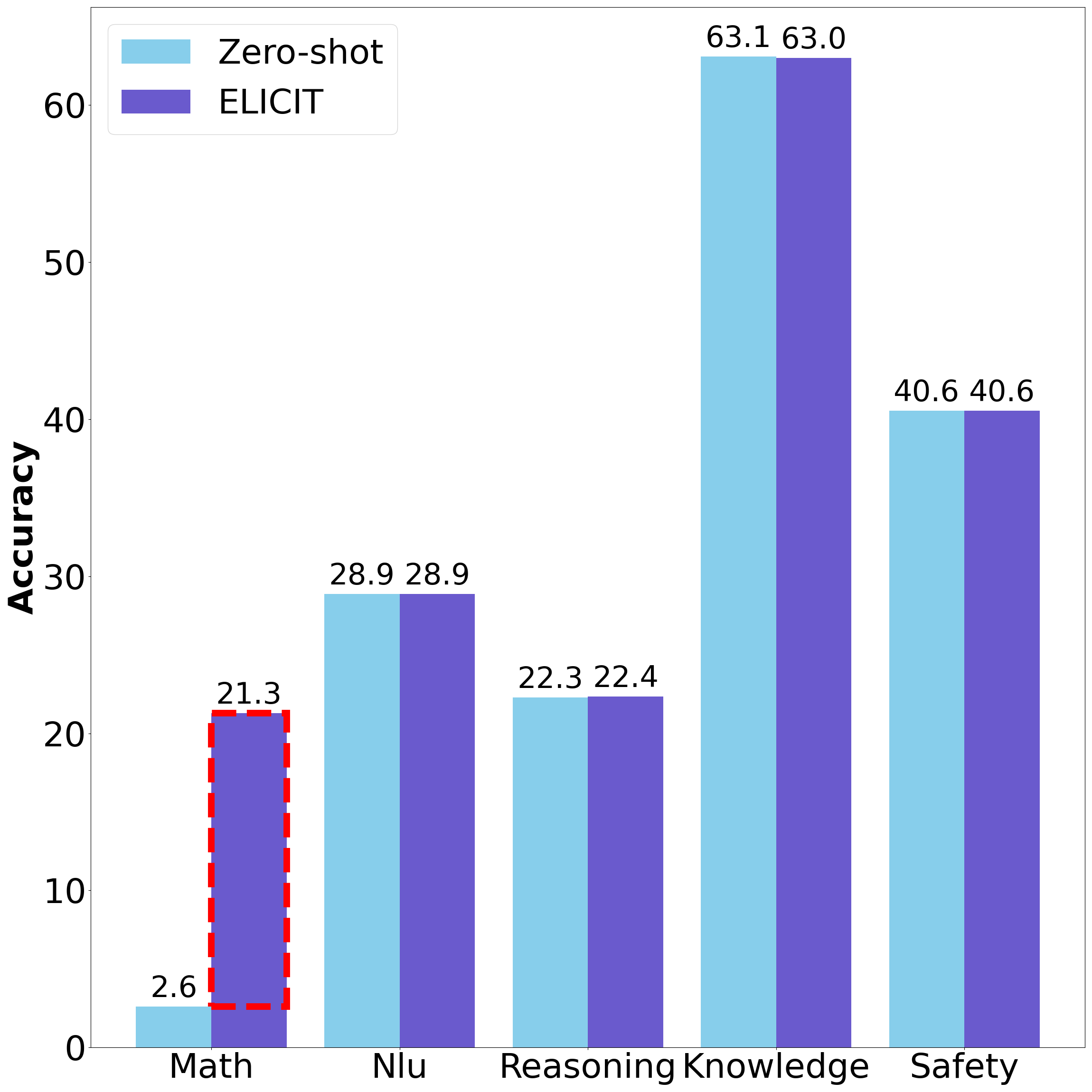}
    \vspace{-10pt}
    \caption{\small \textbf{Performance on \alg across different domains} when the library only contains math-related task vectors on Mistral.}
    \label{fig:selective}
    \vspace{-20pt}
\end{wrapfigure}
From Tables~\ref{tab:main_results}, comparing the zero-shot baseline and \alg, we observe that \alg significantly elicits model capabilities across most tasks without increasing token usage.
Across the 20 tasks, \alg achieves an average improvement of 11.4\% across different models.
For Llama3, \alg improves over zero-shot by 12.3\% while using the same 108.2 tokens.
\alg demonstrates substantial gains in Reasoning (e.g., +15.1\% for LLama3) and Safety tasks (e.g., +14.4\% for LLama3).
In some cases, \alg's performance is comparable to or surpasses that of 16-shot and BM25 methods, despite their higher token requirements.
Furthermore, it exhibits robustness across various template formats, highlighting its versatility.

\vspace{-8pt}
\subsection{Selective Adaptive Activation of Capabilities} \label{sec:adaptive_activation}
\paragraph{\emph{\alg elicits capability when necessary.}}

\setlength{\intextsep}{0pt}
We demonstrate selective activation by constructing a library containing only math-related task vectors, as shown in Figure~\ref{fig:selective}.
The results clearly illustrate that \alg significantly boosts performance in the Math domain, with a dramatic increase from $2.6$\% to $21.3$\%, while maintaining performance in other domains for Mistral.
Notably, the Reasoning domain also shows a slight improvement, increasing from $22.3$\% to $22.4$\%.
This behavior stems from ELICIT's selective application of task vectors from library, which are not applied when no relevant tasks vectors are detected. More discussion is presented in Appendix~\ref{appendix:analysis_selective_activation}.
Results for other models, presented in Appendix~\ref{appendix:adaptie_elicitation}, demonstrate a similar trend.
The striking improvement in Math performance, coupled with the subtle gain in Reasoning and the stability in other domains, demonstrates \alg's capacity for targeted capability activation, making it a flexible and efficient performance enhancer.

\vspace{-8pt}
\subsection{ELICIT generalize to Unseen Tasks without addtional information} \label{sec:generalizable}
\paragraph{\emph{\alg generalizes across unseen tasks.}}
In Table~\ref{tab:unseen}, we observe that \alg significantly improves model performance on unseen tasks (GLUE-COLA, BBQ Religion, Deepmind \citep{saxton2019analysing}, MMLU-Psychology, and BBH-Logical-Deduction-Five-objects) not present in its capability library.
Across all models, \alg consistently outperforms the Zero-shot baseline.
In several cases, it even approaches or surpasses the BM25 retrieval baseline, despite using substantially fewer tokens.
For instance, on the MMLU-Psychology task, \alg achieves a $15.8$\% absolute improvement over Zero-shot for Llama3 Model.
These results are achieved without additional token usage and task information, demonstrating \alg's efficiency, flexibility, and generalization ability.
\looseness=-1
\begin{table}[!t]
    \caption{\small
        \textbf{\alg can generalize to unseen tasks.}
        \alg achieves significant performance gains without additional token usage across different models and unseen tasks.
        We sample 100 examples per task across three random seeds. We use BM25 retrieval of 16 examples as baseline.
    }
    \vspace{-1em}
    \label{tab:unseen}
    \resizebox{\textwidth}{!}{%
        \begin{tabular}{cc|lccccc|c}
            \toprule
            \multicolumn{1}{l}{{\color[HTML]{0D0D0D} \textbf{}}} & \multicolumn{1}{l|}{}                & \textbf{\# Tokens}                   & \multicolumn{1}{l}{\textbf{GLUE COLA}} & \multicolumn{1}{l}{\textbf{BBQ Religion}} & \multicolumn{1}{l}{\textbf{Deepmind}} & \multicolumn{1}{l}{\textbf{MMLU-Psychology}} & \multicolumn{1}{l|}{\textbf{BBH-five-objects}} & \multicolumn{1}{l}{\textbf{Avg}}  \\ \midrule
                                                                 & {\color[HTML]{999999} \textbf{BM25}} & {\color[HTML]{999999} 2502.8 ± 26.0} & {\color[HTML]{999999} 55.4 ± 1.0}      & {\color[HTML]{999999} 64.6 ± 1.3}         & {\color[HTML]{999999} 30.7 ± 1.7}     & {\color[HTML]{999999} 83.0 ± 0.1}            & {\color[HTML]{999999} 48.3 ± 0.0}              & {\color[HTML]{999999} 56.4 ± 0.4} \\
                                                                 & \textbf{Zero-shot}                   & 103.6 ± 47.7                         & \textbf{72.0 ± 0.7}                    & 38.6 ± 1.1                                & 17.5 ± 2.6                            & 54.2 ± 0.3                                   & 17.1 ± 0.0                                     & 39.9 ± 0.8                        \\
            \multirow{-3}{*}{\textbf{Llama}}                     & \textbf{\alg}                        & 103.6 ± 47.7                         & 63.4 ± 0.9                             & \textbf{45.0 ± 0.7}                       & \textbf{23.7 ± 3.4}                   & \textbf{70.0 ± 0.6}                          & \textbf{25.7 ± 0.0}                            & \textbf{45.6 ± 0.4}               \\ \midrule
                                                                 & {\color[HTML]{999999} \textbf{BM25}} & {\color[HTML]{999999} 2804.6 ± 27.6} & {\color[HTML]{999999} 44.4 ± 2.2}      & {\color[HTML]{999999} 70.7 ± 0.7}         & {\color[HTML]{999999} 26.6 ± 3.9}     & {\color[HTML]{999999} 78.7 ± 1.1}            & {\color[HTML]{999999} 25.7 ± 0.0}              & {\color[HTML]{999999} 49.2 ± 0.3} \\
                                                                 & \textbf{Zero-shot}                   & 115.4 ± 51.0                         & \textbf{43.3 ± 1.1}                    & 35.4 ± 3.3                                & 9.0 ± 0.4                             & 57.9 ± 0.7                                   & 7.4 ± 0.0                                      & 30.6 ± 1.0                        \\
            \multirow{-3}{*}{\textbf{Mistral}}                   & \textbf{\alg}                        & 115.4 ± 51.0                         & 41.7 ± 0.8                             & \textbf{42.1 ± 2.5}                       & \textbf{25.1 ± 1.2}                   & \textbf{65.6 ± 0.6}                          & \textbf{15.6 ± 0.0}                            & \textbf{38.0 ± 0.6}               \\ \midrule
                                                                 & {\color[HTML]{999999} \textbf{BM25}} & {\color[HTML]{999999} 2600.0 ± 28.3} & {\color[HTML]{999999} 5.8 ± 1.0}       & {\color[HTML]{999999} 19.1 ± 1.2}         & {\color[HTML]{999999} 14.1 ± 1.2}     & {\color[HTML]{999999} 4.7 ± 0.3}             & {\color[HTML]{999999} 1.0 ± 0.0}               & {\color[HTML]{999999} 8.9 ± 0.3}  \\
                                                                 & \textbf{Zero-shot}                   & 106.7 ± 49.6                         & \textbf{48.5 ± 0.6}                    & 21.7 ± 1.7                                & 9.7 ± 1.2                             & 20.1 ± 0.8                                   & 7.6 ± 0.0                                      & 21.5 ± 0.1                        \\
            \multirow{-3}{*}{\textbf{Pythia}}                    & \textbf{\alg}                        & 106.7 ± 49.6                         & 45.4 ± 0.6                             & \textbf{30.3 ± 4.2}                       & \textbf{14.2 ± 1.8}                   & \textbf{20.4 ± 0.6}                          & \textbf{14.3 ± 0.0}                            & \textbf{24.9 ± 0.6}               \\ \midrule
                                                                 & {\color[HTML]{999999} \textbf{BM25}} & {\color[HTML]{999999} 2600.0 ± 28.3} & {\color[HTML]{999999} 48.1 ± 3.1}      & {\color[HTML]{999999} 30.6 ± 1.1}         & {\color[HTML]{999999} 21.6 ± 3.3}     & {\color[HTML]{999999} 19.1 ± 0.9}            & {\color[HTML]{999999} 25.8 ± 0.0}              & {\color[HTML]{999999} 29.0 ± 0.9} \\
                                                                 & \textbf{Zero-shot}                   & 106.7 ± 49.6                         & \textbf{70.3 ± 1.0}                    & 21.3 ± 2.9                                & 10.9 ± 0.7                            & 13.9 ± 0.5                                   & 6.2 ± 0.0                                      & 24.5 ± 0.4                        \\
            \multirow{-3}{*}{\textbf{Mamba}}                     & \textbf{\alg}                        & 106.7 ± 49.6                         & 63.6 ± 0.4                             & \textbf{31.5 ± 2.5}                       & \textbf{22.1 ± 3.3}                   & \textbf{20.4 ± 0.2}                          & \textbf{14.4 ± 0.0}                            & \textbf{30.4 ± 0.9}               \\ \bottomrule
        \end{tabular}%
    }
    \vspace{-1.5em}
\end{table}

\vspace{-8pt}
\subsection{Complementary Intergration} \label{sec:orthogonal}
\paragraph*{\emph{\alg shows potential as a plug-and-play performance booster.}}
While \alg demonstrates compatibility with existing solutions like BM25 retrieval, Table \ref{tab:bm25+ours_in_domain} reveals nuanced performance patterns.
For smaller models (Pythia-2.8B and Mamba-2.8B), combining ELICIT with BM25 yields consistent improvements, with Pythia's average performance increasing from 22.1\% to 28.3\% (+5.9\%).
However, larger models (Llama3-8B and Mistral-7B) exhibit mixed results: while NLU and Reasoning tasks show modest gains (e.g., +2.6\% for Llama3), Knowledge and Safety tasks experience slight declines.
Aligning with the findings of \cite{li2024context}, this phenomenon can be attributed to two factors: (1) smaller models' relatively weak in-context learning capabilities benefit more from additional task-relevant information provided by our method, while (2) larger models' inherently stronger in-context adaptation abilities may be disrupted by the introduction of additional context that alters their learned representations.
Future work could investigate this scale-dependent phenomenon.

\vspace{10pt}
\begin{table}[!t]
    \caption{\textbf{\alg as a potential plug-and-play performance booster}: performance when combined with BM25 on in-domain tasks.
        Results indicate stronger complementary effects for smaller models (Pythia, Mamba), while larger models (Llama3, Mistral) show task-specific variations.}
    \vspace{-1em}
    \label{tab:bm25+ours_in_domain}
    \resizebox{\columnwidth}{!}{%
        \begin{tabular}{cc|ccccc|c}
            \toprule
            { \textbf{Model}} &                      & \textbf{NLU}        & \textbf{Reasoning}  & \textbf{Knowledge}  & \textbf{Math}       & \textbf{Safety}     & \textbf{Avg.}       \\ \midrule
            \textbf{Llama}    & \textbf{BM25}        & 56.1 ± 1.5          & \textbf{68.8 ± 0.2} & \textbf{69.5 ± 0.9} & \textbf{28.0 ± 2.3} & \textbf{56.7 ± 2.0} & \textbf{55.8 ± 0.7} \\
                              & \textbf{BM25+ELICIT} & \textbf{58.0 ± 0.4} & 62.7 ± 0.5          & 65.1 ± 0.5          & 25.1 ± 1.6          & 54.5 ± 3.5          & 53.1 ± 1.1          \\ \midrule
            \textbf{Mistral}  & \textbf{BM25}        & \textbf{55.2 ± 0.3} & \textbf{66.0 ± 0.5} & \textbf{70.2 ± 1.9} & 24.1 ± 0.4          & \textbf{62.1 ± 0.5} & \textbf{55.5 ± 0.4} \\
                              & \textbf{BM25+ELICIT} & 54.5 ± 0.8          & 62.6 ± 0.4          & 67.5 ± 1.7          & \textbf{24.8 ± 1.9} & 58.0 ± 1.4          & 53.5 ± 0.5          \\ \midrule
            \textbf{Pythia}   & \textbf{BM25}        & 33.3 ± 2.2          & 25.8 ± 0.4          & 12.9 ± 0.5          & 11.0 ± 1.8          & 27.3 ± 2.1          & 22.1 ± 0.5          \\
                              & \textbf{BM25+ELICIT} & \textbf{53.5 ± 1.5} & \textbf{26.5 ± 1.1} & \textbf{14.5 ± 0.6} & \textbf{13.2 ± 1.7} & \textbf{33.7 ± 0.7} & \textbf{28.3 ± 0.3} \\ \midrule
            \textbf{Mamba}    & \textbf{BM25}        & 29.3 ± 2.2          & \textbf{34.9 ± 0.9} & 24.7 ± 0.5          & 15.1 ± 2.2          & 35.4 ± 1.2          & 27.9 ± 0.3          \\
                              & \textbf{BM25+ELICIT} & \textbf{38.4 ± 1.2} & 31.6 ± 0.4          & \textbf{28.9 ± 0.2} & \textbf{15.2 ± 3.8} & \textbf{42.9 ± 1.9} & \textbf{31.4 ± 0.3} \\
            \bottomrule
        \end{tabular}%
    }
    \vspace{-1.5em}
\end{table}

\vspace{-10pt}
\section{Ablation Study}

\subsection{Similarity-Based Retrieve} \label{sec:similarity-based_methods}

\begin{figure}[h]
    \centering
    \subfloat[Cosine Similarity]{%
        \includegraphics[width=0.33\textwidth]{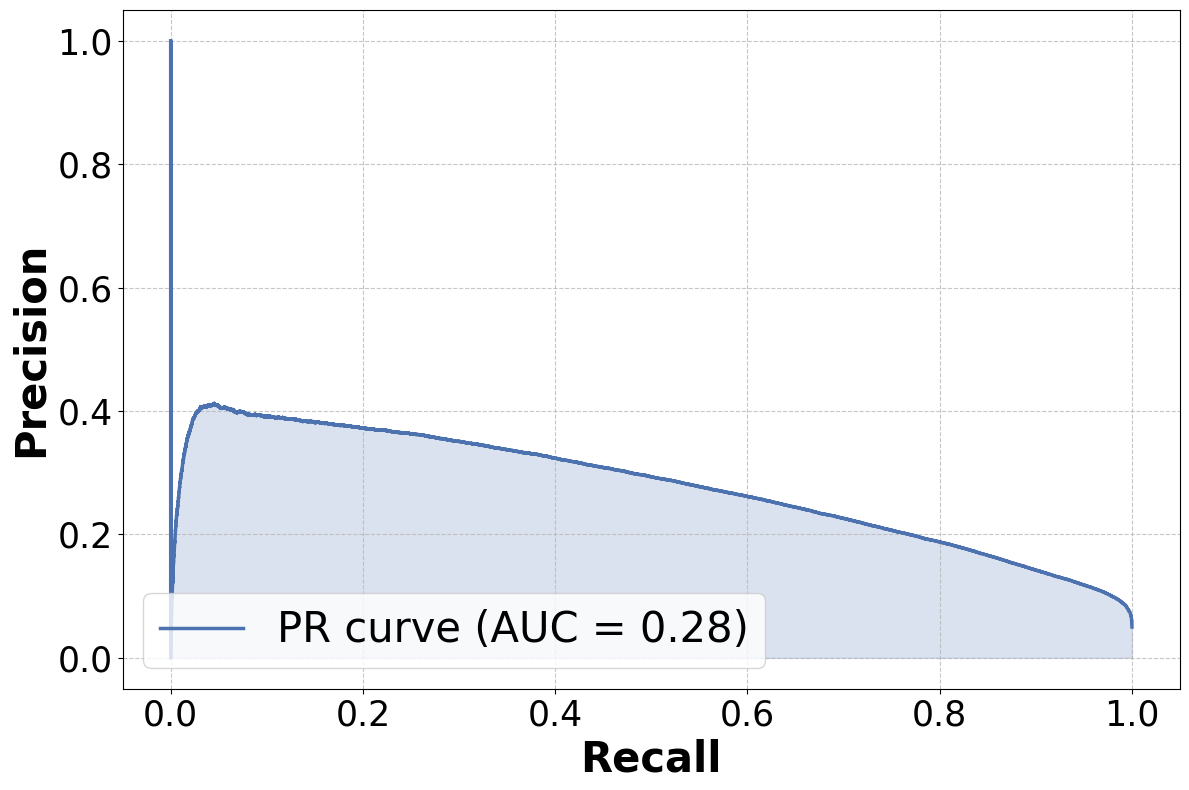}%
        \label{fig:figure4}%
    }%
    \hfill
    \subfloat[Euclidean Distance]{%
        \includegraphics[width=0.33\textwidth]{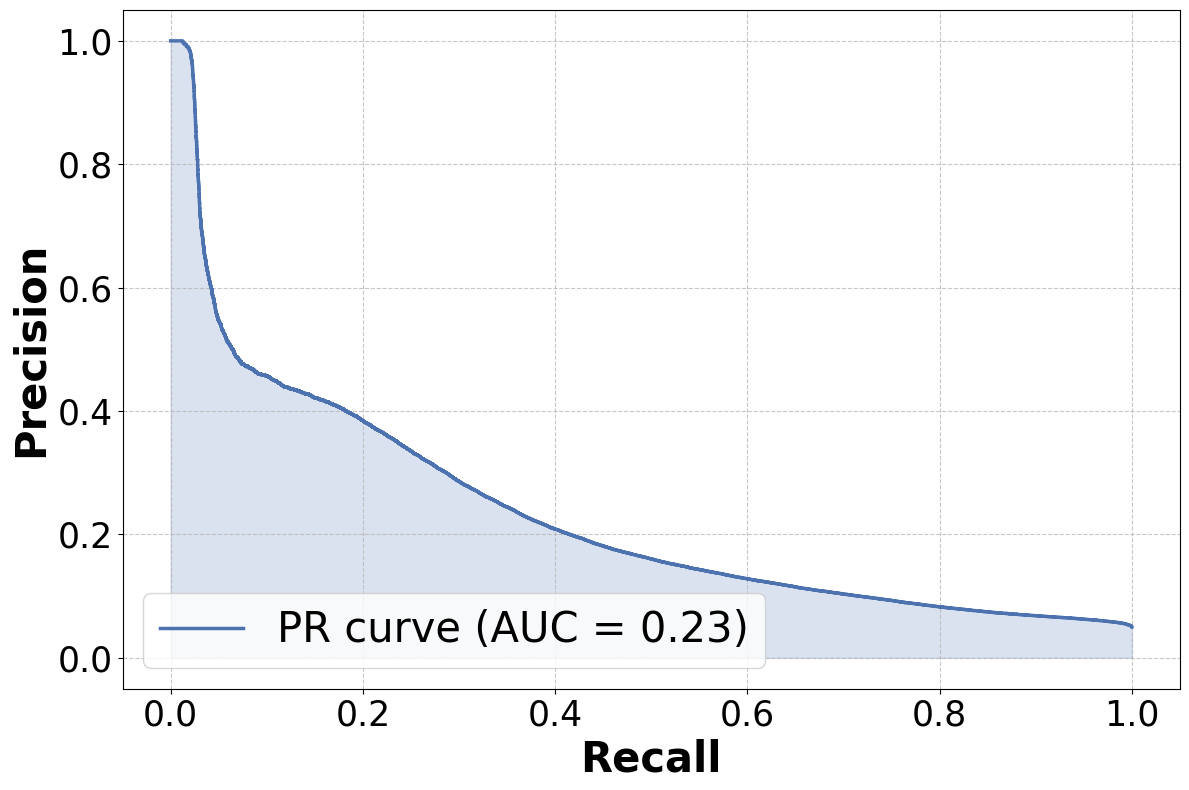}%
        \label{fig:figure5}%
    }%
    \hfill
    \subfloat[t-SNE Distance]{%
        \includegraphics[width=0.33\textwidth]{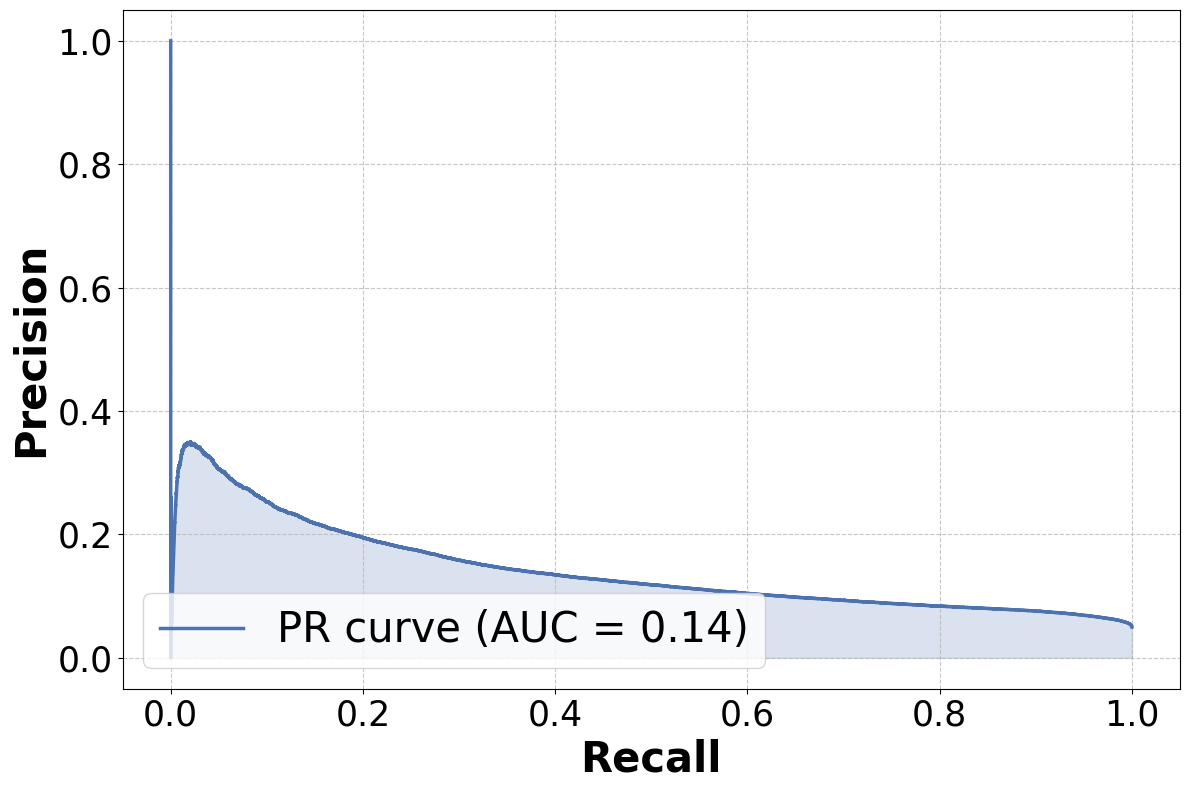}%
        \label{fig:figure6}%
    }
    \vspace{-1em}
    \caption{\small \textbf{Precision-Recall Curves for Similarity-based methods for Llama3}.}
    \label{fig:recall_precision_other_methods}
\end{figure}

We also explored similarity-based retrieval methods, such as cosine similarity, t-SNE distance, and Euclidean distance between the query embedding and the task vectors $\mtheta$ in capability library.
However, as illustrated in Figure~\ref{fig:recall_precision_other_methods}, the precision-recall curves for these methods on Llama3 exhibit very low AUC scores, with the highest being a mere 0.28.
These poor AUC values indicate that the discrimination ability of these similarity-based approaches is inadequate for effectively identifying relevant task vectors from the library.
The precision-recall curves for similarity-based methods on other models are presented in Appendix~\ref{appendix:similarity-based_retrieve_method}, further highlighting their suboptimal performance.
In stark contrast, the trained retriever in our proposed design can achieve a remarkably high AUC of 0.96 (Figure~\ref{fig:text_recall_precision}).
This substantial improvement in retrieval performance underscores the benefits of our design, which effectively leverages learning-based techniques to intelligently retrieve and integrate relevant capabilities from the library.

\subsection{Selection after threshold filtering} \label{sec:ablation_study}
\begin{wraptable}{r}{0.35\textwidth}
    \caption{\small \textbf{Ablation study} on DDT on valid for Llama3-8B.}
    \small
    \label{tab:ablation_study}
    \centering
    \begin{tabular}{@{}lc@{}}
        \toprule
                                & \multicolumn{1}{c@{}}{\textbf{TP}} \\
        \midrule
        \textbf{zs-shot}        & 33.62                              \\
        \midrule
        \textbf{$n$=5}          & 44.71                              \\
        \textbf{$n$=15}         & 45.02                              \\
        \textbf{$n$=10,w/o DTT} & 45.17                              \\
        \midrule
        \textbf{$n$=10}         & \textbf{45.29}                     \\
        \bottomrule
    \end{tabular}%
\end{wraptable}

In Table~\ref{tab:ablation_study}, we investigate the impact of the number of selected states $n$ when the top1 similarity score reaches the threshold, as well as the effect of using Dynamic Top-K Thresholding (DTT) or not as Section \ref{sec:retrieval} mentioned.
The results show that selecting $n=10$ and using Dynamic Top-K Thresholding achieves the best performance on the validation set.
Choosing fewer vector quantities (e.g., n=5) would limit the method's potential, while selecting too many (e.g., n=15) could introduce irrelevant noise, thereby degrading performance.
Dynamic Top-K adaptively sets the similarity threshold to ensure that only sufficiently relevant vectors are utilized.
This ablation study highlights the rationale and effectiveness of our design choices.

\section{Conclusion}
In this paper, we explore the vision of eliciting and harnessing the potential of large language models' inherent capabilities when adapting to new tasks, akin to in-context learning (ICL), while maintaining efficiency and flexibility.
We propose \alg, a novel framework consisting of two key modules: Build Capability Library and Dynamic Capability Elicitation.
\alg achieves consistent improvements across diverse tasks, input formats, and model architectures.
Our results show that \alg not only has the potential to harness models' latent abilities without introducing substantial additional computational cost, but also advances language models' performance, versatility, adaptability, and scalability.

\clearpage
\section*{Acknowledgement}
This work was supported in part by the National Science and Technology Major Project (No.\ 2022ZD0115101), Research Center for Industries of the Future (RCIF) at Westlake University, Westlake Education Foundation, and Westlake University Center for High-performance Computing.

\bibliography{sources/reference}

\begin{thebibliography}{80}
\providecommand{\natexlab}[1]{#1}
\providecommand{\url}[1]{\texttt{#1}}
\expandafter\ifx\csname urlstyle\endcsname\relax
  \providecommand{\doi}[1]{doi: #1}\else
  \providecommand{\doi}{doi: \begingroup \urlstyle{rm}\Url}\fi

\bibitem[Achiam et~al.(2023)Achiam, Adler, Agarwal, Ahmad, Akkaya, Aleman, Almeida, Altenschmidt, Altman, Anadkat, et~al.]{achiam2023gpt}
Josh Achiam, Steven Adler, Sandhini Agarwal, Lama Ahmad, Ilge Akkaya, Florencia~Leoni Aleman, Diogo Almeida, Janko Altenschmidt, Sam Altman, Shyamal Anadkat, et~al.
\newblock Gpt-4 technical report.
\newblock \emph{arXiv preprint arXiv:2303.08774}, 2023.

\bibitem[Agarwal et~al.(2024)Agarwal, Singh, Zhang, Bohnet, Chan, Anand, Abbas, Nova, Co-Reyes, Chu, et~al.]{agarwal2024many}
Rishabh Agarwal, Avi Singh, Lei~M Zhang, Bernd Bohnet, Stephanie Chan, Ankesh Anand, Zaheer Abbas, Azade Nova, John~D Co-Reyes, Eric Chu, et~al.
\newblock Many-shot in-context learning.
\newblock \emph{arXiv preprint arXiv:2404.11018}, 2024.

\bibitem[Amini et~al.(2019)Amini, Gabriel, Lin, Koncel-Kedziorski, Choi, and Hajishirzi]{amini2019mathqa}
Aida Amini, Saadia Gabriel, Peter Lin, Rik Koncel-Kedziorski, Yejin Choi, and Hannaneh Hajishirzi.
\newblock Mathqa: Towards interpretable math word problem solving with operation-based formalisms.
\newblock \emph{arXiv preprint arXiv:1905.13319}, 2019.

\bibitem[Anil et~al.(2024)Anil, Durmus, Sharma, Benton, Kundu, Batson, Rimsky, Tong, Mu, Ford, et~al.]{anil2024many}
Cem Anil, Esin Durmus, Mrinank Sharma, Joe Benton, Sandipan Kundu, Joshua Batson, Nina Rimsky, Meg Tong, Jesse Mu, Daniel Ford, et~al.
\newblock Many-shot jailbreaking.
\newblock \emph{Anthropic, April}, 2024.

\bibitem[Bertsch et~al.(2024)Bertsch, Ivgi, Alon, Berant, Gormley, and Neubig]{bertsch2024context}
Amanda Bertsch, Maor Ivgi, Uri Alon, Jonathan Berant, Matthew~R Gormley, and Graham Neubig.
\newblock In-context learning with long-context models: An in-depth exploration.
\newblock \emph{arXiv preprint arXiv:2405.00200}, 2024.

\bibitem[Biderman et~al.(2023)Biderman, Schoelkopf, Anthony, Bradley, O’Brien, Hallahan, Khan, Purohit, Prashanth, Raff, et~al.]{biderman2023pythia}
Stella Biderman, Hailey Schoelkopf, Quentin~Gregory Anthony, Herbie Bradley, Kyle O’Brien, Eric Hallahan, Mohammad~Aflah Khan, Shivanshu Purohit, USVSN~Sai Prashanth, Edward Raff, et~al.
\newblock Pythia: A suite for analyzing large language models across training and scaling.
\newblock In \emph{International Conference on Machine Learning}, pp.\  2397--2430. PMLR, 2023.

\bibitem[Bommasani et~al.(2021)Bommasani, Hudson, Adeli, Altman, Arora, von Arx, Bernstein, Bohg, Bosselut, Brunskill, et~al.]{bommasani2021opportunities}
Rishi Bommasani, Drew~A Hudson, Ehsan Adeli, Russ Altman, Simran Arora, Sydney von Arx, Michael~S Bernstein, Jeannette Bohg, Antoine Bosselut, Emma Brunskill, et~al.
\newblock On the opportunities and risks of foundation models.
\newblock \emph{arXiv preprint arXiv:2108.07258}, 2021.

\bibitem[Brown(2020)]{brown2020language}
Tom~B Brown.
\newblock Language models are few-shot learners.
\newblock \emph{arXiv preprint arXiv:2005.14165}, 2020.

\bibitem[Chan et~al.(2022)Chan, Santoro, Lampinen, Wang, Singh, Richemond, McClelland, and Hill]{chan2022data}
Stephanie Chan, Adam Santoro, Andrew Lampinen, Jane Wang, Aaditya Singh, Pierre Richemond, James McClelland, and Felix Hill.
\newblock Data distributional properties drive emergent in-context learning in transformers.
\newblock \emph{Advances in Neural Information Processing Systems}, 35:\penalty0 18878--18891, 2022.

\bibitem[Clark et~al.(2019)Clark, Lee, Chang, Kwiatkowski, Collins, and Toutanova]{clark2019boolq}
Christopher Clark, Kenton Lee, Ming-Wei Chang, Tom Kwiatkowski, Michael Collins, and Kristina Toutanova.
\newblock Boolq: Exploring the surprising difficulty of natural yes/no questions.
\newblock \emph{arXiv preprint arXiv:1905.10044}, 2019.

\bibitem[Clark et~al.(2018)Clark, Cowhey, Etzioni, Khot, Sabharwal, Schoenick, and Tafjord]{clark2018think}
Peter Clark, Isaac Cowhey, Oren Etzioni, Tushar Khot, Ashish Sabharwal, Carissa Schoenick, and Oyvind Tafjord.
\newblock Think you have solved question answering? try arc, the ai2 reasoning challenge.
\newblock \emph{arXiv preprint arXiv:1803.05457}, 2018.

\bibitem[Devlin(2018)]{devlin2018bert}
Jacob Devlin.
\newblock Bert: Pre-training of deep bidirectional transformers for language understanding.
\newblock \emph{arXiv preprint arXiv:1810.04805}, 2018.

\bibitem[Ding et~al.(2021)Ding, Hu, Zhao, Chen, Liu, Zheng, and Sun]{ding2021openprompt}
Ning Ding, Shengding Hu, Weilin Zhao, Yulin Chen, Zhiyuan Liu, Hai-Tao Zheng, and Maosong Sun.
\newblock Openprompt: An open-source framework for prompt-learning.
\newblock \emph{arXiv preprint arXiv:2111.01998}, 2021.

\bibitem[Ding et~al.(2023)Ding, Qin, Yang, Wei, Yang, Su, Hu, Chen, Chan, Chen, et~al.]{ding2023parameter}
Ning Ding, Yujia Qin, Guang Yang, Fuchao Wei, Zonghan Yang, Yusheng Su, Shengding Hu, Yulin Chen, Chi-Min Chan, Weize Chen, et~al.
\newblock Parameter-efficient fine-tuning of large-scale pre-trained language models.
\newblock \emph{Nature Machine Intelligence}, 5\penalty0 (3):\penalty0 220--235, 2023.

\bibitem[Dinh et~al.(2022)Dinh, Zeng, Zhang, Lin, Gira, Rajput, Sohn, Papailiopoulos, and Lee]{dinh2022lift}
Tuan Dinh, Yuchen Zeng, Ruisu Zhang, Ziqian Lin, Michael Gira, Shashank Rajput, Jy-yong Sohn, Dimitris Papailiopoulos, and Kangwook Lee.
\newblock Lift: Language-interfaced fine-tuning for non-language machine learning tasks.
\newblock \emph{Advances in Neural Information Processing Systems}, 35:\penalty0 11763--11784, 2022.

\bibitem[Dong et~al.(2022)Dong, Li, Dai, Zheng, Wu, Chang, Sun, Xu, and Sui]{dong2022survey}
Qingxiu Dong, Lei Li, Damai Dai, Ce~Zheng, Zhiyong Wu, Baobao Chang, Xu~Sun, Jingjing Xu, and Zhifang Sui.
\newblock A survey on in-context learning.
\newblock \emph{arXiv preprint arXiv:2301.00234}, 2022.

\bibitem[Dubey et~al.(2024)Dubey, Jauhri, Pandey, Kadian, Al-Dahle, Letman, Mathur, Schelten, Yang, Fan, et~al.]{dubey2024llama}
Abhimanyu Dubey, Abhinav Jauhri, Abhinav Pandey, Abhishek Kadian, Ahmad Al-Dahle, Aiesha Letman, Akhil Mathur, Alan Schelten, Amy Yang, Angela Fan, et~al.
\newblock The llama 3 herd of models.
\newblock \emph{arXiv preprint arXiv:2407.21783}, 2024.

\bibitem[Fedus et~al.(2022)Fedus, Dean, and Zoph]{fedus2022review}
William Fedus, Jeff Dean, and Barret Zoph.
\newblock A review of sparse expert models in deep learning.
\newblock \emph{arXiv preprint arXiv:2209.01667}, 2022.

\bibitem[Gao et~al.(2023{\natexlab{a}})Gao, Huang, Li, and Chen]{gao2023roles}
Changjiang Gao, Shujian Huang, Jixing Li, and Jiajun Chen.
\newblock Roles of scaling and instruction tuning in language perception: Model vs. human attention.
\newblock \emph{arXiv preprint arXiv:2310.19084}, 2023{\natexlab{a}}.

\bibitem[Gao et~al.(2023{\natexlab{b}})Gao, Tow, Abbasi, Biderman, Black, DiPofi, Foster, Golding, Hsu, Le~Noac'h, Li, McDonell, Muennighoff, Ociepa, Phang, Reynolds, Schoelkopf, Skowron, Sutawika, Tang, Thite, Wang, Wang, and Zou]{eval-harness}
Leo Gao, Jonathan Tow, Baber Abbasi, Stella Biderman, Sid Black, Anthony DiPofi, Charles Foster, Laurence Golding, Jeffrey Hsu, Alain Le~Noac'h, Haonan Li, Kyle McDonell, Niklas Muennighoff, Chris Ociepa, Jason Phang, Laria Reynolds, Hailey Schoelkopf, Aviya Skowron, Lintang Sutawika, Eric Tang, Anish Thite, Ben Wang, Kevin Wang, and Andy Zou.
\newblock A framework for few-shot language model evaluation, 12 2023{\natexlab{b}}.
\newblock URL \url{https://zenodo.org/records/10256836}.

\bibitem[Golchin et~al.(2024)Golchin, Surdeanu, Bethard, Blanco, and Riloff]{golchin2024memorization}
Shahriar Golchin, Mihai Surdeanu, Steven Bethard, Eduardo Blanco, and Ellen Riloff.
\newblock Memorization in in-context learning.
\newblock \emph{arXiv preprint arXiv:2408.11546}, 2024.

\bibitem[Gu \& Dao(2023)Gu and Dao]{gu2023mamba}
Albert Gu and Tri Dao.
\newblock Mamba: Linear-time sequence modeling with selective state spaces.
\newblock \emph{arXiv preprint arXiv:2312.00752}, 2023.

\bibitem[Guo et~al.(2020)Guo, Rush, and Kim]{guo2020parameter}
Demi Guo, Alexander~M Rush, and Yoon Kim.
\newblock Parameter-efficient transfer learning with diff pruning.
\newblock \emph{arXiv preprint arXiv:2012.07463}, 2020.

\bibitem[Gururangan et~al.(2020)Gururangan, Marasovi{\'c}, Swayamdipta, Lo, Beltagy, Downey, and Smith]{gururangan2020don}
Suchin Gururangan, Ana Marasovi{\'c}, Swabha Swayamdipta, Kyle Lo, Iz~Beltagy, Doug Downey, and Noah~A Smith.
\newblock Don't stop pretraining: Adapt language models to domains and tasks.
\newblock \emph{arXiv preprint arXiv:2004.10964}, 2020.

\bibitem[Han et~al.(2021)Han, Zhang, Ding, Gu, Liu, Huo, Qiu, Yao, Zhang, Zhang, et~al.]{han2021pre}
Xu~Han, Zhengyan Zhang, Ning Ding, Yuxian Gu, Xiao Liu, Yuqi Huo, Jiezhong Qiu, Yuan Yao, Ao~Zhang, Liang Zhang, et~al.
\newblock Pre-trained models: Past, present and future.
\newblock \emph{AI Open}, 2:\penalty0 225--250, 2021.

\bibitem[Hendel et~al.(2023)Hendel, Geva, and Globerson]{hendel2023context}
Roee Hendel, Mor Geva, and Amir Globerson.
\newblock In-context learning creates task vectors.
\newblock \emph{arXiv preprint arXiv:2310.15916}, 2023.

\bibitem[Hojel et~al.(2025)Hojel, Bai, Darrell, Globerson, and Bar]{hojel2025finding}
Alberto Hojel, Yutong Bai, Trevor Darrell, Amir Globerson, and Amir Bar.
\newblock Finding visual task vectors.
\newblock In \emph{European Conference on Computer Vision}, pp.\  257--273. Springer, 2025.

\bibitem[Houlsby et~al.(2019)Houlsby, Giurgiu, Jastrzebski, Morrone, De~Laroussilhe, Gesmundo, Attariyan, and Gelly]{houlsby2019parameter}
Neil Houlsby, Andrei Giurgiu, Stanislaw Jastrzebski, Bruna Morrone, Quentin De~Laroussilhe, Andrea Gesmundo, Mona Attariyan, and Sylvain Gelly.
\newblock Parameter-efficient transfer learning for nlp.
\newblock In \emph{International conference on machine learning}, pp.\  2790--2799. PMLR, 2019.

\bibitem[Huang et~al.(2024)Huang, Mitra, Arbelle, Karlinsky, Darrell, and Herzig]{huang2024multimodal}
Brandon Huang, Chancharik Mitra, Assaf Arbelle, Leonid Karlinsky, Trevor Darrell, and Roei Herzig.
\newblock Multimodal task vectors enable many-shot multimodal in-context learning.
\newblock \emph{arXiv preprint arXiv:2406.15334}, 2024.

\bibitem[Ilharco et~al.(2022)Ilharco, Ribeiro, Wortsman, Gururangan, Schmidt, Hajishirzi, and Farhadi]{ilharco2022editing}
Gabriel Ilharco, Marco~Tulio Ribeiro, Mitchell Wortsman, Suchin Gururangan, Ludwig Schmidt, Hannaneh Hajishirzi, and Ali Farhadi.
\newblock Editing models with task arithmetic.
\newblock \emph{arXiv preprint arXiv:2212.04089}, 2022.

\bibitem[Jiang et~al.(2023)Jiang, Sablayrolles, Mensch, Bamford, Chaplot, Casas, Bressand, Lengyel, Lample, Saulnier, et~al.]{jiang2023mistral}
Albert~Q Jiang, Alexandre Sablayrolles, Arthur Mensch, Chris Bamford, Devendra~Singh Chaplot, Diego de~las Casas, Florian Bressand, Gianna Lengyel, Guillaume Lample, Lucile Saulnier, et~al.
\newblock Mistral 7b.
\newblock \emph{arXiv preprint arXiv:2310.06825}, 2023.

\bibitem[Li et~al.(2024)Li, Liu, Hu, Sun, Hu, and Zhang]{li2024context}
Dongfang Li, Zhenyu Liu, Xinshuo Hu, Zetian Sun, Baotian Hu, and Min Zhang.
\newblock In-context learning state vector with inner and momentum optimization.
\newblock \emph{arXiv preprint arXiv:2404.11225}, 2024.

\bibitem[Li et~al.(2023)Li, Lv, Yan, Lin, Zhu, Ni, Xie, Wang, and Qiu]{li2023unified}
Xiaonan Li, Kai Lv, Hang Yan, Tianyang Lin, Wei Zhu, Yuan Ni, Guotong Xie, Xiaoling Wang, and Xipeng Qiu.
\newblock Unified demonstration retriever for in-context learning.
\newblock \emph{arXiv preprint arXiv:2305.04320}, 2023.

\bibitem[Liu et~al.(2021)Liu, Shen, Zhang, Dolan, Carin, and Chen]{liu2021makes}
Jiachang Liu, Dinghan Shen, Yizhe Zhang, Bill Dolan, Lawrence Carin, and Weizhu Chen.
\newblock What makes good in-context examples for gpt-$3 $?
\newblock \emph{arXiv preprint arXiv:2101.06804}, 2021.

\bibitem[Liu et~al.(2023{\natexlab{a}})Liu, Yuan, Fu, Jiang, Hayashi, and Neubig]{liu2023pre}
Pengfei Liu, Weizhe Yuan, Jinlan Fu, Zhengbao Jiang, Hiroaki Hayashi, and Graham Neubig.
\newblock Pre-train, prompt, and predict: A systematic survey of prompting methods in natural language processing.
\newblock \emph{ACM Computing Surveys}, 55\penalty0 (9):\penalty0 1--35, 2023{\natexlab{a}}.

\bibitem[Liu et~al.(2023{\natexlab{b}})Liu, Xing, and Zou]{liu2023context}
Sheng Liu, Lei Xing, and James Zou.
\newblock In-context vectors: Making in context learning more effective and controllable through latent space steering.
\newblock \emph{arXiv preprint arXiv:2311.06668}, 2023{\natexlab{b}}.

\bibitem[Lu et~al.(2021)Lu, Bartolo, Moore, Riedel, and Stenetorp]{lu2021fantastically}
Yao Lu, Max Bartolo, Alastair Moore, Sebastian Riedel, and Pontus Stenetorp.
\newblock Fantastically ordered prompts and where to find them: Overcoming few-shot prompt order sensitivity.
\newblock \emph{arXiv preprint arXiv:2104.08786}, 2021.

\bibitem[Merity et~al.(2016)Merity, Xiong, Bradbury, and Socher]{merity2016pointer}
Stephen Merity, Caiming Xiong, James Bradbury, and Richard Socher.
\newblock Pointer sentinel mixture models.
\newblock \emph{arXiv preprint arXiv:1609.07843}, 2016.

\bibitem[Merullo et~al.(2023)Merullo, Eickhoff, and Pavlick]{merullo2023language}
Jack Merullo, Carsten Eickhoff, and Ellie Pavlick.
\newblock Language models implement simple word2vec-style vector arithmetic.
\newblock \emph{arXiv preprint arXiv:2305.16130}, 2023.

\bibitem[Mihaylov et~al.(2018)Mihaylov, Clark, Khot, and Sabharwal]{mihaylov2018can}
Todor Mihaylov, Peter Clark, Tushar Khot, and Ashish Sabharwal.
\newblock Can a suit of armor conduct electricity? a new dataset for open book question answering.
\newblock \emph{arXiv preprint arXiv:1809.02789}, 2018.

\bibitem[Min et~al.(2021)Min, Lewis, Zettlemoyer, and Hajishirzi]{min2021metaicl}
Sewon Min, Mike Lewis, Luke Zettlemoyer, and Hannaneh Hajishirzi.
\newblock Metaicl: Learning to learn in context.
\newblock \emph{arXiv preprint arXiv:2110.15943}, 2021.

\bibitem[Mu et~al.(2024)Mu, Li, and Goodman]{mu2024learning}
Jesse Mu, Xiang Li, and Noah Goodman.
\newblock Learning to compress prompts with gist tokens.
\newblock \emph{Advances in Neural Information Processing Systems}, 36, 2024.

\bibitem[Nangia et~al.(2020)Nangia, Vania, Bhalerao, and Bowman]{nangia2020crows}
Nikita Nangia, Clara Vania, Rasika Bhalerao, and Samuel~R Bowman.
\newblock Crows-pairs: A challenge dataset for measuring social biases in masked language models.
\newblock \emph{arXiv preprint arXiv:2010.00133}, 2020.

\bibitem[Nie et~al.(2022)Nie, Chen, Zhang, and Cheng]{nie2022improving}
Feng Nie, Meixi Chen, Zhirui Zhang, and Xu~Cheng.
\newblock Improving few-shot performance of language models via nearest neighbor calibration.
\newblock \emph{arXiv preprint arXiv:2212.02216}, 2022.

\bibitem[Olsson et~al.(2022)Olsson, Elhage, Nanda, Joseph, DasSarma, Henighan, Mann, Askell, Bai, Chen, et~al.]{olsson2022context}
Catherine Olsson, Nelson Elhage, Neel Nanda, Nicholas Joseph, Nova DasSarma, Tom Henighan, Ben Mann, Amanda Askell, Yuntao Bai, Anna Chen, et~al.
\newblock In-context learning and induction heads.
\newblock \emph{arXiv preprint arXiv:2209.11895}, 2022.

\bibitem[Panigrahi et~al.(2023)Panigrahi, Saunshi, Zhao, and Arora]{panigrahi2023task}
Abhishek Panigrahi, Nikunj Saunshi, Haoyu Zhao, and Sanjeev Arora.
\newblock Task-specific skill localization in fine-tuned language models.
\newblock In \emph{International Conference on Machine Learning}, pp.\  27011--27033. PMLR, 2023.

\bibitem[Parrish et~al.(2021)Parrish, Chen, Nangia, Padmakumar, Phang, Thompson, Htut, and Bowman]{parrish2021bbq}
Alicia Parrish, Angelica Chen, Nikita Nangia, Vishakh Padmakumar, Jason Phang, Jana Thompson, Phu~Mon Htut, and Samuel~R Bowman.
\newblock Bbq: A hand-built bias benchmark for question answering.
\newblock \emph{arXiv preprint arXiv:2110.08193}, 2021.

\bibitem[Pfeiffer et~al.(2020)Pfeiffer, Vuli{\'c}, Gurevych, and Ruder]{pfeiffer2020mad}
Jonas Pfeiffer, Ivan Vuli{\'c}, Iryna Gurevych, and Sebastian Ruder.
\newblock Mad-x: An adapter-based framework for multi-task cross-lingual transfer.
\newblock \emph{arXiv preprint arXiv:2005.00052}, 2020.

\bibitem[Pfeiffer et~al.(2023)Pfeiffer, Ruder, Vuli{\'c}, and Ponti]{pfeiffer2023modular}
Jonas Pfeiffer, Sebastian Ruder, Ivan Vuli{\'c}, and Edoardo~Maria Ponti.
\newblock Modular deep learning.
\newblock \emph{arXiv preprint arXiv:2302.11529}, 2023.

\bibitem[Robertson et~al.(2009)Robertson, Zaragoza, et~al.]{robertson2009probabilistic}
Stephen Robertson, Hugo Zaragoza, et~al.
\newblock The probabilistic relevance framework: Bm25 and beyond.
\newblock \emph{Foundations and Trends{\textregistered} in Information Retrieval}, 3\penalty0 (4):\penalty0 333--389, 2009.

\bibitem[Rubin et~al.(2021)Rubin, Herzig, and Berant]{rubin2021learning}
Ohad Rubin, Jonathan Herzig, and Jonathan Berant.
\newblock Learning to retrieve prompts for in-context learning.
\newblock \emph{arXiv preprint arXiv:2112.08633}, 2021.

\bibitem[Saxton et~al.(2019)Saxton, Grefenstette, Hill, and Kohli]{saxton2019analysing}
David Saxton, Edward Grefenstette, Felix Hill, and Pushmeet Kohli.
\newblock Analysing mathematical reasoning abilities of neural models.
\newblock \emph{arXiv preprint arXiv:1904.01557}, 2019.

\bibitem[Shao et~al.(2023)Shao, Cai, Liao, Zheng, Yang, et~al.]{shao2023compositional}
Nan Shao, Zefan Cai, Chonghua Liao, Yanan Zheng, Zhilin Yang, et~al.
\newblock Compositional task representations for large language models.
\newblock In \emph{The Eleventh International Conference on Learning Representations}, 2023.

\bibitem[Shi et~al.(2022)Shi, Zhang, Bai, and Lin]{shi2022xricl}
Peng Shi, Rui Zhang, He~Bai, and Jimmy Lin.
\newblock Xricl: Cross-lingual retrieval-augmented in-context learning for cross-lingual text-to-sql semantic parsing.
\newblock \emph{arXiv preprint arXiv:2210.13693}, 2022.

\bibitem[Singh et~al.(2024)Singh, Moskovitz, Hill, Chan, and Saxe]{singh2024needs}
Aaditya~K Singh, Ted Moskovitz, Felix Hill, Stephanie~CY Chan, and Andrew~M Saxe.
\newblock What needs to go right for an induction head? a mechanistic study of in-context learning circuits and their formation.
\newblock \emph{arXiv preprint arXiv:2404.07129}, 2024.

\bibitem[Su et~al.(2022)Su, Kasai, Wu, Shi, Wang, Xin, Zhang, Ostendorf, Zettlemoyer, Smith, et~al.]{su2022selective}
Hongjin Su, Jungo Kasai, Chen~Henry Wu, Weijia Shi, Tianlu Wang, Jiayi Xin, Rui Zhang, Mari Ostendorf, Luke Zettlemoyer, Noah~A Smith, et~al.
\newblock Selective annotation makes language models better few-shot learners.
\newblock \emph{arXiv preprint arXiv:2209.01975}, 2022.

\bibitem[Sun et~al.(2023)Sun, Shaib, and Wallace]{sun2023evaluating}
Jiuding Sun, Chantal Shaib, and Byron~C Wallace.
\newblock Evaluating the zero-shot robustness of instruction-tuned language models.
\newblock \emph{arXiv preprint arXiv:2306.11270}, 2023.

\bibitem[Suzgun et~al.(2022)Suzgun, Scales, Sch{\"a}rli, Gehrmann, Tay, Chung, Chowdhery, Le, Chi, Zhou, et~al.]{suzgun2022challenging}
Mirac Suzgun, Nathan Scales, Nathanael Sch{\"a}rli, Sebastian Gehrmann, Yi~Tay, Hyung~Won Chung, Aakanksha Chowdhery, Quoc~V Le, Ed~H Chi, Denny Zhou, et~al.
\newblock Challenging big-bench tasks and whether chain-of-thought can solve them.
\newblock \emph{arXiv preprint arXiv:2210.09261}, 2022.

\bibitem[Talmor et~al.(2018)Talmor, Herzig, Lourie, and Berant]{talmor2018commonsenseqa}
Alon Talmor, Jonathan Herzig, Nicholas Lourie, and Jonathan Berant.
\newblock Commonsenseqa: A question answering challenge targeting commonsense knowledge.
\newblock \emph{arXiv preprint arXiv:1811.00937}, 2018.

\bibitem[Team et~al.(2023)Team, Anil, Borgeaud, Wu, Alayrac, Yu, Soricut, Schalkwyk, Dai, Hauth, et~al.]{team2023gemini}
Gemini Team, Rohan Anil, Sebastian Borgeaud, Yonghui Wu, Jean-Baptiste Alayrac, Jiahui Yu, Radu Soricut, Johan Schalkwyk, Andrew~M Dai, Anja Hauth, et~al.
\newblock Gemini: a family of highly capable multimodal models.
\newblock \emph{arXiv preprint arXiv:2312.11805}, 2023.

\bibitem[Thirunavukarasu et~al.(2023)Thirunavukarasu, Ting, Elangovan, Gutierrez, Tan, and Ting]{thirunavukarasu2023large}
Arun~James Thirunavukarasu, Darren Shu~Jeng Ting, Kabilan Elangovan, Laura Gutierrez, Ting~Fang Tan, and Daniel Shu~Wei Ting.
\newblock Large language models in medicine.
\newblock \emph{Nature medicine}, 29\penalty0 (8):\penalty0 1930--1940, 2023.

\bibitem[Todd et~al.(2023)Todd, Li, Sharma, Mueller, Wallace, and Bau]{todd2023function}
Eric Todd, Millicent~L Li, Arnab~Sen Sharma, Aaron Mueller, Byron~C Wallace, and David Bau.
\newblock Function vectors in large language models.
\newblock \emph{arXiv preprint arXiv:2310.15213}, 2023.

\bibitem[Touvron et~al.(2023)Touvron, Lavril, Izacard, Martinet, Lachaux, Lacroix, Rozi{\`e}re, Goyal, Hambro, Azhar, et~al.]{touvron2023llama}
Hugo Touvron, Thibaut Lavril, Gautier Izacard, Xavier Martinet, Marie-Anne Lachaux, Timoth{\'e}e Lacroix, Baptiste Rozi{\`e}re, Naman Goyal, Eric Hambro, Faisal Azhar, et~al.
\newblock Llama: Open and efficient foundation language models.
\newblock \emph{arXiv preprint arXiv:2302.13971}, 2023.

\bibitem[Vacareanu et~al.(2024)Vacareanu, Negru, Suciu, and Surdeanu]{vacareanu2024words}
Robert Vacareanu, Vlad-Andrei Negru, Vasile Suciu, and Mihai Surdeanu.
\newblock From words to numbers: Your large language model is secretly a capable regressor when given in-context examples.
\newblock \emph{arXiv preprint arXiv:2404.07544}, 2024.

\bibitem[Vaswani(2017)]{vaswani2017attention}
A~Vaswani.
\newblock Attention is all you need.
\newblock \emph{Advances in Neural Information Processing Systems}, 2017.

\bibitem[Von~Oswald et~al.(2023)Von~Oswald, Niklasson, Randazzo, Sacramento, Mordvintsev, Zhmoginov, and Vladymyrov]{von2023transformers}
Johannes Von~Oswald, Eyvind Niklasson, Ettore Randazzo, Jo{\~a}o Sacramento, Alexander Mordvintsev, Andrey Zhmoginov, and Max Vladymyrov.
\newblock Transformers learn in-context by gradient descent.
\newblock In \emph{International Conference on Machine Learning}, pp.\  35151--35174. PMLR, 2023.

\bibitem[Wang(2018)]{wang2018glue}
Alex Wang.
\newblock Glue: A multi-task benchmark and analysis platform for natural language understanding.
\newblock \emph{arXiv preprint arXiv:1804.07461}, 2018.

\bibitem[Wang et~al.(2019)Wang, Pruksachatkun, Nangia, Singh, Michael, Hill, Levy, and Bowman]{wang2019superglue}
Alex Wang, Yada Pruksachatkun, Nikita Nangia, Amanpreet Singh, Julian Michael, Felix Hill, Omer Levy, and Samuel Bowman.
\newblock Superglue: A stickier benchmark for general-purpose language understanding systems.
\newblock \emph{Advances in neural information processing systems}, 32, 2019.

\bibitem[Wang et~al.(2024)Wang, Ma, Zhang, Ni, Chandra, Guo, Ren, Arulraj, He, Jiang, et~al.]{wang2024mmlu}
Yubo Wang, Xueguang Ma, Ge~Zhang, Yuansheng Ni, Abhranil Chandra, Shiguang Guo, Weiming Ren, Aaran Arulraj, Xuan He, Ziyan Jiang, et~al.
\newblock Mmlu-pro: A more robust and challenging multi-task language understanding benchmark.
\newblock \emph{arXiv preprint arXiv:2406.01574}, 2024.

\bibitem[Wolf et~al.(2020)Wolf, Debut, Sanh, Chaumond, Delangue, Moi, Cistac, Rault, Louf, Funtowicz, et~al.]{wolf2020transformers}
Thomas Wolf, Lysandre Debut, Victor Sanh, Julien Chaumond, Clement Delangue, Anthony Moi, Pierric Cistac, Tim Rault, R{\'e}mi Louf, Morgan Funtowicz, et~al.
\newblock Transformers: State-of-the-art natural language processing.
\newblock In \emph{Proceedings of the 2020 conference on empirical methods in natural language processing: system demonstrations}, pp.\  38--45, 2020.

\bibitem[Xiao et~al.(2024)Xiao, Zhang, Song, Jiang, Yao, Han, Wang, Wang, Huang, Lin, et~al.]{xiao2024configurable}
Chaojun Xiao, Zhengyan Zhang, Chenyang Song, Dazhi Jiang, Feng Yao, Xu~Han, Xiaozhi Wang, Shuo Wang, Yufei Huang, Guanyu Lin, et~al.
\newblock Configurable foundation models: Building llms from a modular perspective.
\newblock \emph{arXiv preprint arXiv:2409.02877}, 2024.

\bibitem[Yan et~al.(2023)Yan, Xu, Song, Wu, Li, and Zhang]{yan2023understanding}
Jianhao Yan, Jin Xu, Chiyu Song, Chenming Wu, Yafu Li, and Yue Zhang.
\newblock Understanding in-context learning from repetitions.
\newblock \emph{arXiv preprint arXiv:2310.00297}, 2023.

\bibitem[Yu et~al.(2024)Yu, Yu, Yu, Huang, and Li]{yu2024language}
Le~Yu, Bowen Yu, Haiyang Yu, Fei Huang, and Yongbin Li.
\newblock Language models are super mario: Absorbing abilities from homologous models as a free lunch.
\newblock In \emph{Forty-first International Conference on Machine Learning}, 2024.

\bibitem[Zaken et~al.(2021)Zaken, Ravfogel, and Goldberg]{zaken2021bitfit}
Elad~Ben Zaken, Shauli Ravfogel, and Yoav Goldberg.
\newblock Bitfit: Simple parameter-efficient fine-tuning for transformer-based masked language-models.
\newblock \emph{arXiv preprint arXiv:2106.10199}, 2021.

\bibitem[Zellers et~al.(2019)Zellers, Holtzman, Bisk, Farhadi, and Choi]{zellers2019hellaswag}
Rowan Zellers, Ari Holtzman, Yonatan Bisk, Ali Farhadi, and Yejin Choi.
\newblock Hellaswag: Can a machine really finish your sentence?
\newblock \emph{arXiv preprint arXiv:1905.07830}, 2019.

\bibitem[Zhang et~al.(2023{\natexlab{a}})Zhang, Deng, Liu, Pan, and Bing]{zhang2023sentiment}
Wenxuan Zhang, Yue Deng, Bing Liu, Sinno~Jialin Pan, and Lidong Bing.
\newblock Sentiment analysis in the era of large language models: A reality check.
\newblock \emph{arXiv preprint arXiv:2305.15005}, 2023{\natexlab{a}}.

\bibitem[Zhang et~al.(2023{\natexlab{b}})Zhang, Zeng, Lin, Xiao, Wang, Han, Liu, Xie, Sun, and Zhou]{zhang2023emergent}
Zhengyan Zhang, Zhiyuan Zeng, Yankai Lin, Chaojun Xiao, Xiaozhi Wang, Xu~Han, Zhiyuan Liu, Ruobing Xie, Maosong Sun, and Jie Zhou.
\newblock Emergent modularity in pre-trained transformers.
\newblock \emph{arXiv preprint arXiv:2305.18390}, 2023{\natexlab{b}}.

\bibitem[Zhao et~al.(2022)Zhao, Xu, Yang, and Gao]{zhao2022consistent}
Kang Zhao, Hua Xu, Jiangong Yang, and Kai Gao.
\newblock Consistent representation learning for continual relation extraction.
\newblock \emph{arXiv preprint arXiv:2203.02721}, 2022.

\bibitem[Zhao et~al.(2021)Zhao, Wallace, Feng, Klein, and Singh]{zhao2021calibrate}
Zihao Zhao, Eric Wallace, Shi Feng, Dan Klein, and Sameer Singh.
\newblock Calibrate before use: Improving few-shot performance of language models.
\newblock In \emph{International conference on machine learning}, pp.\  12697--12706. PMLR, 2021.

\bibitem[Zou et~al.(2023)Zou, Phan, Chen, Campbell, Guo, Ren, Pan, Yin, Mazeika, Dombrowski, et~al.]{zou2023representation}
Andy Zou, Long Phan, Sarah Chen, James Campbell, Phillip Guo, Richard Ren, Alexander Pan, Xuwang Yin, Mantas Mazeika, Ann-Kathrin Dombrowski, et~al.
\newblock Representation engineering: A top-down approach to ai transparency.
\newblock \emph{arXiv preprint arXiv:2310.01405}, 2023.

\end{thebibliography}
\bibliographystyle{configuration/iclr2025_conference}

\appendix

\section{Intervention Strategies} \label{appendix:intervention_strategy}
\begin{figure}[H]
    \centering 
    \includegraphics[width=1.0\textwidth,]{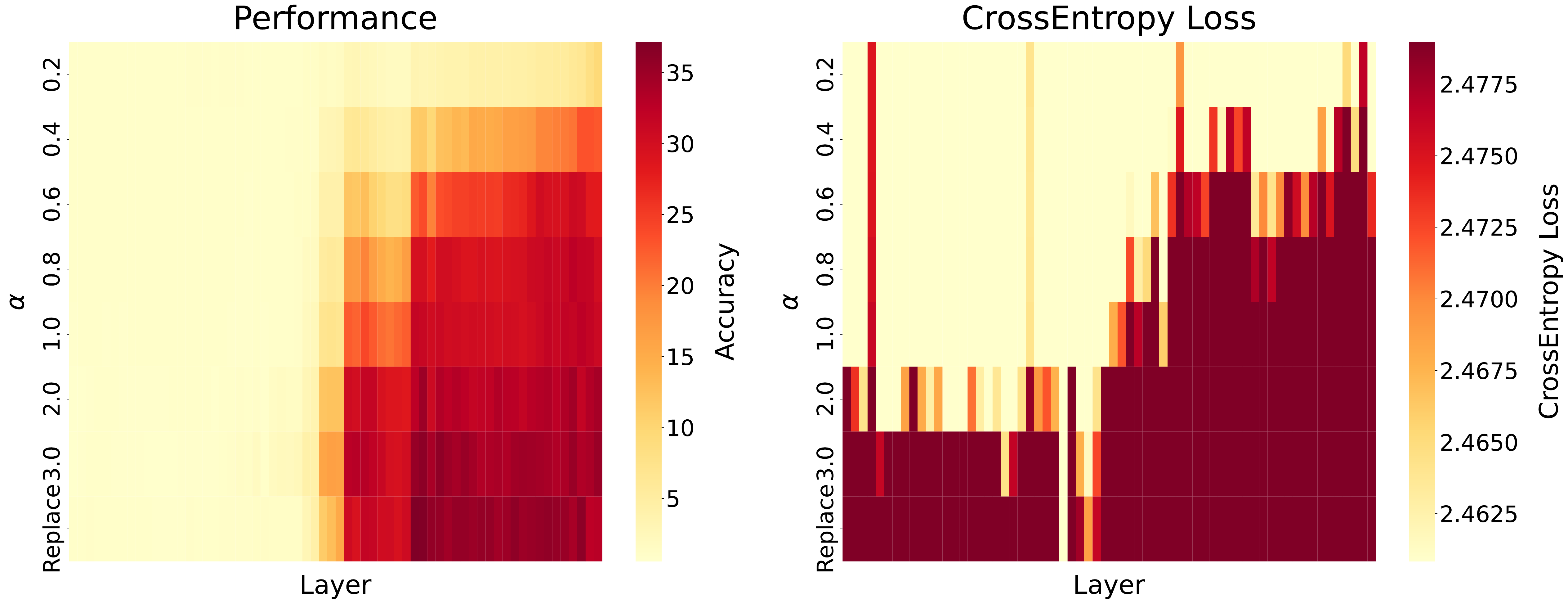}
    \caption{\small Varying intervention strengths affect accuracy and cross-entropy loss in Mamba on valid set
    of 20 tasks across different layer. }
    \label{fig:mamba_acc_ce_loss}
\end{figure}

\begin{figure}[H]
    \centering
    \includegraphics[width=1.0\textwidth,]{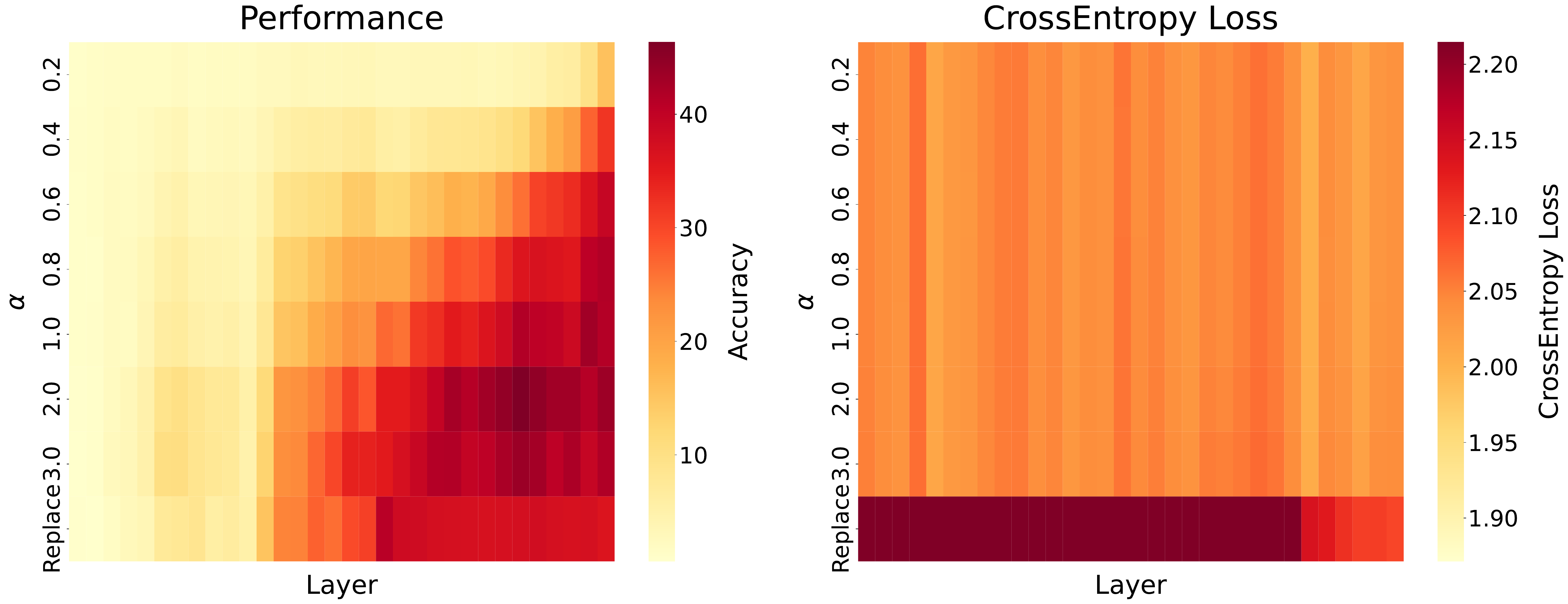}
    \caption{\small Varying intervention strengths affect accuracy and cross-entropy loss in Mistral on valid set
    of 20 tasks across different layer. }
    \label{fig:mistral_acc_ce_loss}
\end{figure}

\begin{figure}[H]
    \centering
    \includegraphics[width=1.0\textwidth,]{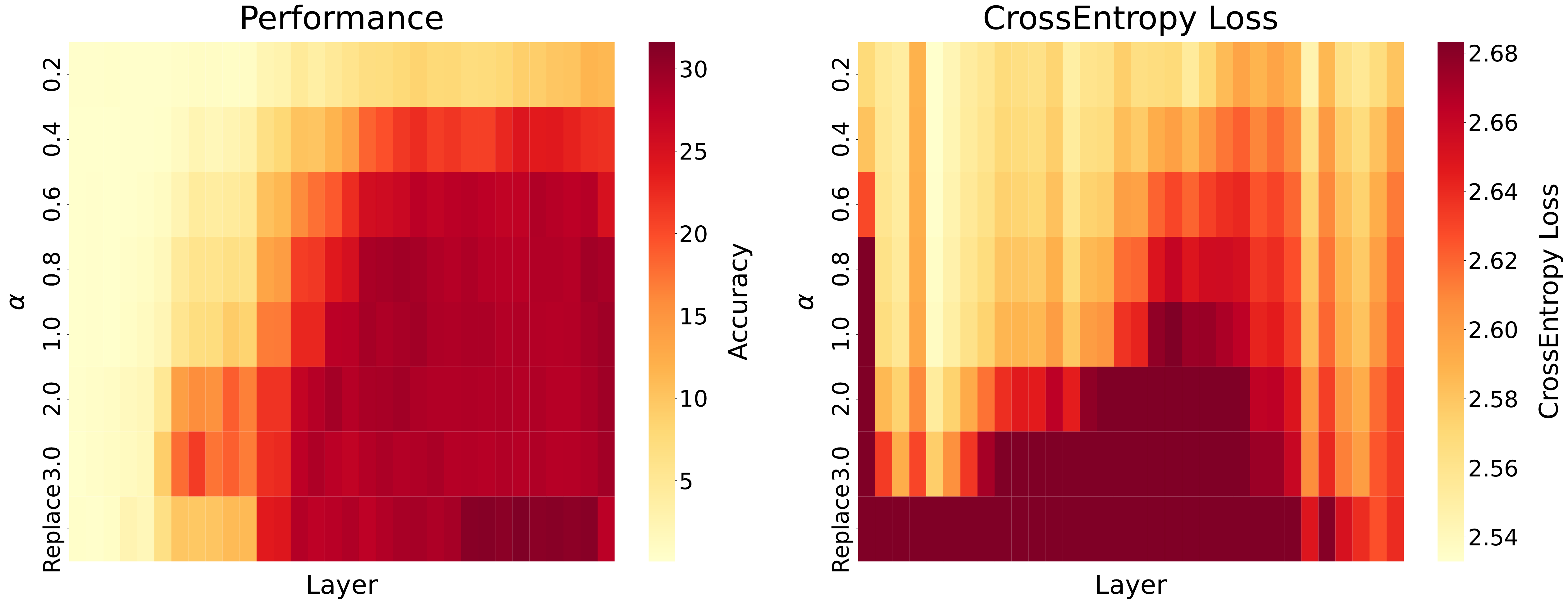}
    \caption{\small Varying intervention strengths affect accuracy and cross-entropy loss in Pythia on valid set
    of 20 tasks across different layer. }
    \label{fig:pyhtia_acc_ce_loss}
\end{figure}

As mentioned in Section~\ref{sec_construct_library}, we choose $\tilde{\mathbf{h}}_l = \mathbf{h}_l + \alpha \cdot \boldsymbol{\theta}_l$ as the intervention strategy, where $\alpha$ is a scaling factor that controls the intervention strength.
We observe the performance and cross-entropy loss across a diverse set of 20 tasks by varying $\alpha$.
The results for Mistral, Mamba, and Pythia are shown in Figures~\ref{fig:mamba_acc_ce_loss}, \ref{fig:mistral_acc_ce_loss}, and \ref{fig:pyhtia_acc_ce_loss}, respectively.
The results reveal a similar trade-off between task performance and language modeling capability as the intervention strength increases.
Among the strategies tested, the additive approach consistently demonstrates superior performance across a wide range of tasks while minimizing degradation in language modeling ability.
Across different models, $\alpha$ can be set to 2.0 to achieve a good balance between task performance and language modeling capability.

\subsection{Possible Interpretation of Trade-off}

The trade-off between intervention strength and language modeling performance can be explained through neural circuit interactions.
Prior work suggests ICL and general language modeling operate with different circuits \citep{chan2022data,olsson2022context,singh2024needs}.
We hypothesize that stronger interventions redirect activation patterns from pretrained language modeling circuitry towards ICL-specific patterns, creating tension between these distinct computational paths.
As intervention strength increases, activations appear to deviate further from their pretrained configurations, potentially explaining the degraded performance on general language modeling tasks like WikiText.
This observation suggests that optimal intervention strength requires careful balancing rather than maximization.
Further investigation of these circuit-level understanding remains an important direction for future research.

\section{Best Layer for different Tasks.} \label{appendix:best_layer_distribution}
\begin{figure}[H]
    \centering
    \includegraphics[width=1.0\textwidth,]{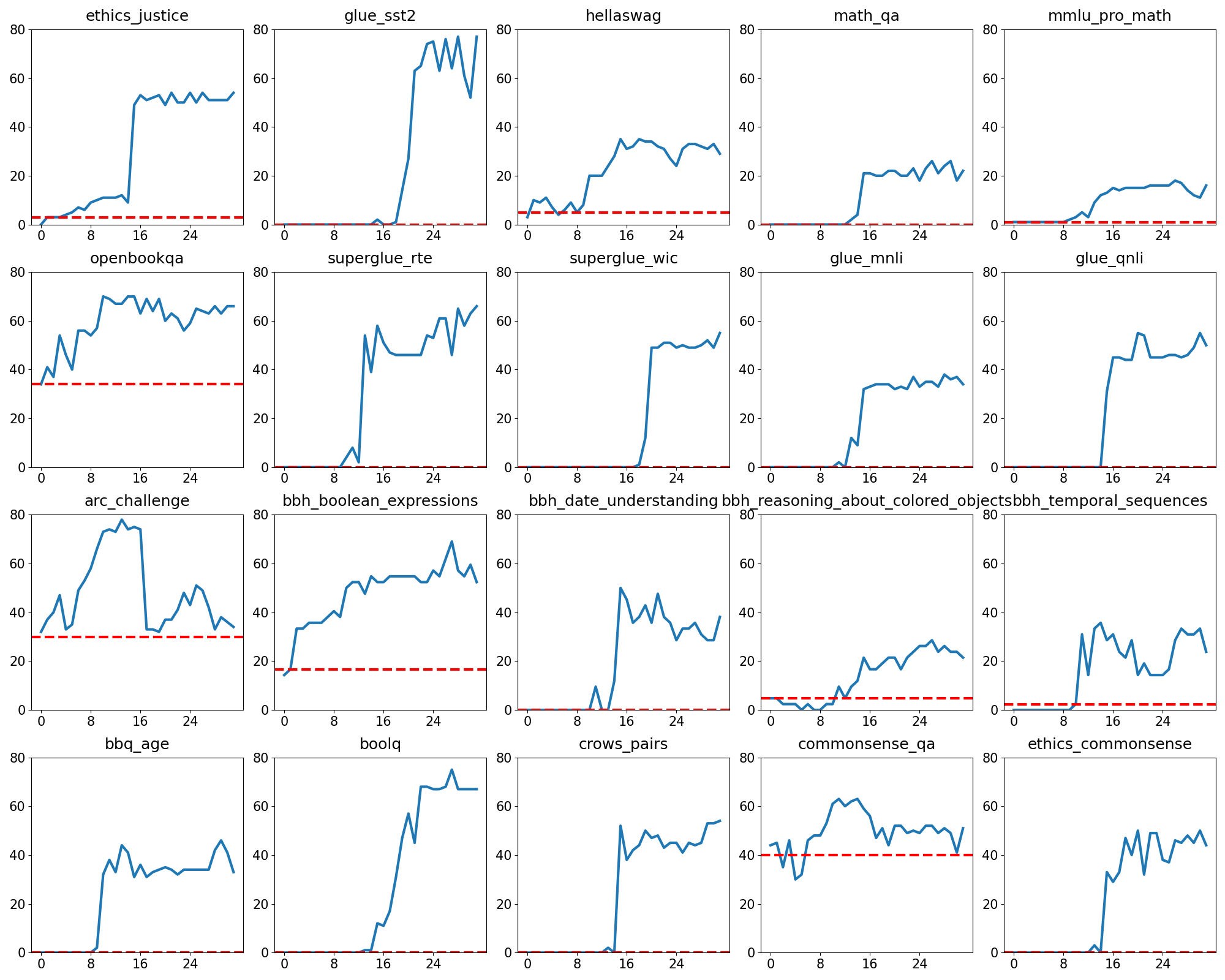}
    
    \caption{\small Performance distribution varying intervention layer on Llama3 8B in 20 tasks on Valid Set.}
    \label{fig:llama_layer_distribution}
    \vspace{-0.95em}
    
\end{figure}

\begin{figure}[H]
    \centering
    \includegraphics[width=1.0\textwidth,]{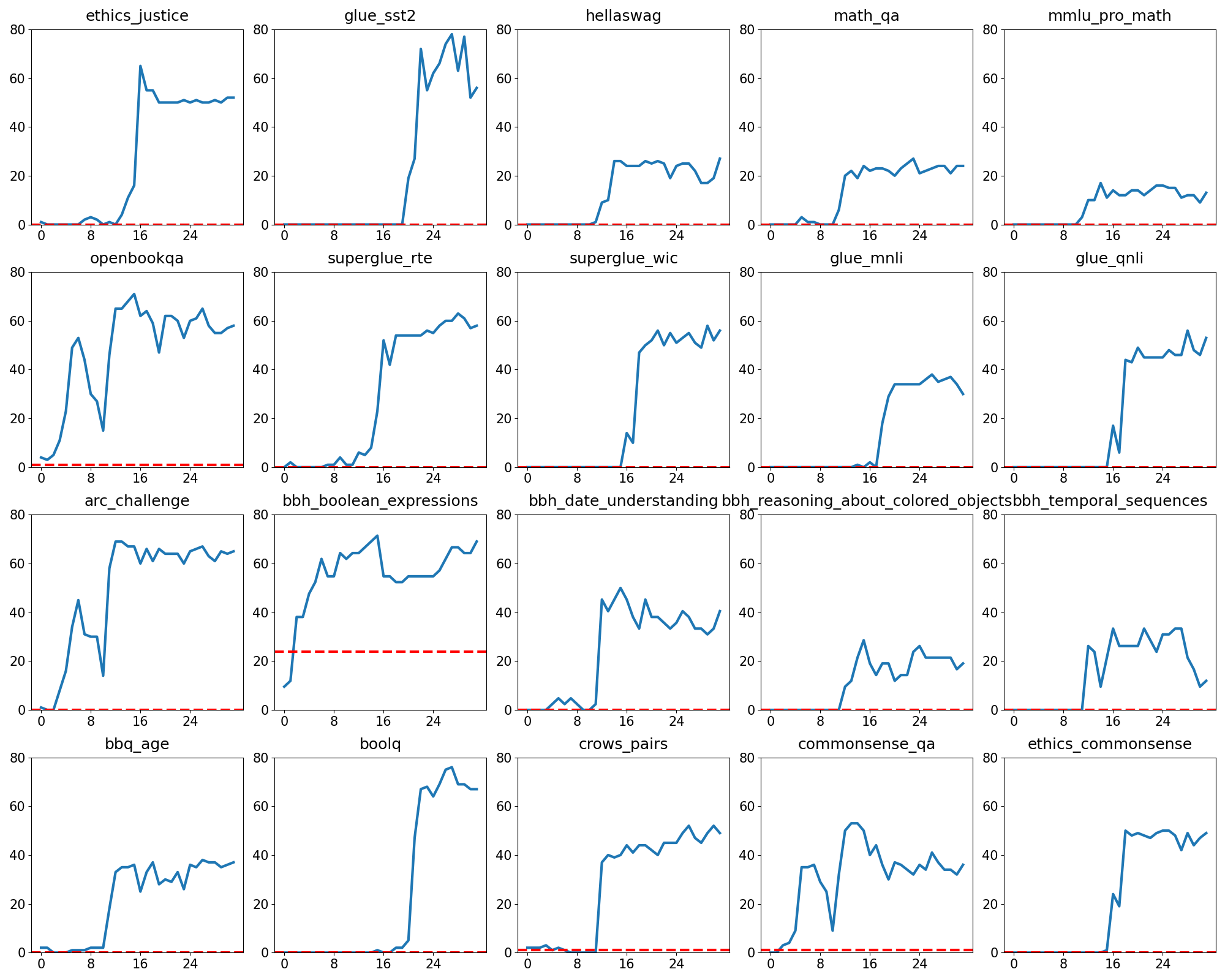}
    \caption{\small Performance distribution varying intervention layer on Mistral in 20 tasks on Valid Set.}
    \label{fig:mistral_layer_distribution}
    \vspace{-0.95em}
    
\end{figure}

\begin{figure}[H]
    \centering
    \includegraphics[width=1.0\textwidth,]{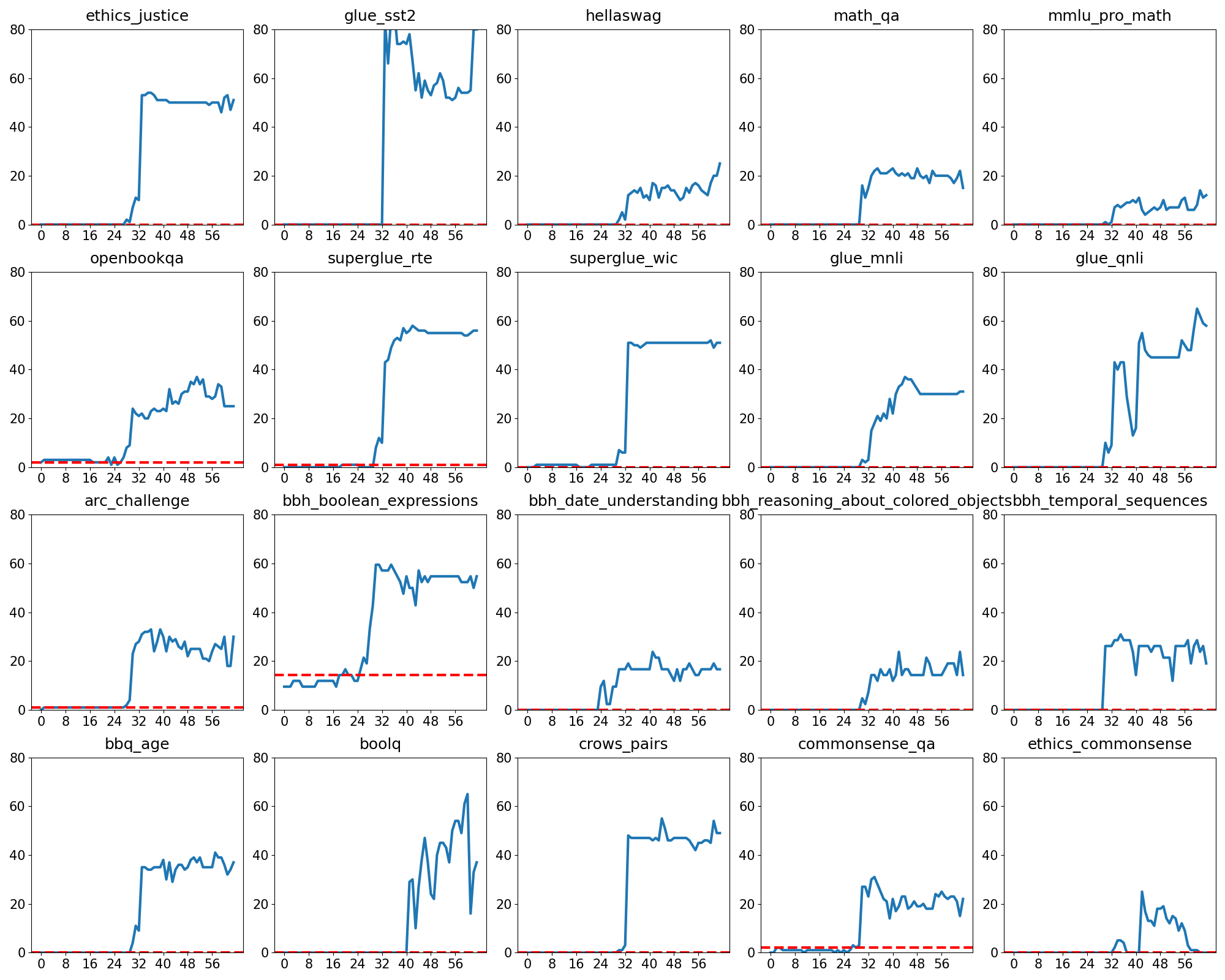}
    \caption{\small Performance distribution varying intervention layer on Mamba in 20 tasks on Valid Set.}
    \label{fig:mamba_layer_distribution}
    \vspace{-0.95em}
    
\end{figure}

As mentioned in Section~\ref{sec_construct_library}, we identify the optimal intervention layer for different tasks.
Figures~\ref{fig:llama_layer_distribution}, \ref{fig:mistral_layer_distribution}, and \ref{fig:mamba_layer_distribution} illustrate that the optimal intervention layer varies significantly across tasks.
For instance, the ARC challenge task achieves the best performance when intervening in the middle layers, while the GLUE SST2 task performs best when intervening in the later layers for the Llama-3 model.
Furthermore, intervening at different layers leads to substantial performance variations.
Therefore, instead of fixing the intervention layer as in \cite{todd2023function}, we propose a dynamic layer selection approach to identify the optimal intervention layer for each task.

\section{Data Construction for Retriever} \label{appendix:data_construction}
As mentioned in Section~\ref{sec:retrieval}, we train a retriever on the constructed data.
To train the retriever as a classifier, we construct data pairs to determine whether a task-specific prompt and an ICL prompt belong to the same task.
For each task and each template, we sample two examples from the validation set.
We create positive and negative pairs, where each pair consists of a task-specific prompt and an ICL prompt from the library.
Negative examples are formed by randomly pairing task-specific prompts with ICL prompts from different tasks.
We balance the data distribution across different tasks, ensuring that each task has an equal number of positive and negative pairs.

\section{Recall Sweep} \label{appendix:recall_swep}
\begin{figure}[]
    \centering
    \begin{subfigure}[b]{0.45\textwidth}
        \centering
        \includegraphics[width=\textwidth]{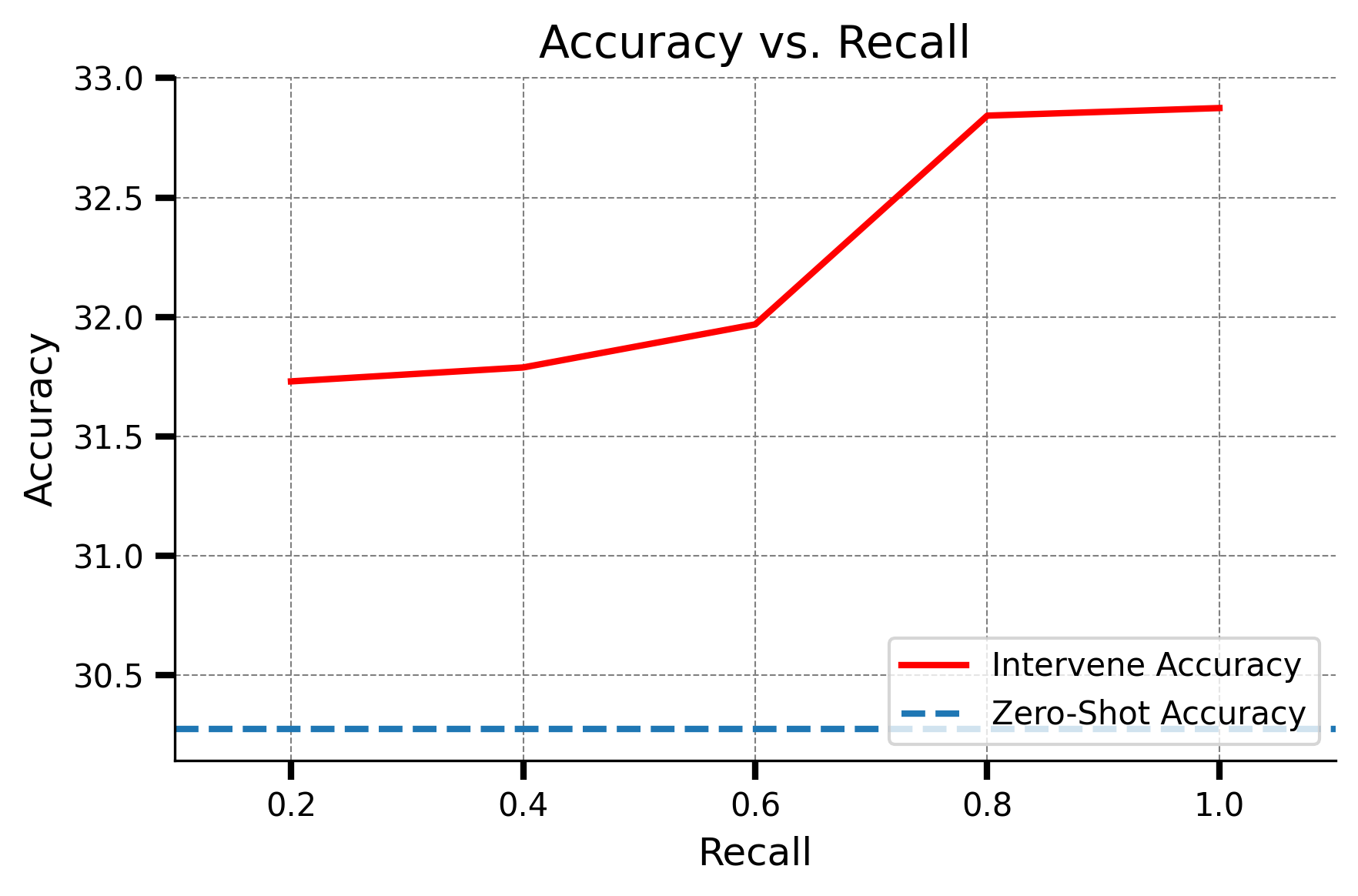}
        \caption{Pythia}
        \label{fig:pythia_recall_sweep}
    \end{subfigure}
    \hfill
    \begin{subfigure}[b]{0.45\textwidth}
        \centering
        \includegraphics[width=\textwidth]{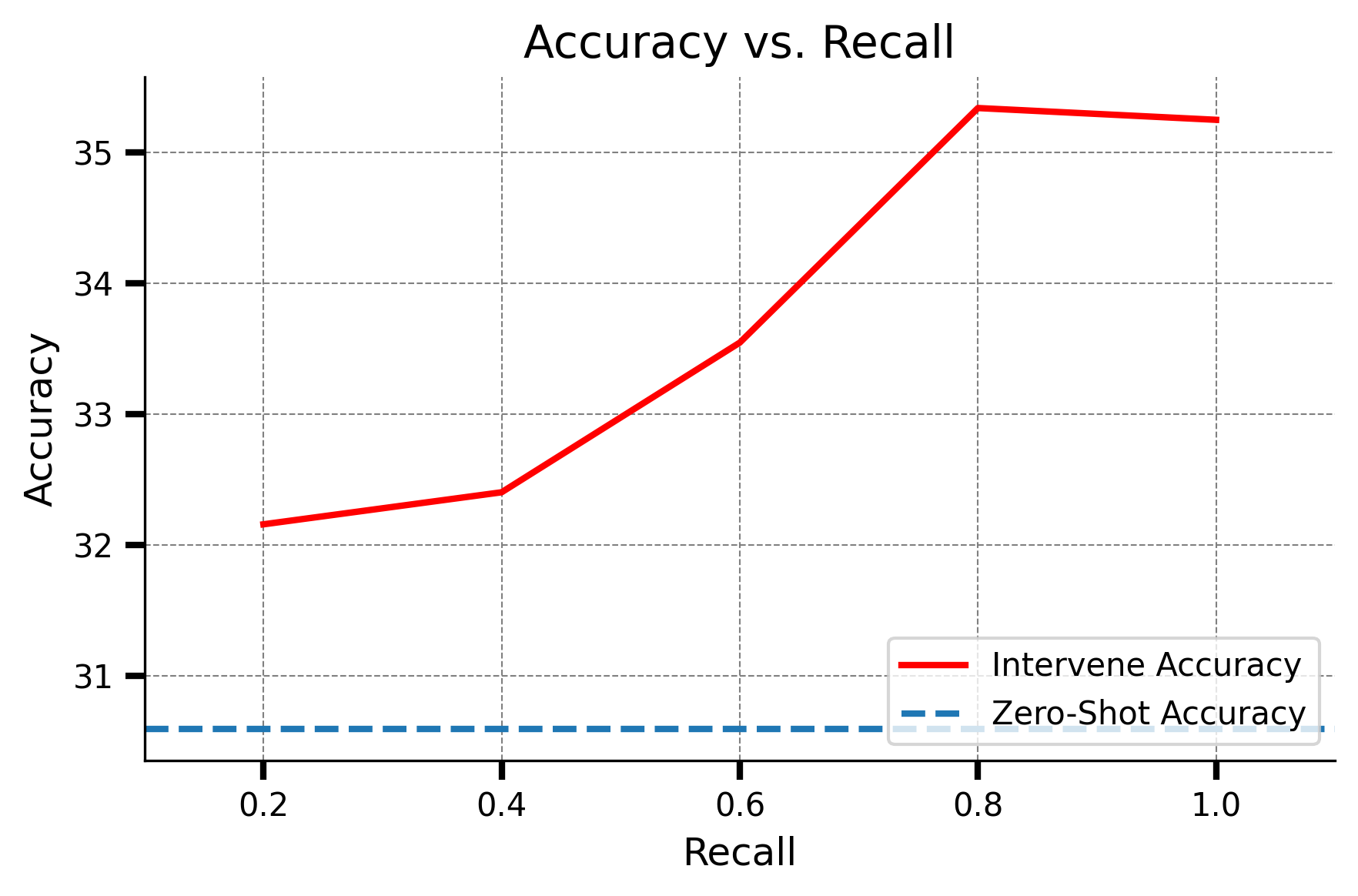}
        \label{fig:mamba_recall_sweep}
        \caption{Mamba}
        
    \end{subfigure}
    \hfill
    \begin{subfigure}[b]{0.45\textwidth}
        \centering
        \includegraphics[width=\textwidth]{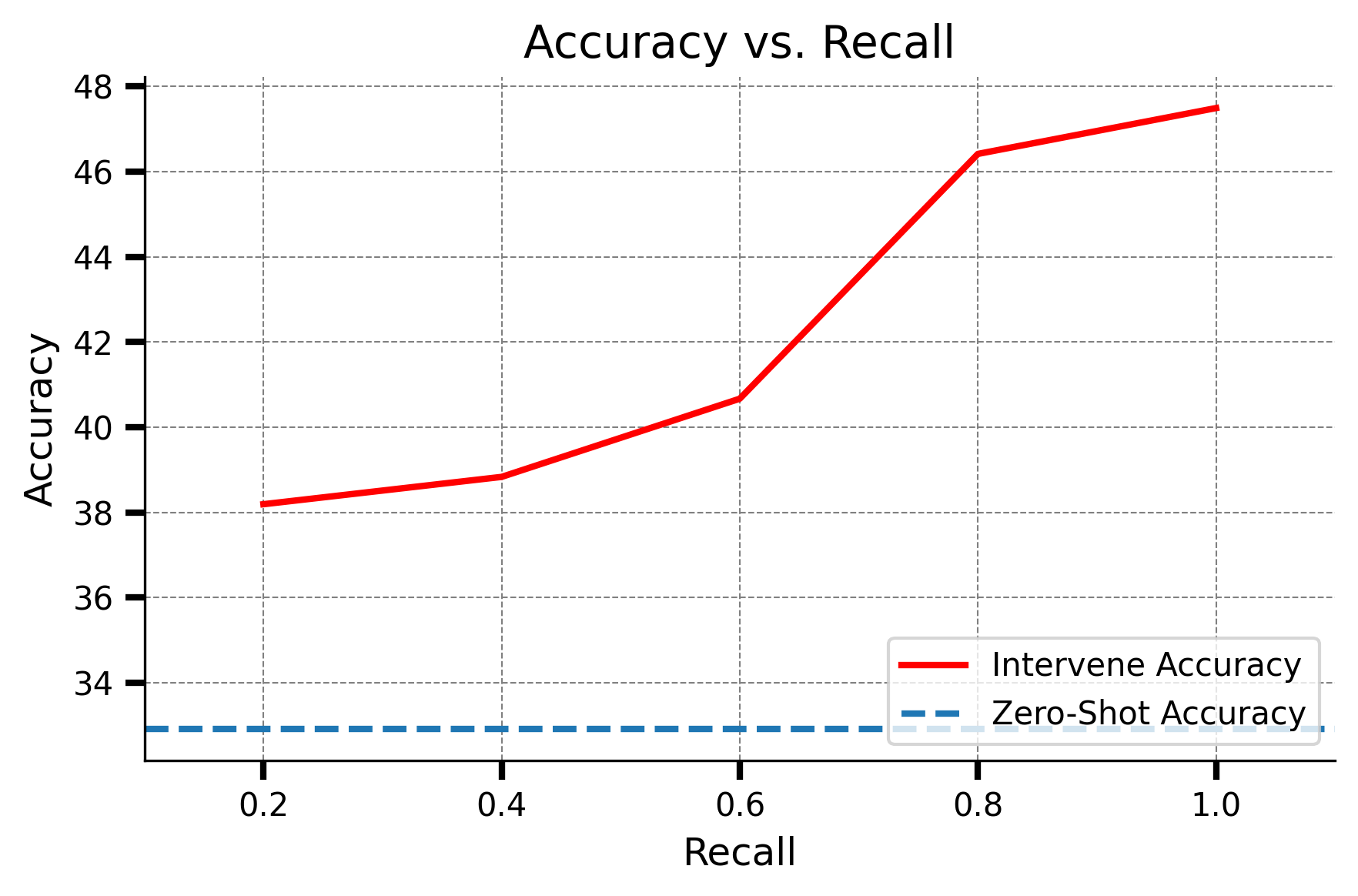}
        \label{fig:mistral_recall_sweep}
        \caption{Mistral}
        
    \end{subfigure}
    \vspace{-1em}
    \caption{\small Accuracy vs. Recall curves for Pythia, Mamba, and Mistral models, illustrating the performance trade-off at different recall levels.}
    \label{fig:other_mdoels_recall_sweep}
  
\end{figure}

As mentioned in Section~\ref{sec:retrieval}, we sweep the threshold determined by recall, and as shown in Figure\ref{fig:other_mdoels_recall_sweep}, the results reveal that a recall of 0.8 provides an optimal balance between accuracy and recall for our pipeline across Pythia, Mamba, and Mistral models.

We observe a similar increasing trend in performance as recall increases for all three models. At lower recall values, the intervene accuracy is close to the zero-shot accuracy, indicating that the retrieved prompts may not be relevant to the task. As recall increases, the intervene accuracy improves significantly, demonstrating the effectiveness of the proposed approach in selecting appropriate prompts for intervention.

Based on these observations, we choose a recall value of 0.8 to determine the threshold for filtering prompts across different models, as it strikes a balance between maximizing accuracy and maintaining a reasonable recall level.

\section{Evaluation Setting} \label{appendix:evaluation_setting}
\subsection{ICL setting}
As \ref{para:evaluation} states, we first primarily implement \alg on the traditional In-Context Learning (ICL) setting.
We find that zero-shot Large Language Models (LLMs) cannot answer properly with contextual guidance.
Although \alg works on such a traditional ICL setting, as shown in Table~\ref{tab:icl_results} and \ref{tab:unseen_icl}, the zero-shot accuracy is almost 0, which is not plausible to evaluate model's performance.
Therefore, we think it's not fair to augment model performance on such traditional zero-shot queries.
\begin{table}[H]
    \centering
    \caption{Performance of \alg across model and tasks in ICL setting.}
    \label{tab:icl_results}
    \resizebox{\columnwidth}{!}{%
    \begin{tabular}{cc|c|ccccc|c}
    \toprule
    {\color[HTML]{0D0D0D} \textbf{model}} & \multicolumn{1}{l|}{}                   & \textbf{\#Tokens}                    & \textbf{NLU}                               & \textbf{Reasoning}                & \textbf{Knowledge}                         & \textbf{Math}                     & \textbf{Safety}                            & \textbf{Avg.}                              \\ \midrule
                                          & {\color[HTML]{999999} 16-shot}          & {\color[HTML]{999999} 1553.4 ± 2.8}  & {\color[HTML]{999999} 60.3 ± 1.2}          & {\color[HTML]{999999} 55.6 ± 0.2} & {\color[HTML]{999999} 69.2 ± 2.0}          & {\color[HTML]{999999} 27.0 ± 0.0} & {\color[HTML]{999999} 60.9 ± 0.5}          & {\color[HTML]{999999} 54.6 ± 0.6}          \\
                                          & {\color[HTML]{999999} BM25}             & {\color[HTML]{999999} 1799.2 ± 26.8} & {\color[HTML]{999999} 59.4 ± 0.3}          & {\color[HTML]{999999} 54.7 ± 0.2} & {\color[HTML]{999999} 66.7 ± 0.2}          & {\color[HTML]{999999} 30.3 ± 0.9} & {\color[HTML]{999999} 55.4 ± 1.2}          & {\color[HTML]{999999} 53.3 ± 0.5}          \\
                                          & \textbf{Zero-shot}                         & 87.6 ± 0.8                           & 0.0 ± 0.0                                  & 21.6 ± 0.2                        & 26.2 ± 0.2                                 & 0.0 ± 0.0                         & 3.1 ± 0.6                                  & 10.2 ± 0.1                                 \\
    \multirow{-4}{*}{\textbf{Llama}}      & \textbf{Ours}                           & 87.6 ± 0.8                           & \textbf{45.5 ± 1.5}                        & \textbf{45.2 ± 0.2}               & \textbf{57.6 ± 0.1}                        & \textbf{12.7 ± 0.9}               & \textbf{43.6 ± 2.0}                        & \textbf{40.9 ± 0.1}                        \\ \midrule
                                          & {\color[HTML]{B7B7B7} \textbf{16-shot}} & {\color[HTML]{B7B7B7} 1779.0 ± 3.4}  & {\color[HTML]{B7B7B7} \textbf{59.5 ± 1.5}} & {\color[HTML]{B7B7B7} 51.7 ± 0.7} & {\color[HTML]{B7B7B7} \textbf{69.4 ± 1.6}} & {\color[HTML]{B7B7B7} 24.0 ± 3.7} & {\color[HTML]{B7B7B7} \textbf{62.2 ± 1.8}} & {\color[HTML]{B7B7B7} \textbf{53.4 ± 1.3}} \\
                                          & {\color[HTML]{B7B7B7} \textbf{BM25}}    & {\color[HTML]{B7B7B7} 2045.0 ± 27.9} & {\color[HTML]{B7B7B7} 57.5 ± 1.3}          & {\color[HTML]{B7B7B7} 51.8 ± 0.5} & {\color[HTML]{B7B7B7} 66.2 ± 2.5}          & {\color[HTML]{B7B7B7} 23.8 ± 1.6} & {\color[HTML]{B7B7B7} 59.2 ± 1.3}          & {\color[HTML]{B7B7B7} 51.7 ± 0.6}          \\
                                          & \textbf{Zero-shot}                         & 100.9 ± 1.7                          & 0.1 ± 0.1                                  & 22.6 ± 0.8                        & 16.9 ± 0.9                                 & 0.7 ± 0.5                         & 2.7 ± 0.4                                  & 8.6 ± 0.1                                  \\
    \multirow{-4}{*}{\textbf{Mistral}}    & \textbf{Ours}                           & 100.9 ± 1.7                          & \textbf{31.8 ± 0.5}                        & \textbf{44.4 ± 0.6}               & \textbf{48.0 ± 0.7}                        & \textbf{17.0 ± 3.1}               & \textbf{41.8 ± 1.0}                        & \textbf{36.6 ± 0.1}                        \\ \midrule
                                          & {\color[HTML]{B7B7B7} \textbf{16-shot}} & {\color[HTML]{B7B7B7} 1581.1 ± 0.3}  & {\color[HTML]{B7B7B7} 52.7 ± 1.7}          & {\color[HTML]{B7B7B7} 21.7 ± 1.0} & {\color[HTML]{B7B7B7} 13.7 ± 1.2}          & {\color[HTML]{B7B7B7} 12.2 ± 1.5} & {\color[HTML]{B7B7B7} 34.1 ± 0.2}          & {\color[HTML]{B7B7B7} 26.8 ± 0.2}          \\
                                          & {\color[HTML]{B7B7B7} \textbf{BM25}}    & {\color[HTML]{B7B7B7} 1848.6 ± 26.4} & {\color[HTML]{B7B7B7} 47.9 ± 1.7}          & {\color[HTML]{B7B7B7} 20.8 ± 0.8} & {\color[HTML]{B7B7B7} 20.2 ± 0.8}          & {\color[HTML]{B7B7B7} 12.3 ± 2.8} & {\color[HTML]{B7B7B7} 36.4 ± 1.0}          & {\color[HTML]{B7B7B7} 27.5 ± 0.3}          \\
                                          & \textbf{Zero-shot}                      & 88.3 ± 1.5                           & 0.2 ± 0.0                                  & 7.7 ± 0.3                         & 3.2 ± 0.4                                  & 0.3 ± 0.2                         & 2.0 ± 0.0                                  & 2.7 ± 0.1                                  \\
    \multirow{-4}{*}{\textbf{Pythia}}     & \textbf{Ours}                           & 88.3 ± 1.5                           & \textbf{46.3 ± 0.8}                        & \textbf{23.2 ± 1.1}               & \textbf{11.1 ± 0.8}                        & \textbf{15.2 ± 2.0}               & \textbf{36.9 ± 3.0}                        & \textbf{26.5 ± 0.8}                        \\ \midrule
                                          & {\color[HTML]{B7B7B7} \textbf{16-shot}} & {\color[HTML]{B7B7B7} 1581.1 ± 1.3}  & {\color[HTML]{B7B7B7} 40.4 ± 1.1}          & {\color[HTML]{B7B7B7} 31.9 ± 1.1} & {\color[HTML]{B7B7B7} \textbf{34.2 ± 1.0}} & {\color[HTML]{B7B7B7} 15.0 ± 2.9} & {\color[HTML]{B7B7B7} 40.9 ± 2.1}          & {\color[HTML]{B7B7B7} 32.5 ± 0.3}          \\
                                          & {\color[HTML]{B7B7B7} \textbf{BM25}}    & {\color[HTML]{B7B7B7} 1848.6 ± 26.4} & {\color[HTML]{B7B7B7} 43.7 ± 1.9}          & {\color[HTML]{B7B7B7} 31.4 ± 0.0} & {\color[HTML]{B7B7B7} 25.2 ± 1.9}          & {\color[HTML]{B7B7B7} 14.7 ± 2.1} & {\color[HTML]{B7B7B7} 38.3 ± 0.8}          & {\color[HTML]{B7B7B7} 30.7 ± 0.5}          \\
                                          & \textbf{Zero-shot}                      & 88.3 ± 1.5                           & 0.2 ± 0.2                                  & 12.3 ± 0.2                        & 0.6 ± 0.2                                  & 0.0 ± 0.0                         & 5.1 ± 1.5                                  & 3.6 ± 0.3                                  \\
    \multirow{-4}{*}{\textbf{Mamba}}      & \textbf{Ours}                           & 88.3 ± 1.5                           & \textbf{33.7 ± 0.6}                        & \textbf{26.5 ± 0.6}               & \textbf{25.6 ± 1.1}                        & \textbf{16.3 ± 2.2}               & \textbf{39.2 ± 0.5}                        & \textbf{28.3 ± 0.2}                        \\ \bottomrule
    \end{tabular}%
    }
    \end{table}

    \begin{table}[H]
        \caption{Unseen task for ICL Setting}
        \label{tab:unseen_icl}
        \resizebox{\textwidth}{!}{%
        \begin{tabular}{ccccccccc}
        \toprule
                                           & \multicolumn{1}{c|}{}                                     & \multicolumn{1}{c|}{\textbf{\# Tokens}}                   & \textbf{GLUE COLA}                & \textbf{BBQ Religion}             & \textbf{Deepmind}                 & \textbf{MMLU-Psychology}          & \multicolumn{1}{c|}{\textbf{BBH-five-objects}}         & \textbf{Avg}                      \\ \midrule
                                           & \multicolumn{1}{c|}{{\color[HTML]{999999} \textbf{BM25}}} & \multicolumn{1}{c|}{{\color[HTML]{999999} 1684.3 ± 3.3}}  & {\color[HTML]{999999} 33.3 ± 1.9} & {\color[HTML]{999999} 69.3 ± 2.4} & {\color[HTML]{999999} 32.0 ± 1.4} & {\color[HTML]{999999} 83.7 ± 0.5} & \multicolumn{1}{c|}{{\color[HTML]{999999} 35.0 ± 0.0}} & {\color[HTML]{999999} 50.7 ± 0.1} \\
                                           & \multicolumn{1}{c|}{\textbf{Zero-shot}}                   & \multicolumn{1}{c|}{76.5 ± 0.2}                           & 0.7 ± 0.5                         & 5.3 ± 2.4                         & 0.0 ± 0.0                         & 62.7 ± 0.5                        & \multicolumn{1}{c|}{0.0 ± 0.0}                         & 13.7 ± 0.7                        \\
        \multirow{-3}{*}{\textbf{Llama}}   & \multicolumn{1}{c|}{\textbf{Ours}}                        & \multicolumn{1}{c|}{76.5 ± 0.2}                           & \textbf{1.3 ± 0.5}                & \textbf{24.3 ± 3.8}               & \textbf{19.3 ± 2.4}               & \textbf{66.0 ± 0.0}               & \multicolumn{1}{c|}{\textbf{5.0 ± 0.0}}                & \textbf{23.2 ± 1.1}               \\ \midrule
                                           & \multicolumn{1}{c|}{{\color[HTML]{999999} \textbf{BM25}}} & \multicolumn{1}{c|}{{\color[HTML]{999999} 1913.3 ± 20.0}} & {\color[HTML]{999999} 27.0 ± 1.6} & {\color[HTML]{999999} 69.0 ± 1.4} & {\color[HTML]{999999} 28.3 ± 3.9} & {\color[HTML]{999999} 79.7 ± 1.2} & \multicolumn{1}{c|}{{\color[HTML]{999999} 26.2 ± 0.0}} & {\color[HTML]{999999} 46.1 ± 0.6} \\
                                           & \multicolumn{1}{c|}{\textbf{Zero-shot}}                   & \multicolumn{1}{c|}{85.5 ± 0.8}                           & \textbf{1.7 ± 1.2}                & 1.0 ± 0.8                         & 1.3 ± 1.2                         & 35.3 ± 1.2                        & \multicolumn{1}{c|}{0.0 ± 0.0}                         & 7.9 ± 0.3                         \\
        \multirow{-3}{*}{\textbf{Mistral}} & \multicolumn{1}{c|}{\textbf{Ours}}                        & \multicolumn{1}{c|}{85.5 ± 0.8}                           & 1.3 ± 1.2                         & \textbf{18.3 ± 2.1}               & \textbf{20.0 ± 0.8}               & \textbf{54.7 ± 1.2}               & \multicolumn{1}{c|}{\textbf{10.0 ± 0.0}}               & \textbf{20.9 ± 1.1}               \\ \midrule
                                           
                                           & \multicolumn{1}{c|}{{\color[HTML]{999999} \textbf{BM25}}} & \multicolumn{1}{c|}{{\color[HTML]{999999} 1747.6 ± 20.2}} & {\color[HTML]{999999} 13.7 ± 1.9} & {\color[HTML]{999999} 33.7 ± 1.7} & {\color[HTML]{999999} 20.3 ± 2.6} & {\color[HTML]{999999} 18.7 ± 1.7} & \multicolumn{1}{c|}{{\color[HTML]{999999} 6.2 ± 0.0}}  & {\color[HTML]{999999} 18.5 ± 1.2} \\
                                           & \multicolumn{1}{c|}{\textbf{Zero-shot}}                   & \multicolumn{1}{c|}{78.1 ± 0.7}                           & \textbf{43.0 ± 2.9}               & 0.0 ± 0.0                         & 0.3 ± 0.5                         & 3.3 ± 0.5                         & \multicolumn{1}{c|}{0.0 ± 0.0}                         & 9.3 ± 0.7                         \\
        \multirow{-3}{*}{\textbf{pythia}}  & \multicolumn{1}{c|}{\textbf{Ours}}                        & \multicolumn{1}{c|}{78.1 ± 0.7}                           & 37.0 ± 2.4                        & \textbf{14.7 ± 4.6}               & \textbf{15.0 ± 0.8}               & \textbf{13.0 ± 1.4}               & \multicolumn{1}{c|}{\textbf{6.2 ± 0.0}}                & \textbf{17.2 ± 0.8}               \\ \midrule
                                            & \multicolumn{1}{c|}{{\color[HTML]{999999} \textbf{BM25}}} & \multicolumn{1}{c|}{{\color[HTML]{999999} 1747.6 ± 20.2}} & {\color[HTML]{999999} 36.7 ± 1.2} & {\color[HTML]{999999} 33.3 ± 2.6} & {\color[HTML]{999999} 25.3 ± 4.5} & {\color[HTML]{999999} 25.3 ± 0.5} & \multicolumn{1}{c|}{{\color[HTML]{999999} 21.2 ± 0.0}} & {\color[HTML]{999999} 28.4 ± 1.1} \\
                                           & \multicolumn{1}{c|}{\textbf{Zero-shot}}                   & \multicolumn{1}{c|}{78.1 ± 0.7}                           & \textbf{19.7 ± 2.5}               & 0.0 ± 0.0                         & 0.0 ± 0.0                         & 0.0 ± 0.0                         & \multicolumn{1}{c|}{0.0 ± 0.0}                         & 3.9 ± 0.5                         \\
        \multirow{-3}{*}{\textbf{Mamba}}   & \multicolumn{1}{c|}{\textbf{Ours}}                        & \multicolumn{1}{c|}{78.1 ± 0.7}                           & 18.0 ± 1.4                        & \textbf{19.3 ± 3.4}               & \textbf{13.7 ± 3.8}               & \textbf{13.0 ± 0.8}               & \multicolumn{1}{c|}{\textbf{7.5 ± 0.0}}                & \textbf{14.3 ± 1.1}               \\  \bottomrule
        \end{tabular}%
        }
        \end{table}

\subsection{Task Specific Prompt}
To ensure fair comparisons in zero-shot scenarios, we prepend task-specific prompts before each test query. These task-specific prompts are manually crafted following guidelines from \texttt{lm-harness} \citep{eval-harness}  and the \texttt{chain-of-thought-hub} \footnote{https://github.com/FranxYao/chain-of-thought-hub.git}.
The complete set of prompts used is provided in Figure ~\ref{tab:all_prompts}.
\begin{longtable}{|l|p{0.55\linewidth}|}
    \caption{Task Specific Prompts for Various Tasks} \label{tab:all_prompts} \small\\
    \hline
    \textbf{Task} & \textbf{Prompts} \\
    \hline
    \endfirsthead
    \multicolumn{2}{c}{\textbf{Prompts for Various Tasks (Continued)}} \\
    \hline
    \textbf{Task} & \textbf{Prompts} \\
    \hline
    \endhead
    \hline
    \multicolumn{2}{r}{\textit{Continued on next page}}
    \endfoot
    \endlastfoot

    bbh\_date\_understanding & \begin{itemize}[noitemsep,topsep=0pt]
        \item Infer the date from context. Finish your answer with 'X' where X is the correct letter choice.\newline\newline Question: \{input\}
        \item Determine the date based on contextual clues. End your response with 'X', where X represents the correct option.\newline\newline Question: \{input\}
        \item Use the given context to deduce the date. Conclude your answer with 'X', X being the right letter choice.\newline\newline Question: \{input\}
    \end{itemize} \\
    \hline

    bbh\_boolean\_expressions & \begin{itemize}[noitemsep,topsep=0pt]
        \item Evaluate the result of a random Boolean expression.\newline\newline Question: \{input\}
        \item Calculate the outcome of a given Boolean expression.\newline\newline Question: \{input\}
        \item Determine the result of the provided Boolean logic statement.\newline\newline Question: \{input\}
    \end{itemize} \\
    \hline

    bbh\_date\_understanding & \begin{itemize}[noitemsep,topsep=0pt]
        \item Infer the date from context. Finish your answer with 'X' where X is the correct letter choice.\newline\newline Question: \{input\}
        \item Determine the date based on contextual clues. End your response with 'X', where X represents the correct option.\newline\newline Question: \{input\}
        \item Use the given context to deduce the date. Conclude your answer with 'X', X being the right letter choice.\newline\newline Question: \{input\}
    \end{itemize} \\
    \hline
    bbh\_boolean\_expressions & \begin{itemize}[noitemsep,topsep=0pt]
        \item Evaluate the result of a random Boolean expression.\newline\newline Question: \{input\}
        \item Calculate the outcome of a given Boolean expression.\newline\newline Question: \{input\}
        \item Determine the result of the provided Boolean logic statement.\newline\newline Question: \{input\}
    \end{itemize} \\
    \hline
    mmlu\_pro\_math & \begin{itemize}[noitemsep,topsep=0pt]
        \item The following are multiple choice questions (with answers) about math. Finish your answer with 'X' where X is the correct letter choice.\newline\newline Question: \{input\}
        \item Below are multiple-choice math questions. Conclude your response with 'X', X being the correct option.\newline\newline Question: \{input\}
        \item Answer these math multiple-choice questions. Answer with 'X', where X is the right letter choice.\newline\newline Question: \{input\}
    \end{itemize} \\
    \hline
    superglue\_rte & \begin{itemize}[noitemsep,topsep=0pt]
        \item \{input\}\newline Question: Is the hypothesis correct based on the premise? Answer True or False.
        \item \{input\}\newline Question: Based on the premise, is the hypothesis accurate? Respond with True or False.
        \item \{input\}\newline Question: Does the premise support the hypothesis? Answer True or False.
    \end{itemize} \\
    \hline
    bbh\_temporal\_sequences & \begin{itemize}[noitemsep,topsep=0pt]
        \item Answer questions about which times certain events could have occurred. Finish your answer with 'X' where X is the correct letter choice.\newline\newline Q: \{input\}
        \item Determine possible occurrence times for specific events. Answer with 'X', X being the correct option.\newline\newline Q: \{input\}
        \item Identify when certain events could have happened. Conclude with 'X', where X is the right letter choice.\newline\newline Q: \{input\}
    \end{itemize} \\
    \hline
    bbh\_reasoning\_about\_colored\_objects & \begin{itemize}[noitemsep,topsep=0pt]
        \item Answer extremely simple questions about the colors of objects on a surface. Finish your answer with 'X' where X is the correct letter choice.\newline\newline Q: \{input\}
        \item Respond to basic questions about object colors on a surface. Answer with 'X', X being the correct option.\newline\newline Q: \{input\}
        \item Address simple queries regarding the colors of items on a surface. Conclude with 'X', where X is the right letter choice.\newline\newline Q: \{input\}
    \end{itemize} \\
    \hline
    ethics\_commonsense & \begin{itemize}[noitemsep,topsep=0pt]
        \item \{input\}\newline Question: Is this wrong? Answer Yes or No.
        \item \{input\}\newline Question: Is this ethically incorrect? Respond with Yes or No.
        \item \{input\}\newline Question: Does this violate ethical norms? Answer Yes or No.
    \end{itemize} \\
    \hline
    ethics\_justice & \begin{itemize}[noitemsep,topsep=0pt]
        \item \{input\}\newline Question: Would most people believe this reasonable or unreasonable to say? Answer Yes or No.
        \item \{input\}\newline Question: Would the majority consider this statement reasonable or unreasonable? Answer Yes or No.
        \item \{input\}\newline Question: Is this statement generally perceived as reasonable or unreasonable? Respond with Yes or No.
    \end{itemize} \\
    \hline
    glue\_mnli & \begin{itemize}[noitemsep,topsep=0pt]
        \item \{input\} True, False or Neither?
        \item \{input\} Answer selecting from: True, False, or Neither?
        \item \{input\} Choose one as answer: True, False, or Neither?
    \end{itemize} \\
    \hline
    glue\_sst2 & \begin{itemize}[noitemsep,topsep=0pt]
        \item \{input\}\newline Question: Is this sentence positive or negative? Answer positive or negative.
        \item \{input\}\newline Question: Does this sentence convey a positive or negative sentiment? Answer positive or negative.
        \item \{input\}\newline Question: Is the tone of this sentence positive or negative? Respond with positive or negative.
    \end{itemize} \\
    \hline
    glue\_qnli & \begin{itemize}[noitemsep,topsep=0pt]
        \item \{input\}\newline Question: Does this response answer the question? Answer Yes or No.
        \item \{input\}\newline Question: Is the response answer to the question? Answer Yes or No.
        \item \{input\}\newline Question: Does the given response address the question? Respond with Yes or No.
    \end{itemize} \\
    \hline
    hellaswag & \begin{itemize}[noitemsep,topsep=0pt]
        \item The following are multiple choice questions (with answers) about commonsense NLI. Finish your answer with 'X' where X is the correct letter choice.\newline\newline Question: \{input\}
        \item The following are multiple-choice questions about commonsense natural language inference. Answer with 'X', X being the correct option.\newline\newline Question: \{input\}
        \item Answer these multiple-choice questions on commonsense language understanding. Conclude with 'X', where X is the right letter choice.\newline\newline Question: \{input\}
    \end{itemize} \\
    \hline
    crows\_pairs & \begin{itemize}[noitemsep,topsep=0pt]
        \item \{input\}\newline Question: Which sentence is more stereotypical? Answer A or B.
        \item \{input\}\newline Question: Which sentence contains more stereotypical content? Answer A or B.
        \item \{input\}\newline Question: Identify the sentence with stronger stereotypical elements. Respond with A or B.
    \end{itemize} \\
    \hline
    boolq & \begin{itemize}[noitemsep,topsep=0pt]
        \item \{input\}\newline Answer True or False.
        \item \{input\}\newline Respond with True or False.
        \item \{input\}\newline Is this statement correct? Answer True or False.
    \end{itemize} \\
    \hline
    math\_qa & \begin{itemize}[noitemsep,topsep=0pt]
        \item The following are multiple choice questions (with answers) about math word problem. Finish your answer with 'X' where X is the correct letter choice.\newline\newline Question: \{input\}
        \item Below are multiple-choice questions about math word problems. Answer with 'X', X being the correct option.\newline\newline Question: \{input\}
        \item Solve these multiple-choice math word problems. Conclude with 'X', where X is the right letter choice.\newline\newline Question: \{input\}
    \end{itemize} \\
    \hline
    superglue\_wic & \begin{itemize}[noitemsep,topsep=0pt]
        \item \{input\}\newline Question: Is the word used in the same way in the two sentences above? Answer Yes or No.
        \item \{input\}\newline Question: Is the word used similarly in both sentences above? Respond with Yes or No.
        \item \{input\}\newline Question: Does the word have the same meaning in the two given sentences? Answer Yes or No.
    \end{itemize} \\
    \hline
    openbookqa & \begin{itemize}[noitemsep,topsep=0pt]
        \item The following are multiple choice questions (with answers) about multi-step reasoning. Finish your answer with 'X' where X is the correct letter choice.\newline\newline Question: \{input\}
        \item The following are multiple-choice questions testing multi-step reasoning. Answer with 'X', X being the correct option.\newline\newline Question: \{input\}
        \item Answer these multiple-choice questions involving multi-step logical thinking. Conclude with 'X', where X is the right letter choice.\newline\newline Question: \{input\}
    \end{itemize} \\
    \hline
    commonsense\_qa & \begin{itemize}[noitemsep,topsep=0pt]
        \item The following are multiple choice questions (with answers) about commonsense knowledge reasoning. Finish your answer with 'X' where X is the correct letter choice.\newline\newline Question: \{input\}
        \item Below are multiple-choice questions about commonsense reasoning. Answer with 'X', X being the correct option.\newline\newline Question: \{input\}
        \item Respond to these multiple-choice questions on commonsense knowledge. Conclude with 'X', where X is the right letter choice.\newline\newline Question: \{input\}
    \end{itemize} \\
    \hline
    bbq\_age & \begin{itemize}[noitemsep,topsep=0pt]
        \item The following are multiple choice questions (with answers) about social bias on age. Finish your answer with 'X' where X is the correct letter choice.\newline\newline Question: \{input\}
        \item The following are multiple-choice questions about age-related social bias. Answer with 'X', X being the correct option.\newline\newline Question: \{input\}
        \item Answer these multiple-choice questions on social bias regarding age. Conclude with 'X', where X is the right letter choice.\newline\newline Question: \{input\}
    \end{itemize} \\
    \hline
    arc\_challenge & \begin{itemize}[noitemsep,topsep=0pt]
        \item The following are multiple choice questions (with answers) about science. Finish your answer with 'X' where X is the correct letter choice.\newline\newline Question: \{input\}
        \item Below are multiple-choice science questions. Answer with 'X', X being the correct option.\newline\newline Question: \{input\}
        \item Respond to these multiple-choice questions on scientific topics. Conclude with 'X', where X is the right letter choice.\newline\newline Question: \{input\}
    \end{itemize} \\
    \hline  
    glue\_cola & \begin{itemize}[noitemsep,topsep=0pt]
        \item \{input\}\newline Question: Does this sentence make sense? Answer Yes or No.
        \item \{input\}\newline Question: Is this sentence logically coherent? Respond with Yes or No.
        \item \{input\}\newline Question: Evaluate if this sentence is meaningful. Reply with Yes or No.
    \end{itemize} \\
    \hline
    bbh\_logical\_deduction\_five\_objects & \begin{itemize}[noitemsep,topsep=0pt]
        \item A logical deduction task which requires deducing the order of a sequence of objects. Finish your answer with 'X' where X is the correct letter choice.\newline\newline Question: \{input\}
        \item This challenge involves logically determining the sequence of a set of objects. Conclude your response with 'X', where X is the appropriate letter option.\newline\newline Question: \{input\}
        \item In this logical reasoning exercise, deduce the correct order of a series of objects. End your answer with 'X', X being the right letter choice.\newline\newline Question: \{input\}
    \end{itemize} \\
    \hline
    mmlu\_high\_school\_psychology & \begin{itemize}[noitemsep,topsep=0pt]
        \item The following are multiple choice questions (with answers) about high school psychology. Finish your answer with 'X' where X is the correct letter choice.\newline\newline Question: \{input\}
        \item Below are multiple-choice questions testing high school level psychology knowledge. Conclude your response with 'X', X representing the correct option.\newline\newline Question: \{input\}
        \item These questions assess understanding of high school psychology concepts. End your answer with 'X', where X is the letter of the correct choice.\newline\newline Question: \{input\}
    \end{itemize} \\
    \hline
    bbq\_religion & \begin{itemize}[noitemsep,topsep=0pt]
        \item The following are multiple choice questions (with answers) about social bias on religion. Finish your answer with 'X' where X is the correct letter choice.\newline\newline Question: \{input\}
        \item Here are multiple-choice questions addressing social biases related to religion. Conclude your answer with 'X', X being the correct letter option.\newline\newline Question: \{input\}
        \item These questions explore social biases in the context of religion. End your response with 'X', where X represents the right letter choice.\newline\newline Question: \{input\}
    \end{itemize} \\
    \hline
    deepmind & \begin{itemize}[noitemsep,topsep=0pt]
        \item The following are multiple choice questions (with answers) about algebraic word problems. Finish your answer with 'X' where X is the correct letter choice.\newline\newline Question: \{input\}
        \item Below are multiple-choice questions testing algebraic word problem solving skills. Conclude your answer with 'X', X being the correct option letter.\newline\newline Question: \{input\}
        \item These questions assess your ability to solve algebraic word problems. End your response with 'X', where X is the letter of the right choice.\newline\newline Question: \{input\}
    \end{itemize} \\
    \hline
\end{longtable}

\section{Adaptive Elicitation} \label{appendix:adaptie_elicitation}
\begin{figure}[H]
    \centering
    \begin{subfigure}[b]{0.3\textwidth}
        \centering
        \includegraphics[width=\textwidth]{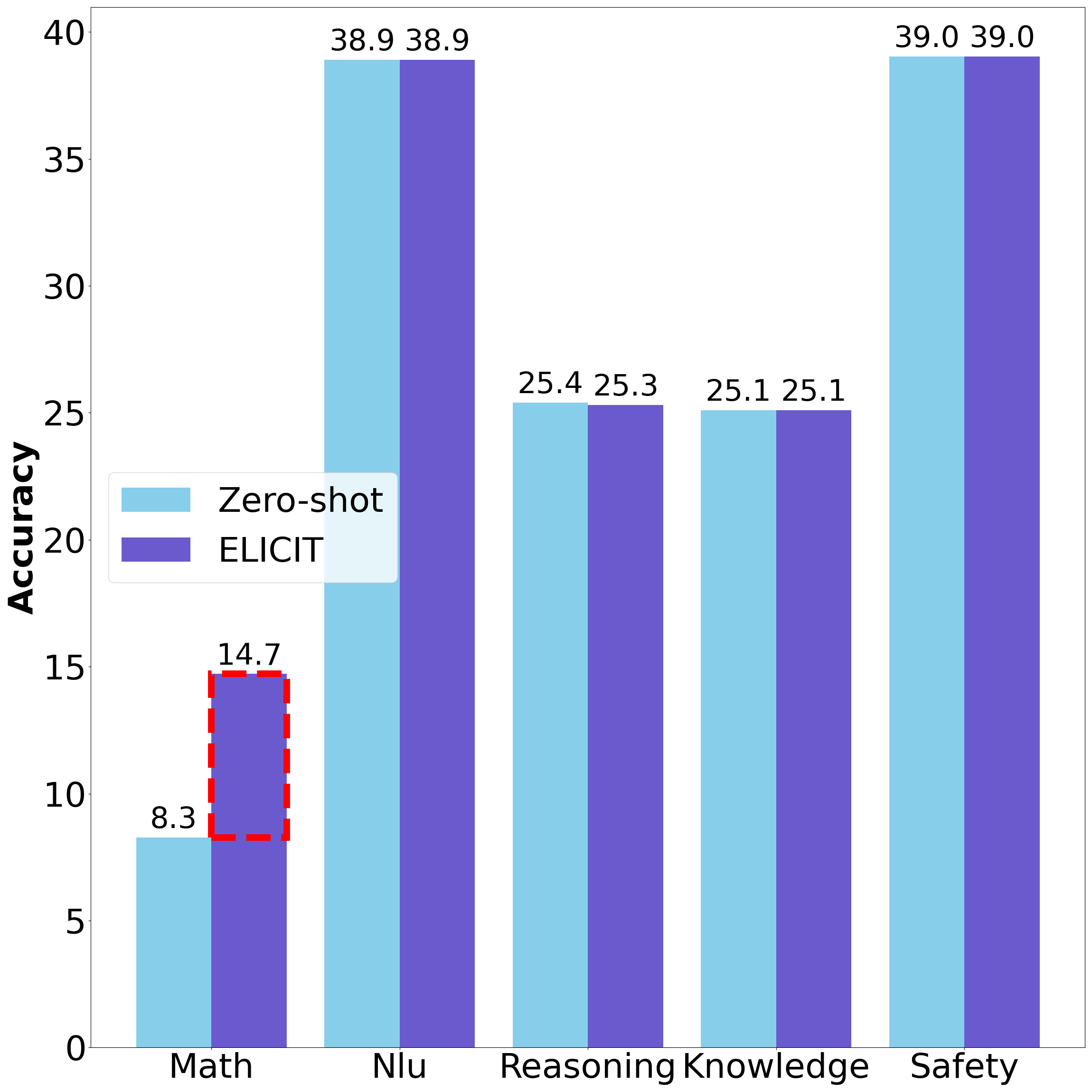}
        \caption{Mamba}
       
    \end{subfigure}
    \hfill
    \begin{subfigure}[b]{0.3\textwidth}
        \centering
        \includegraphics[width=\textwidth]{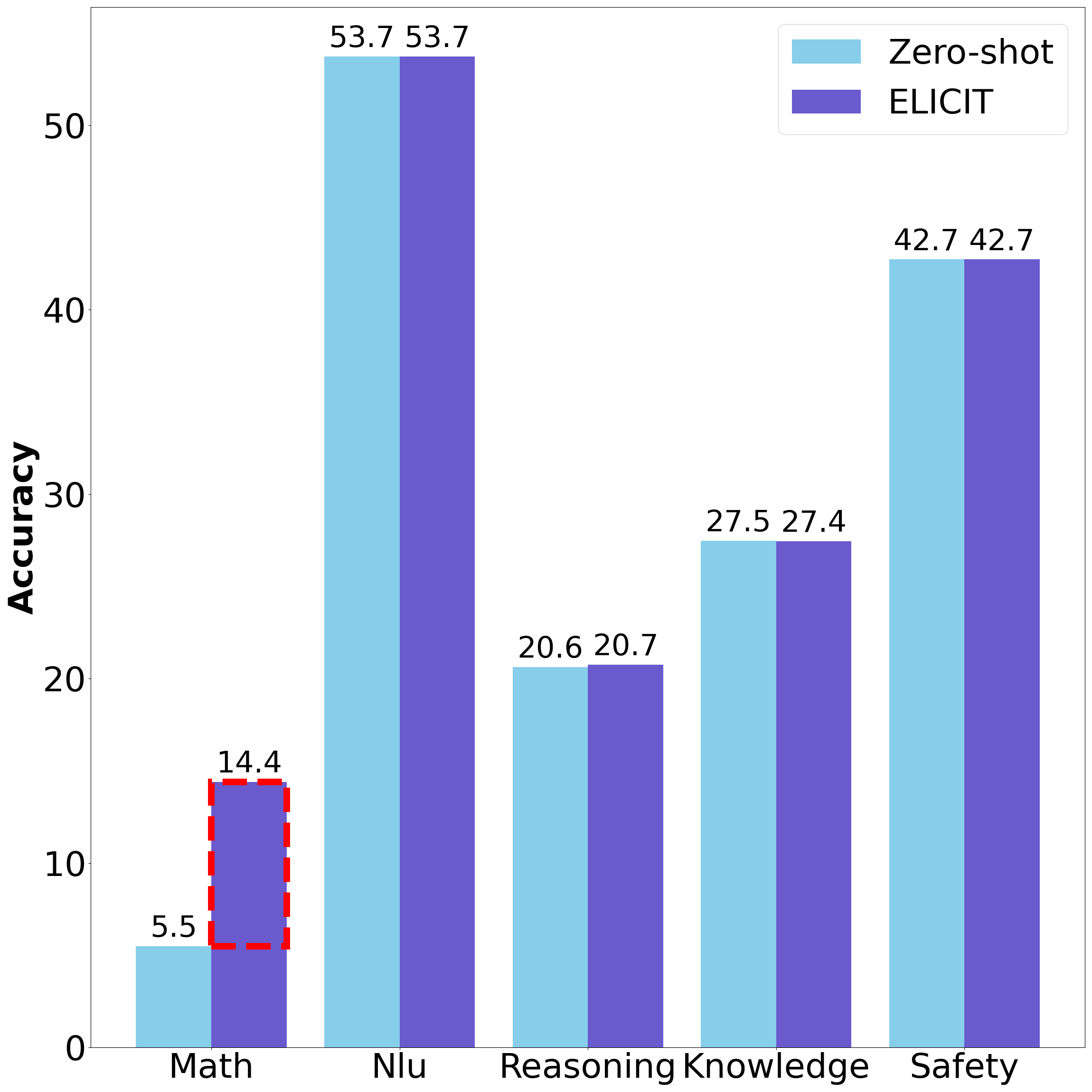}
        \caption{Pythia}
        
    \end{subfigure}
    \hfill
    \begin{subfigure}[b]{0.3\textwidth}
        \centering
        \includegraphics[width=\textwidth]{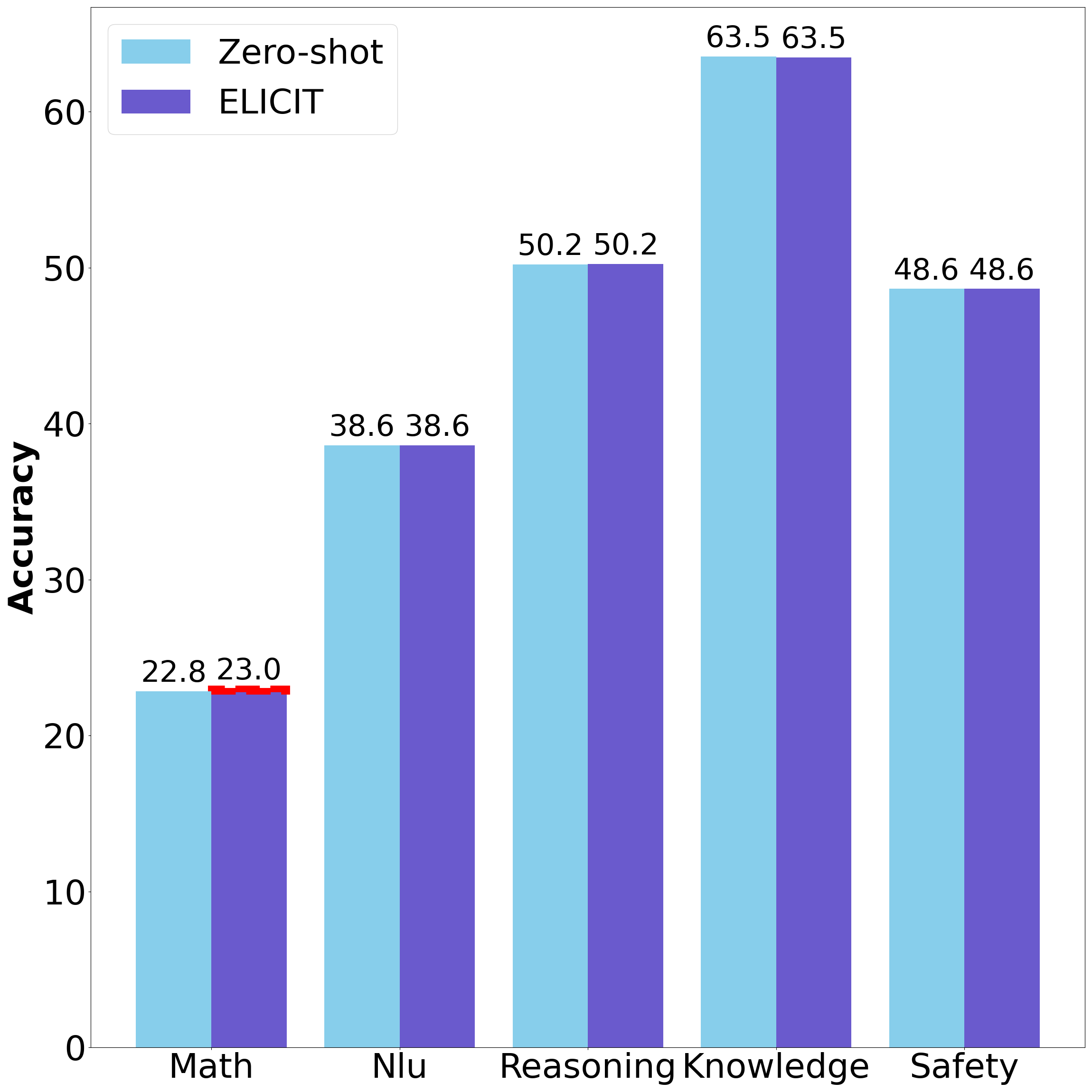}
        \caption{Llama3}
        
    \end{subfigure}
    \vspace{-1em}
    \caption{\small Performance on ELICITacross different domains when the li-brary only contains math-related taskvectors on Mamba, Pythia, and Llama3.}
    \label{fig:math_other_models}
\end{figure}

As mentioned in Section \ref{sec:adaptive_activation}, we show that when provided the library with only math-related task vectors, performance shows a significant improvement on the math domain while retaining or slightly improving in other domains for Mistral.
Figure~\ref{fig:math_other_models} illustrates similar results on other models such as Mamba, Pythia, and Llama3.

\section{Similarity-Based Retrieve Method} \label{appendix:similarity-based_retrieve_method}
\begin{figure}[H]
    
    \centering
    \subfloat[]{%
        \includegraphics[width=0.33\textwidth]{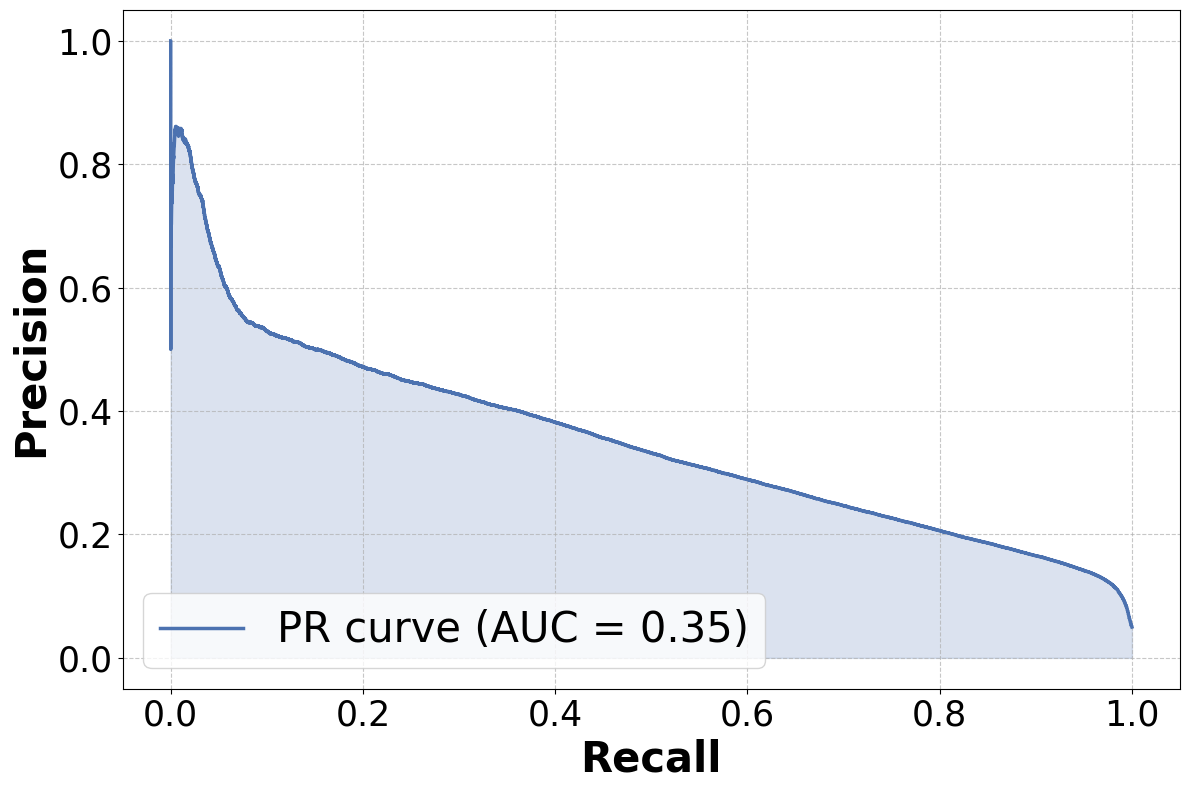}%
        \label{1}%
    }%
    \hfill
    \subfloat[]{%
        \includegraphics[width=0.33\textwidth]{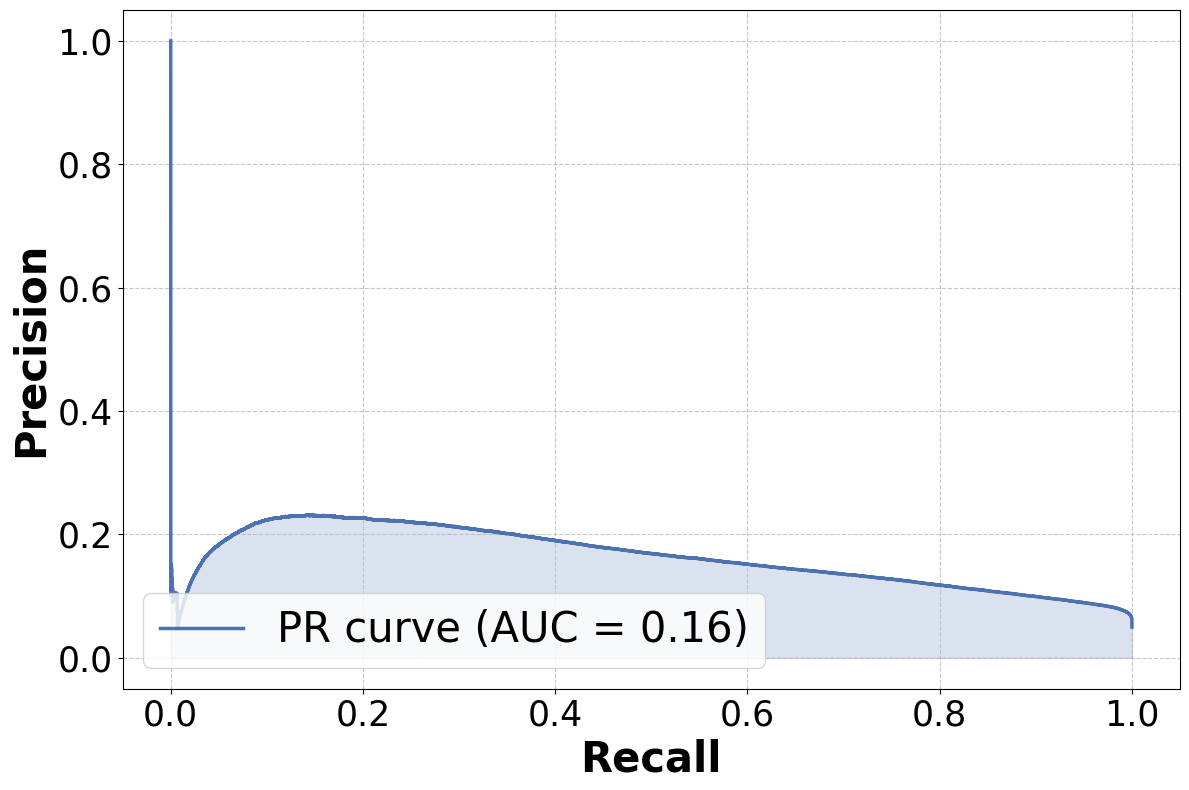}%
        \label{2}%
    }%
    \hfill
    \subfloat[]{%
        \includegraphics[width=0.33\textwidth]{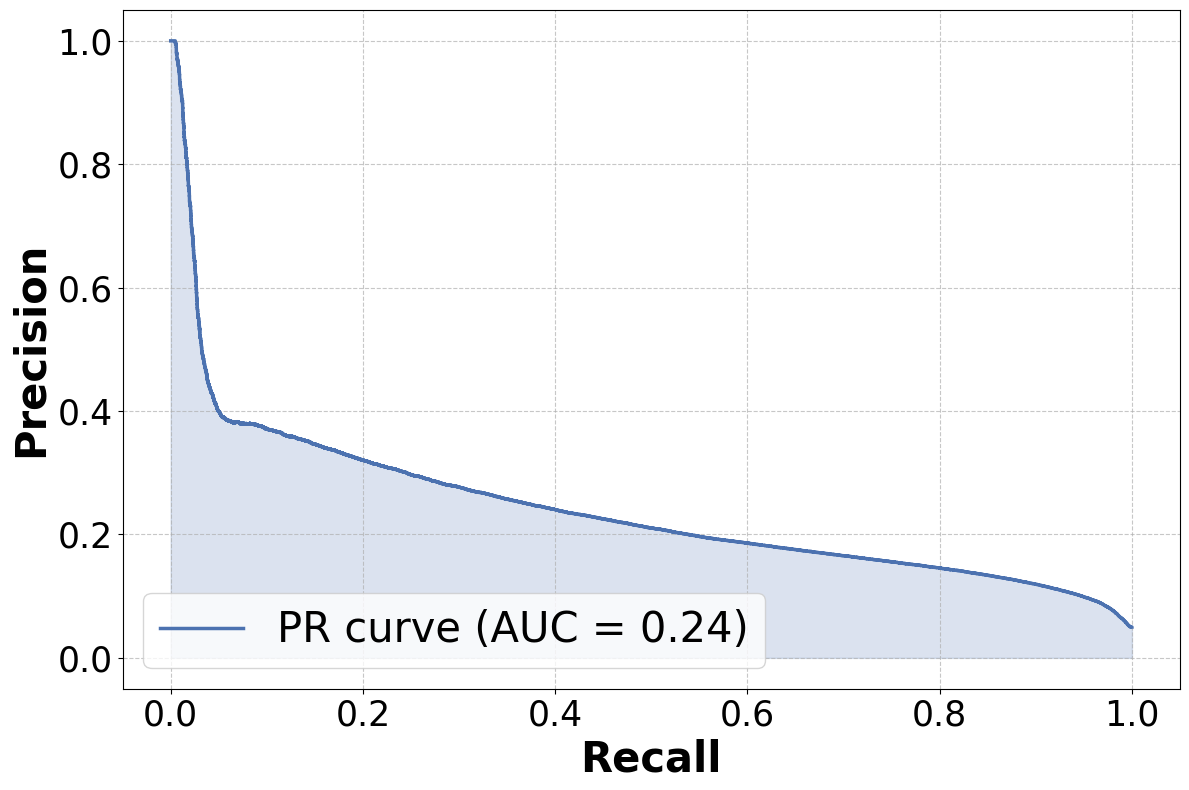}%
        \label{3}%
    }

    \vspace{1em}

    \subfloat[]{%
        \includegraphics[width=0.33\textwidth]{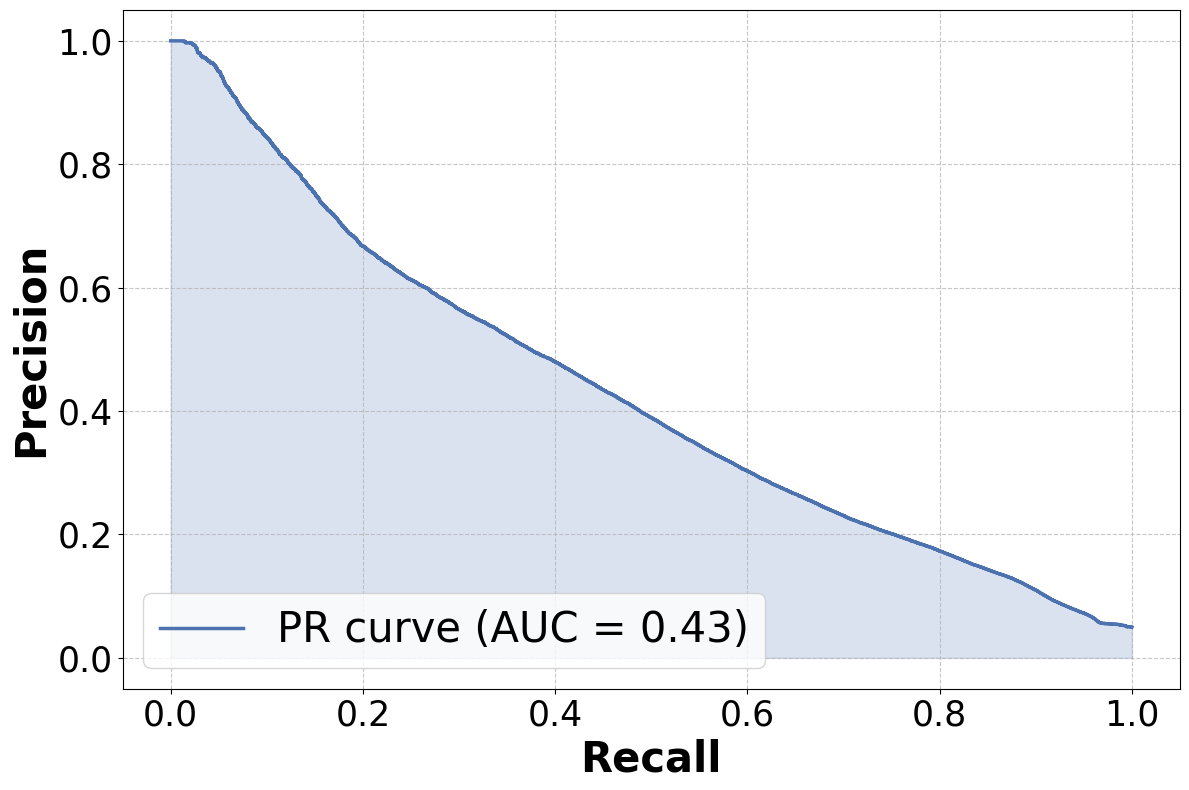}%
        \label{4}%
    }%
    \hfill
    \subfloat[]{%
        \includegraphics[width=0.33\textwidth]{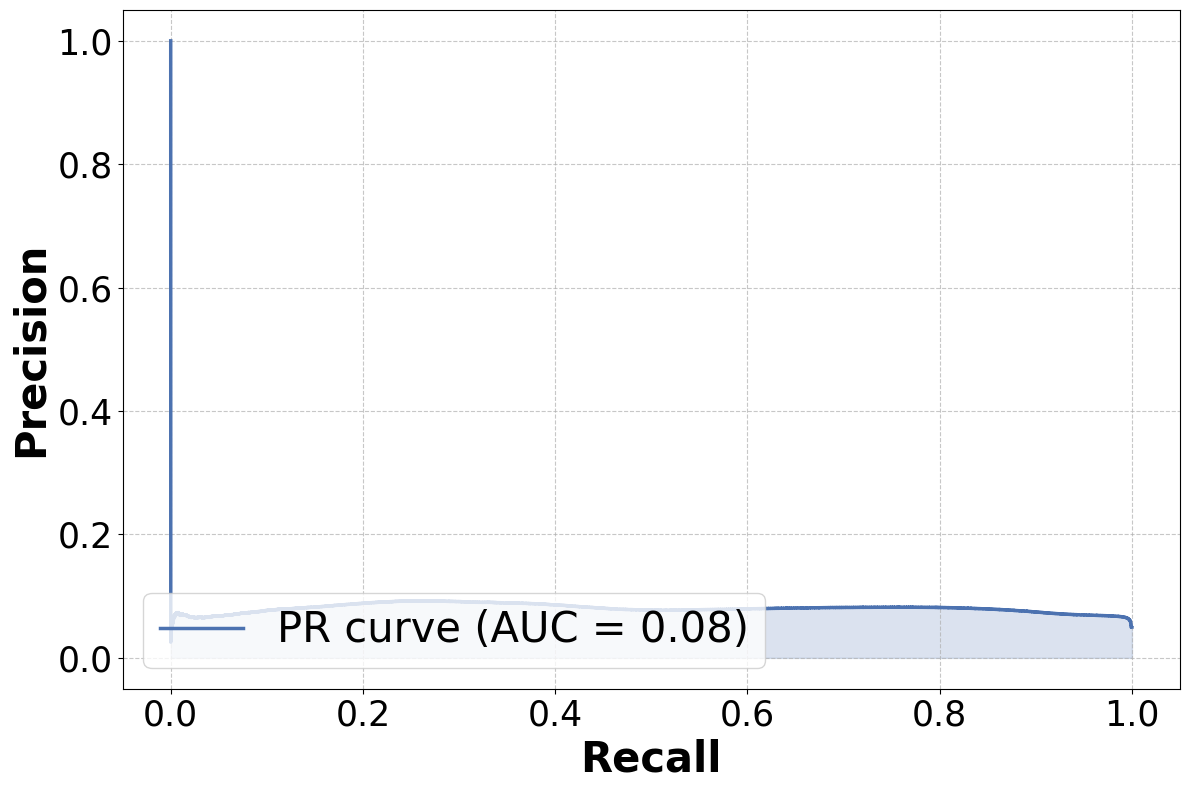}%
        \label{fig:5}%
    }%
    \hfill
    \subfloat[]{%
        \includegraphics[width=0.33\textwidth]{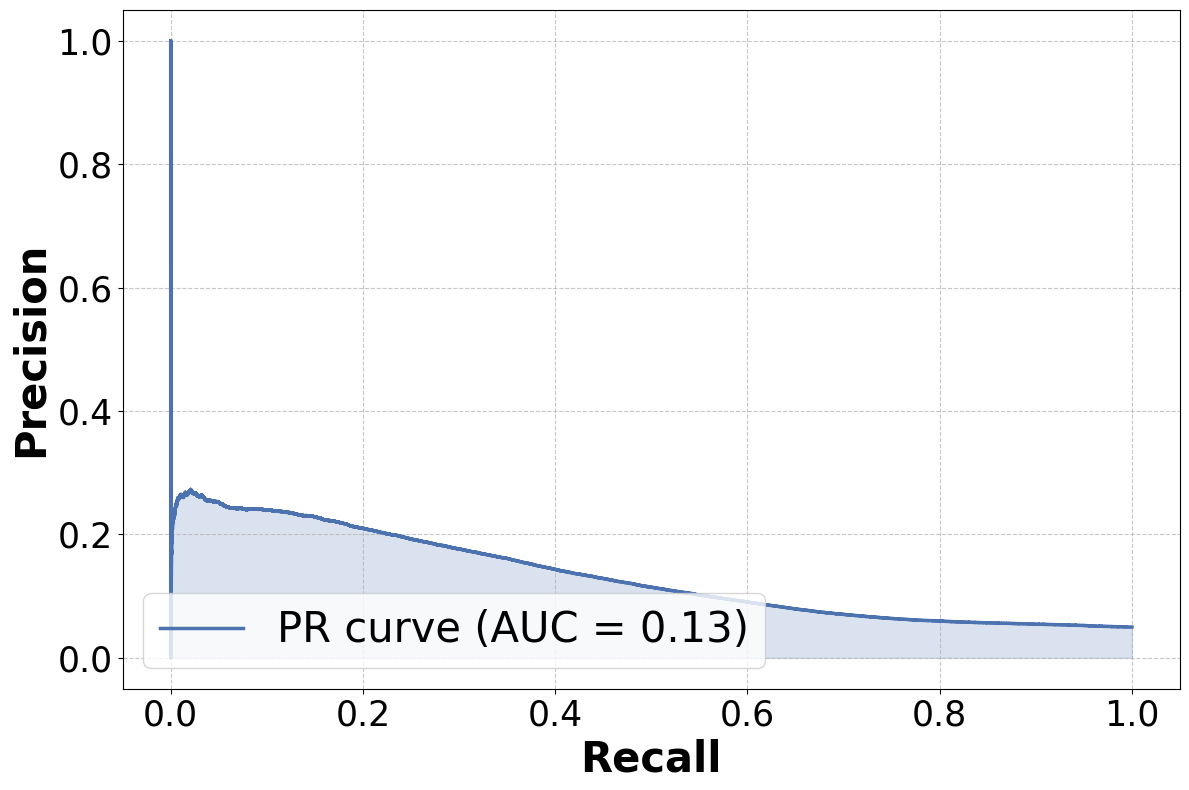}%
        \label{fig:6}%
    }

    \caption{Precision-Recall Curves for Similarity-based Prompt Retrieval Methods on Mistral and Mamba Models. Top Row: Mistral, Bottom Row: Mamba. From Left to Right: Cosine Similarity, t-SNE, and Distance-based Methods.}
    \label{fig:recall_precision_other_models_similarity_based}
\end{figure}

Section~\ref{sec:similarity-based_methods} demonstrates the poor precision-recall performance of similarity-based retrieval methods on the Llama3 model.
Figure~\ref{fig:recall_precision_other_models_similarity_based}  presents the Precision-Recall curves for Mistral and Mamba under different similarity-based approaches, which also exhibit poor results.
In contrast, our proposed retrieval module achieves significantly higher precision and recall across all models.
This highlights the effectiveness of our method in accurately retrieving relevant task vectors to support different tasks. 

\section{Orthogonal Augmentation in Unseen Tasks.} \label{appendix:orthogonal_augmentation}
As shown in Section~\ref{sec:orthogonal}, \alg demonstrates its plug-and-play capability by seamlessly integrating with existing methods such as BM25 Retrieval for in-domain tasks.
Table~\ref{tab:bm25+ours_unseen} showcases \alg combined with BM25 in unseen tasks, combined with BM25 for unseen tasks, where we can observe performance improvements across various models and tasks. 
This highlights the versatility and effectiveness of \alg in augmenting existing methods and tasks.
\begin{table}[H]
    \caption{\alg combined with BM25 on unseen tasks.
    Improvements are concentrated in smaller models (Pythia, Mamba), while larger models exhibit task-specific trade-offs (e.g., Llama3 shows gains in Deepmind but declines in GLUE-COLA).}
    \label{tab:bm25+ours_unseen}
    \resizebox{\columnwidth}{!}{%
    \begin{tabular}{lc|ccccc|c}
        \toprule
        {\color[HTML]{0D0D0D} \textbf{Model}} & \multicolumn{1}{l|}{} & \textbf{GLUE COLA}  & \textbf{BBQ Religion} & \textbf{Deepmind}   & \textbf{MMLU-Psychology} & \textbf{BBH-five-objects} & \textbf{Avg}        \\ \midrule
        \multicolumn{1}{c}{\textbf{Llama}}    & \textbf{BM25}         & \textbf{55.4 ± 1.0} & \textbf{64.6 ± 1.3}   & \textbf{30.7 ± 1.7} & \textbf{83.0 ± 0.1}      & \textbf{48.3 ± 0.0}       & \textbf{56.4 ± 0.4} \\
                                              & \textbf{BM25+ELICIT}  & 47.6 ± 2.2          & 60.6 ± 1.0            & 26.4 ± 1.0          & 81.4 ± 0.8               & 44.4 ± 0.0                & 52.1 ± 0.3          \\ \midrule
        \multicolumn{1}{c}{\textbf{Mistral}}  & \textbf{BM25}         & \textbf{44.4 ± 2.2} & \textbf{70.7 ± 0.7}   & \textbf{26.6 ± 3.9} & \textbf{78.7 ± 1.1}      & 25.7 ± 0.0                & \textbf{49.2 ± 0.3} \\
                                              & \textbf{BM25+ELICIT}  & 36.4 ± 1.1          & 59.4 ± 1.8            & 25.2 ± 1.6          & 70.5 ± 0.3               & \textbf{26.9 ± 0.0}       & 43.7 ± 0.5          \\ \midrule
        \multicolumn{1}{c}{\textbf{Pythia}}   & \textbf{BM25}         & 5.8 ± 1.0           & 19.1 ± 1.2            & \textbf{14.1 ± 1.2} & 4.7 ± 0.3                & 1.0 ± 0.0                 & 8.9 ± 0.3           \\
                                              & \textbf{BM25+ELICIT}  & \textbf{7.3 ± 0.8}  & \textbf{30.9 ± 3.3}   & 14.0 ± 0.6          & \textbf{11.9 ± 0.6}      & \textbf{3.5 ± 0.0}        & \textbf{13.5 ± 0.7} \\ \midrule
        \multicolumn{1}{c}{\textbf{Mamba}}    & \textbf{BM25}         & \textbf{48.1 ± 3.1} & 30.6 ± 1.1            & 21.6 ± 3.3          & 19.1 ± 0.9               & \textbf{25.8 ± 0.0}       & \textbf{29.0 ± 0.9} \\
                                              & \textbf{BM25+ELICIT}  & 46.6 ± 1.7          & \textbf{30.9 ± 1.8}   & \textbf{22.7 ± 0.6} & \textbf{22.7 ± 0.4}      & 21.8 ± 0.0                & 28.9 ± 0.5          \\ \bottomrule
        \end{tabular}%
    }
\end{table}

\section{Dataset Splits}

We provide detailed information about our dataset curation and splitting strategies to ensure reproducibility.
Our primary objective was to maintain robust evaluation capabilities while ensuring sufficient training data for ICL prompt construction.
For datasets with pre-existing splits (ARC-Challenge, Ethics, GLUE, MathQA, OpenbookQA), we preserved the original partitioning.
When handling datasets with only train-valid splits, we employed two approaches: for those with validation sets exceeding 350 samples (e.g., BoolQ, Hellaswag), we split the validation set into new validation and test sets at a 7:3 ratio;
for those with smaller validation sets (e.g., CommonsenseQA), we divided the training set into new train and test sets (7:3).
For test-only datasets, we implemented different strategies based on size: smaller datasets like BBH (250 samples) were split to ensure 128 samples for training and 80-100 samples for testing, with remaining samples allocated to validation.
Larger test-only datasets (>1000 samples) such as MMLU-Pro-Math, BBQ, and Crows Pairs were split into train-valid-test sets at a 7:2:1 ratio.
The same 7:2:1 split was applied to train-only datasets like SuperGLUE and DeepMind.
This systematic approach ensures a minimum of 80 test samples for reliable evaluation metrics and at least 128 training samples for ICL prompt construction across all tasks.

\section{Analysis of \alg 's Selective Activation} \label{appendix:analysis_selective_activation}

We investigate why ELICIT can selectively activate capability in Figure~\ref{fig:selective} and the importance of this mechanism.
Using a library containing only math-related task vectors on Mistral, we analyzed the number of chosen states per domain, shown in Table~\ref{table:chosen_nums}.
Math-related tasks showed consistent high utilization (9.8 ± 0.1 chosen states), while other domains maintained minimal selection (approximately 0.0).
This pattern confirms that ELICIT's performance improvements stem from its dynamic retrieval and selective activation of relevant capabilities.

\vspace{10pt}
\begin{table}[h]
    \caption{The average number of chosen numbers per domain per sample. The statistics come from Mistral when the capability library only contains math-related task vectors.}
    \label{table:chosen_nums}
    \resizebox{\columnwidth}{!}{%
    \begin{tabular}{cccccc}
    \toprule
    \multicolumn{6}{c}{{\textbf{in-domain}}}                                                                                                                                                                                                                                                \\ \midrule
    \multicolumn{1}{c|}{{}}                     & {\textbf{NLU}}       & {\textbf{Reasoning}}    & {\textbf{Knowledge}} & {\textbf{Math}}            & {\textbf{Safety}}           \\
    \multicolumn{1}{c|}{{\textbf{chosen nums}}} & {0.0 ± 0.0}          & {0.1 ± 0.0}             & {0.0 ± 0.0}          & {\textbf{9.8 ± 0.1}}       & {0.0 ± 0.0}                 \\ \midrule
    \multicolumn{6}{c}{{\textbf{Out-of-domain}}}                                                                                                                                                                                                                                                     \\ \midrule
    \multicolumn{1}{c|}{{}}                     & {\textbf{GLUE COLA}} & {\textbf{BBQ Religion}} & {\textbf{Deepmind}}  & {\textbf{MMLU-Psychology}} & {\textbf{BBH-five-objects}} \\
    \multicolumn{1}{c|}{{\textbf{chosen nums}}} & {0.0 ± 0.0}          & {0.0 ± 0.0}             & {\textbf{9.9 ± 0.1}} & {0.0 ± 0.0}                & {0.0 ± 0.0}                 \\ \bottomrule
    \end{tabular}%
    }
    \end{table}

We observed minor improvements in reasoning tasks, exemplified by this ARC Challenge case in Table~\ref{fig:case_arc_challenge}.
It demonstrates our pipeline's ability to selectively activate relevant capabilities based solely on query and handle unseen tasks flexibly, without requiring explicit task information.

\vspace{10pt}
\begin{table}[h]
    \caption{A successful case from Arc-Challenge when capability library only contains math-related task vectors on Mistral.}
    \label{fig:case_arc_challenge}
    \small
    \centering
    \begin{tabular}{l|l}
    \toprule
    \multicolumn{1}{c|}{{\textbf{input}}} & {\begin{tabular}[c]{@{}l@{}}Below are multiple-choice science questions.Answer with 'X',\\  X being the correct option.\textbackslash{}n\textbackslash{}nQuestion: An unbalanced \\ equation for the reaction of methane gas (CH\_\{4\}) with oxygen \\ is shown below. CH\_\{4\} + \textbackslash{}\textbackslash{}Box O\_\{2\} -\textgreater 2CO\_\{2\} + 4H\_\{2\}O \\ How many molecules of oxygen gas (O\_\{2\}) are needed to \\ properly balance this equation?\textbackslash{}nOptions:\textbackslash{}nA. 1\textbackslash{}nB. 2\textbackslash{}nC. 3\\ \textbackslash{}nD. 4\textbackslash{}nAnswer:\end{tabular}} \\ \midrule
    {\textbf{chosen task vectors}}         & {10 task vectors from MathQA}                                                                                                                   \\ \midrule
    {\textbf{Original Output}}             & {B}                                                                                                                                 \\ \midrule
    {\textbf{ELICIT Output}}               & {\textbf{D (correct)}}                                                                                                                                                        \\ \bottomrule
    \end{tabular}%
\end{table}

Experiments forcing the application of top task vectors to all queries (Table~\ref{table:force_application}), showed significant performance degradation in NLU and knowledge tasks, highlighting the importance of selective activation.

\vspace{10pt}
\begin{table}[h]
    \caption{The results of forcibly applying the top task vectors for each query. The experiments were conducted on Mistral. Domains with degraded performance are marked in bold.}
    \label{table:force_application}
    \centering
    \begin{tabular}{l|lllll}
    \toprule
    \multicolumn{1}{c|}{{\textbf{}}} & \multicolumn{1}{c}{{\textbf{nlu}}} & \multicolumn{1}{c}{{\textbf{reasoning}}} & \multicolumn{1}{c}{{\textbf{knowledge}}} & \multicolumn{1}{c}{{\textbf{math}}} & \multicolumn{1}{c}{{\textbf{safety}}} \\ \midrule
    {\textbf{Zero-shot}}             & {28.8}                              & {27.4}                                    & {58.8}                                    & {4.0}                                & {42.2}                                \\ \midrule
    {\textbf{ELICIT}}                & {\textbf{15.7}}                              & {31.4}                                    & {\textbf{47.8}}                                    & {18.3}                               & {53.1}                                \\ \bottomrule
    \end{tabular}%
\end{table}

These experimental results demonstrate that ELICIT's performance improvement stems from its selective activation mechanism and the importance of selectively using only task-relevant vectors to dynamically activate capabilities.

We analyzed the usage frequency of task vectors in the capability library, which contains 20 distinct task vector types. The analysis was performed on Pythia-6.9B while evaluating 25 tasks in total: 20 in-domain tasks and 5 out-of-domain tasks. Our findings confirmed that all 20 task vector types in the library were utilized during the evaluation.

\begin{figure}[H]
    \centering 
    \includegraphics[width=01.0\textwidth,]{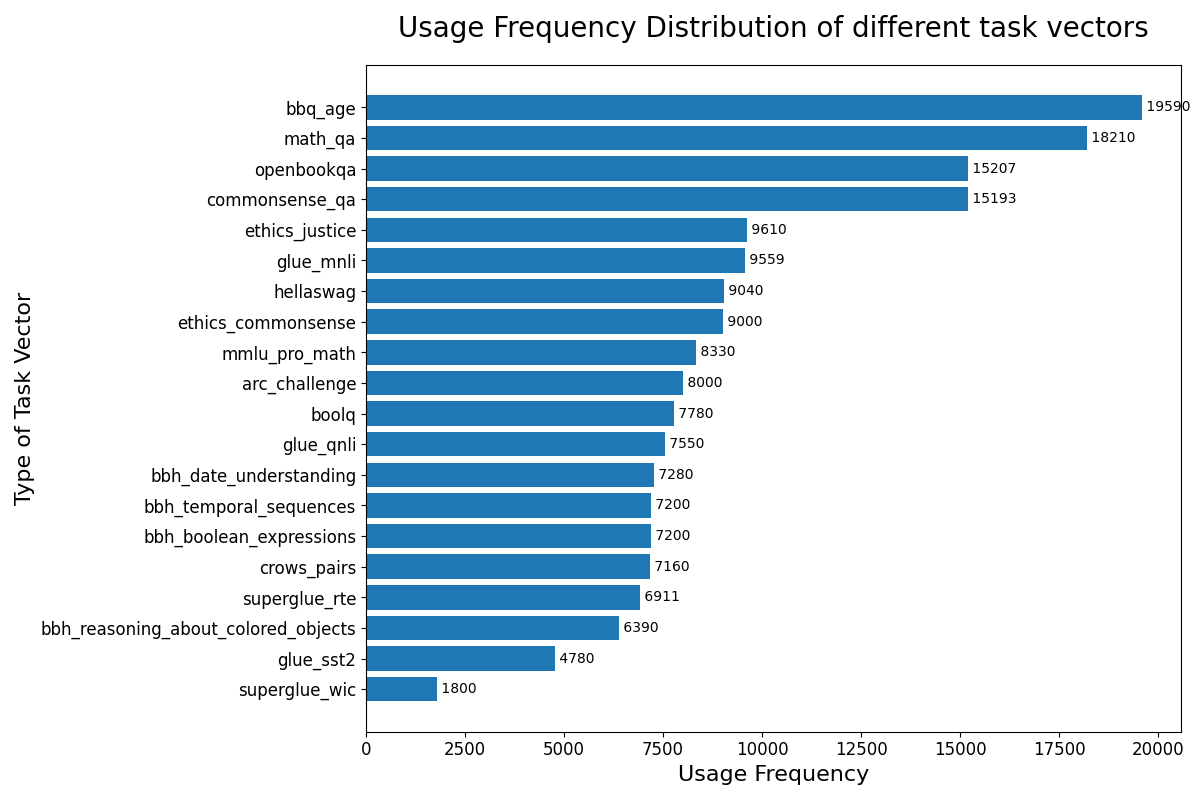}
    \caption{\small Usage Frequency Distribution of Different Types of Task Vectors Across all In-Domain and OOD (Out-of-Domain) Samples. The results is based on Pythia-6.9B.}
    \label{fig:usage_distribution}
\end{figure}

\section{More models and tasks on \alg}

Beyond our primary experiments, we evaluate the scalability and generalizability of \alg across larger language models and more challenging tasks.
As shown in Table~\ref{table:more_models}, \alg maintains its performance advantages when applied to more base models.
Furthermore, Table~\ref{table:more_datasets} demonstrates that \alg achieves consistent improvements across a diverse set of complex tasks, validating its effectiveness and versatility.

We further explore the applicability of \alg to instruction-tuned models, with preliminary results shown in Table~\ref{table:instrut_model}.
While this initial experiment suggest the potential compatibility of \alg with instruction-tuned models, several challenges remain.
Instruction-tuned models exhibit heightened sensitivity to prompts and instructions \citep{sun2023evaluating,gao2023roles}, necessitating more investigation and analysis.
Key challenges include identifying effective task vectors for in-context learning (ICL) and developing robust methods for zero-shot performance evaluation.
We leave the comprehensive adaptation of \alg for instruction-tuned models as promising future work.

\vspace{10pt}
\begin{table}[h]
    \caption{Performance of \alg on more different models. \alg are effective for larger models.}
    \label{table:more_models}
    \resizebox{\columnwidth}{!}{%
    \begin{tabular}{cc|cccccc|c}
        \toprule
        \multicolumn{1}{l}{{}}                                        & \multicolumn{1}{l|}{{}} & {\textbf{Length}} & {\textbf{nlu}}        & {\textbf{reasoning}}  & {\textbf{knowledge}}  & {\textbf{math}}       & {\textbf{safety}}     & {\textbf{avg}}        \\ \midrule
        \multicolumn{1}{c|}{{}}                                       & {\textbf{zs}}           & {109.8 ± 1.5}     & {37.6 ± 0.4}          & {16.1 ± 0.5}          & {17.4 ± 0.6}          & {5.9 ± 0.7}           & {31.7 ± 0.5}          & {21.8 ± 0.1}          \\
        \multicolumn{1}{c|}{\multirow{-2}{*}{{\textbf{Pythia-6.9B}}}} & {\textbf{ELICIT}}       & {109.8 ± 1.5}     & {\textbf{38.7 ± 1.4}} & {\textbf{28.1 ± 0.5}} & {\textbf{27.9 ± 1.0}} & {\textbf{18.2 ± 2.6}} & {\textbf{47.8 ± 2.0}} & {\textbf{32.2 ± 0.7}} \\ \midrule
        \multicolumn{1}{c|}{{}}                                       & {\textbf{zs}}           & {109.8 ± 1.5}     & {34.7 ± 0.6}          & {20.7 ± 0.2}          & {18.1 ± 0.6}          & {7.9 ± 1.7}           & {34.6 ± 0.6}          & {23.2 ± 0.2}          \\
        \multicolumn{1}{c|}{\multirow{-2}{*}{{\textbf{Pythia-12B}}}}  & {\textbf{ELICIT}}       & {109.8 ± 1.5}     & {\textbf{38.5 ± 0.5}} & {\textbf{29.7 ± 0.7}} & {\textbf{29.8 ± 0.6}} & {\textbf{17.5 ± 2.1}} & {\textbf{46.8 ± 0.2}} & {\textbf{32.5 ± 0.5}} \\ \midrule
        \multicolumn{1}{c|}{{}}                                       & {\textbf{zs}}           & {101.1}           & {50.9}                & {66.8}                & {59.7}                & {37.6}                & {44.2}                & {51.8}                \\
        \multicolumn{1}{c|}{\multirow{-2}{*}{{\textbf{Llama3-70B}}}}  & {\textbf{ELICIT}}       & {101.1}           & {\textbf{55.9}}       & {\textbf{80.5}}       & {\textbf{84.6}}       & {\textbf{52.4}}       & {\textbf{67.4}}       & {\textbf{68.2}}       \\ \bottomrule
        \end{tabular}%
    }
    \end{table}

    \vspace{10pt}
    \begin{table}[h]
        \caption{The results of \alg on GSM8K and MMLU-Professional-Law on Llama3-8B. GSM8K is as in-domain task and MMLU-Profeesional-Law is out-of-domain.}
        \label{table:more_datasets}
        \centering
        \begin{tabular}{c|cc}
        \toprule
        {\textbf{}}             & {\textbf{GSM8K}} & {\textbf{MMLU-Professional-Law}} \\ \midrule
        {\textbf{zs}}           & {30.44}          & {31.67}                          \\
        {\textbf{ELICIT}}       & {\textbf{32.44}} & {\textbf{41.11}}                 \\
        \bottomrule
        \end{tabular}%
        \end{table}

\vspace{10pt}

\begin{table}[h]
    \caption{
        The preliminary experiment of ELICIT on Llama3-8B-Instruct.    
    }
    \label{table:instrut_model}
    \centering
        \begin{tabular}{c|cccc|c}
        \toprule
        {\textbf{}}     & {\textbf{nlu}}  & {\textbf{reasoning}} & {\textbf{knowledge}} & {\textbf{safety}} & {\textbf{avg}}  \\ \midrule
        {\textbf{zs}}   & {45.0}          & {4.9}                & {31.9}               & {42.5}            & {31.1}          \\
        {\textbf{\alg}} & {\textbf{52.7}} & {\textbf{36.2}}      & {\textbf{70.9}}      & {\textbf{49.0}}   & {\textbf{52.2}} \\ \bottomrule
        \end{tabular}%
        
        \end{table}

\section{Diversity-Optimizaed Capability Library}

We conduct an experiment on maximizing the diversity of prompts in the given capability library. Instead of random demonstration selection, we construct a new capability library of diversity-optimized prompts as described in \cite{su2022selective}.

Spefically, we used Sentence-BERT to generate embeddings by averaging the resulting vectors over the words in each text input. For each task, after computing embeddings for all training data, we implemented an iterative approach to find diverse examples to construct ICL prompts. Starting with a random example, we selected examples that maximized the distance from previously chosen examples in each iteration. We then conducted a new capability library using these more diverse ICL prompts.

As shown in Table~\ref{table:diversity_results}, the diversity-optimized prompts yielded mixed results. Compared to the original ELICIT, while performance improved in reasoning (+1.1\%), math (+0.5\%) and NLU tasks (+4.5\%), there was a decline in Knowledge (-5.9\%) and Safety (-2.3\%) ability.

This result suggests the potential for future work to improve our pipeline by enhancing the quality of task vectors through better demonstration selection methods.

\begin{table}[h]
    \caption{
        The comparison of ELICIT using different capability library based on different ICL prompts. The experiments are conducted on Llama3-8B.
    }
    \label{table:diversity_results}
    \resizebox{\columnwidth}{!}{%
    \begin{tabular}{c|ccccc|c}
    \toprule
    \multicolumn{1}{c|}{{\textbf{}}} & \multicolumn{1}{c}{{\textbf{NLU}}} & \multicolumn{1}{c}{{\textbf{Reasoning}}} & \multicolumn{1}{c}{{\textbf{Knowledge}}} & \multicolumn{1}{c}{{\textbf{Math}}} & \multicolumn{1}{c|}{{\textbf{Safety}}} & \multicolumn{1}{c}{{\textbf{Avg.}}} \\ \midrule
    {\textbf{Zero-shot}}             & {32.2 ± 1.2}                       & {32.9 ± 0.2}                             & {42.5 ± 1.2}                             & {14.0 ± 1.0}                        & {35.5 ± 1.2}                           & {31.4 ± 0.7}                        \\
    {\textbf{ELICIT}}                & {38.1 ± 0.9}             & {46.1 ± 0.3}                    & {\textbf{60.7 ± 1.2}}                    & {19.4 ± 1.1}               & {\textbf{49.4 ± 2.1}}                  & {\textbf{42.7 ± 0.8}}               \\
    {\textbf{ELICIT (diversity)}}    & {\textbf{42.6 ± 0.3}}              & {\textbf{47.2 ± 0.1}}                    & {54.8 ± 1.5}                             & {\textbf{19.9 ± 0.8}}               & {47.1 ± 2.6}                           & {42.3 ± 0.9}                        \\ \bottomrule
    \end{tabular}%
    }
    \end{table}

\section{Multi-Layer Intervention}

While our primary analysis focuses on single-layer intervention, we also conduct preliminary experiments on multi-layer intervention, with the intervention strength $\alpha=2$ distributed evenly across layers.
We evaluated four settings: (1) the zero-shot baseline, (2) intervention on three consecutive layers (centered on the previously identified optimal layer), (3) intervention across all layers, and (4) our original single-layer implementation.

Results from Llama3-8B (Table~\ref{table:multi_layers}) reveal an intriguing pattern: distributing intervention across multiple layers tends to yield better performance.
This observation opens promising directions for future research into the mechanisms and benefits of multi-layer interventions.

\vspace{10pt}
\begin{table}[H]
    \caption{
            Comparison of multiple intervention layers on ELICIT. The experiments are conducted on Llama3-8B.
        }
    \label{table:multi_layers}
    \centering
    \begin{tabular}{l|lllll|l}
    \toprule
    \multicolumn{1}{c|}{{ \textbf{}}} & \multicolumn{1}{c}{{ \textbf{nlu}}} & \multicolumn{1}{c}{{ \textbf{reasoning}}} & \multicolumn{1}{c}{{ \textbf{knowledge}}} & \multicolumn{1}{c}{{ \textbf{math}}} & \multicolumn{1}{c|}{{ \textbf{safety}}} & \multicolumn{1}{c}{{ \textbf{avg}}} \\ \midrule
    { \textbf{zs}}                    & { 32.4}                             & { 31.8}                                   & { 42.8}                                   & { 15.4}                              & { 36.6}                                 & { 31.8}                             \\
    { \textbf{ELICIT (1 layer)}}      & { 38.3}                             & { 46.9}                                   & { 60.7}                                   & { 20.6}                              & { 51.1}                                 & { 43.5}                             \\
    { \textbf{ELICIT (3 layers)}}     & { 38.2}                             & { \textbf{47.1}}                          & { 61}                                     & { 21.6}                              & { 51.6}                                 & { 43.9}                             \\
    { \textbf{ELICIT (all layers)}}   & { \textbf{40.9}}                    & { 46.3}                                   & { \textbf{61.4}}                          & { \textbf{21.7}}                     & { \textbf{52.4}}                        & { \textbf{44.5}}                    \\ \bottomrule
    \end{tabular}%
    \end{table}

\section{Analysis of Computational Efficiency with Retrieval Module}

To demonstrate the effciency of \alg, We conducted a detailed analysis of ELICIT's computational efficiency using the Pythia-6.9B model, measuring the average processing time per sample across different pipeline stages.
The results are shown in Table~\ref{table:time_table}.
Our quantitative results demonstrate that the integration of the retrieval module maintains the method's efficiency.
Specifically, the retrieval module adds only 0.105 seconds of computational overhead per sample.
The total inference time, including retrieval operations, remains efficient at 0.172 seconds per sample.
ELICIT demonstrates superior efficiency compared to baseline approaches, processing samples 2-3 times faster than both 16-shot inference and BM25-based inference methods.
These results validate that ELICIT's performance improvements do not come at the cost of computational efficiency, even with the addition of retrieval module.

\vspace{10pt}
\begin{table}[H]
    \caption{
            The running time of different stages per sample across different domains.
    }
    \label{table:time_table}
    \resizebox{\columnwidth}{!}{%
    \begin{tabular}{c|cccc|c}
    \toprule
    {\textbf{}}          & {\textbf{zs inference time}} & {\textbf{ELCIT inference time}} & {\textbf{retrieve time}} & {\textbf{bm25 inference time}} & {\textbf{16shot inference time}} \\ \midrule
    {\textbf{nlu}}       & {0.063}                      & {0.064}                         & {0.097}                  & {0.302}                        & {0.181}                          \\
    {\textbf{reasoning}} & {0.065}                      & {0.066}                         & {0.104}                  & {0.349}                        & {0.315}                          \\
    {\textbf{knowledge}} & {0.066}                      & {0.069}                         & {0.108}                  & {0.517}                        & {0.371}                          \\
    {\textbf{math}}      & {0.065}                      & {0.067}                         & {0.111}                  & {0.351}                        & {0.352}                          \\
    {\textbf{safety}}    & {0.067}                      & {0.069}                         & {0.104}                  & {0.611}                        & {0.366}                          \\ \midrule
    {\textbf{avg}}       & {0.065}                      & {0.067}                         & {0.105}                  & {0.426}                        & {0.317}                          \\ \bottomrule
    \end{tabular}%
    }
    \end{table}

\end{document}